\documentclass{llncs}

\usepackage{amsmath}
\usepackage{multirow}
\usepackage{cite}
\usepackage{graphicx}
\usepackage{xcolor}
\usepackage{subcaption}
\usepackage[hidelinks]{hyperref}  
\usepackage{tikz}
\usetikzlibrary{matrix}

\newtheorem{mytheorem}{Theorem}
\newtheorem{mydef}{Definition}

\usepackage{orcidlink}
\usepackage{todonotes}

\begin{document}

\title{Identifying the Truth of Global Model: A Generic Solution to Defend Against Byzantine and Backdoor Attacks in Federated Learning (full version)}

\author{Sheldon C. Ebron\and
Meiying Zhang \and
Kan Yang\orcidlink{0000-0003-4234-9596}}
\institute{Dept. of Computer Science, University of Memphis, USA\\
\email{\{sebron, mzhang6, kan.yang\}@memphis.edu}}
\authorrunning{S. Ebron et al.}

\maketitle
\begin{abstract}
Federated Learning (FL) enables multiple parties to train machine learning models collaboratively without sharing the raw training data. However, the federated nature of FL enables malicious clients to influence a trained model by injecting error model updates via Byzantine or backdoor attacks. To detect malicious model updates, a typical approach is to measure the distance between each model update and a \textit{ground-truth model update}. To find such \textit{ground-truth model updates}, existing defenses either require a benign root dataset on the server (e.g., FLTrust) or simply use trimmed mean or median as the threshold for clipping (e.g., FLAME). However, such benign root datasets are impractical, and the trimmed mean or median may also eliminate contributions from these underrepresented datasets. 
In this paper, we propose a generic solution, namely FedTruth, to defend against model poisoning attacks in FL, where the \textit{ground-truth model update} (i.e., the global model update) will be estimated among all the model updates with dynamic aggregation weights. Specifically, FedTruth does not have specific assumptions on the benign or malicious data distribution or access to a benign root dataset. Moreover, FedTruth considers the potential contributions from all benign clients. Our empirical results show that FedTruth can reduce the impacts of poisoned model updates against both Byzantine and backdoor attacks, and is also efficient in large-scale FL systems. 


\end{abstract}


\section{Introduction}
In traditional machine learning, training data is usually hosted by a centralized server (cloud server) that runs the learning algorithm or is shared among a set of participating nodes for distributed learning. However, in many applications, data cannot be shared with the cloud or other participating nodes due to privacy or legal restrictions, especially when multiple organizations are involved. 
Federated Learning (FL) allows multiple parties, such as clients or devices, to collaboratively train machine learning models without sharing raw training data~\cite{mcmahan2017communication}. All selected clients train the global model on their local datasets and send the local model updates to an aggregator. The aggregator then aggregates all the local model updates and sends the new global model to all the clients selected for the next round of training until convergence is reached. The FL framework is suitable for many AI-driven applications where data is sensitive or legally restricted, such as smart healthcare (e.g., cancer prediction\cite{kourou2015machine}
), smart transportation (e.g., autonomous driving \cite{sallab2017deep}), smart finance (e.g., fraud detection \cite{fiore2019using}), and smart life (e.g., surveillance object detection \cite{zhao2019object}). 

However, the federated nature of FL enables malicious clients to influence a trained model by injecting error model updates. For example, adversaries can control a set of clients to launch \textit{Byzantine attacks} \cite{blanchard2017machine, chen2017distributed} (i.e., sending arbitrary model updates to make the global model converge to a sub-optimal model), or \textit{backdoor attacks} \cite{bhagoji2019analyzing, bagdasaryan2020backdoor, xie2019dba, wang2020attack} (i.e., manipulating local model updates to cause the final model to misclassify certain inputs with high confidence).

 
Towards model poisoning attacks in FL, existing defenses focus on designing robust aggregation rules by: 
\begin{itemize}
    \item \textit{clustering and removing.}
    This approach identifies malicious model updates by clustering model updates (e.g., Krum \cite{blanchard2017machine}, AFA \cite{munoz2019byzantine}, FoolsGold \cite{fung2020limitations} and Auror\cite{shen2016auror}). However, they only work under specific assumptions about the underlying data distribution of malicious clients and benign clients. For example, Krum and Auror assume that the data of benign clients are independent and identically distributed (iid), whereas FoolsGold and AFA assume the benign data are non-iid. Moreover, these defenses cannot detect stealthy attacks (e.g., constraint-and-scale attacks \cite{bagdasaryan2020backdoor}) or adaptive attacks (e.g., Krum attack \cite{fang2020local}). 
    
    \item \textit{clipping and noising}. This approach clips individual weights with a certain threshold and adds random noise to the weights so that the impact of poisoned model updates on the global model can be reduced \cite{bagdasaryan2020backdoor, nguyen2021flame}. In \cite{nguyen2021flame}, the authors propose FLAME, which first applies clustering to filter model updates and then uses clipping and noising with an adaptive clipping threshold and noise level. However, the clipping and noising also eliminate the contributions from benign clients with underrepresented datasets.
    
    \item \textit{trimming and averaging.} This approach finds the mean or median of each weight in the remaining model updates after removing some values that are bigger/smaller than some thresholds (trimmed mean or median \cite{yin2018byzantine}) or with low frequency (FreqFed \cite{fereidooni2024freqfed}). However, the trimmed mean or median can be easily bypassed using adaptive attacks (e.g., Trim attack \cite{fang2020local}). 

    \item \textit{adjusting aggregation weights using root data \cite{cao2021fltrust}. } This approach assigns different weights based on the distance between each model update and the benign model update from the root dataset. However, it requires the aggregator to access the benign root dataset. 
\end{itemize}
 Recently, several works \cite{kusetogullari2020ardis, gorbunov2023variance, farhadkhani2022byzantine} have been proposed to achieve provable Byzantine robustness by integrating variance-reduced algorithms and byzantine-resilient aggregation algorithms. However, they require prior knowledge of the
 variance of the gradients \cite{kusetogullari2020ardis, gorbunov2023variance} or only focus on existing byzantine-resilient aggregation algorithms.

{\textbf{Motivation}}: Based on the above-discussed defenses, we have the following observations:
\begin{enumerate}
    \item  Without knowing clients' local datasets or a benign root dataset, it is difficult to determine whether an outlier is a malicious update or a significant contribution from an underrepresented dataset, especially when local datasets are non-iid. It is not a good idea to remove or clip a benign outlier model update with a significant contribution from under-representative data. 

    \item Only one representative model update is chosen as the global model in many existing Byzantine-resilient aggregation algorithms (e.g., Krum \cite{blanchard2017machine}, trimmed median \cite{yin2018byzantine}), which means the global model is trained with only a single local dataset in each round. In other words, the efforts and contributions of other clients are wasted; 

    \item Due to various qualities of data and trained local model, it is unfair to treat all the clients equally (e.g., FLAME\cite{nguyen2021flame}, FreqFed\cite{fereidooni2024freqfed}) or evaluate client contributions based on the size of the training dataset (e.g., FedAvg \cite{mcmahan2017communication}) during the model aggregation.
\end{enumerate}

This paper aims to design a {generic solution} to defend against model poisoning attacks in FL with the following properties: 1) it does not have specific assumptions on benign or malicious data distribution or accessing to a benign root dataset; 2) it considers potential contributions from all the benign clients (including those with under-representative data); and 3) it reduces the impacts of poisoned model updates from malicious clients. Specifically, we propose a new model aggregation algorithm, namely FedTruth, which enables the aggregator to find the truth among all the received local model updates. The basic idea of FedTruth is inspired by truth discovery mechanisms \cite{li2016conflicts, yin2008truth, li2015discovery, ouyang2016aggregating}, which are developed to extract the truth among multiple conflicting pieces of data from different sources under the assumption that the source reliability is unknown a priori. 
\textit{{In each round of FedTruth, the global model update (i.e., \textit{ground-truth model update}) will be computed as a weighted average of all the local model updates with dynamic weights}}.

The contributions of this paper are summarized as follows: 
\begin{itemize}
    \item We develop FedTruth, a generic solution to defend against model poisoning attacks in FL. Compared with existing solutions, FedTruth eliminates the assumptions of benign or malicious data distribution and the need to access a benign root dataset. 
    
    \item We propose a new approach to estimate the \textit{ground-truth model update} among all the model updates with dynamic aggregation weights in each round. Different from the FedAvg \cite{mcmahan2017communication} (where the aggregation weight is determined by the size of training dataset) or FLAME \cite{nguyen2021flame} (where equal aggregation weight is used regardless of the size of training dataset), the aggregation weights in FedTruth are dynamically chosen based on the distances between the estimated truth and local model updates, following the principle that higher weights will be assigned to more reliable clients. 
    
    \item We extensively evaluate the robustness of our FedTruth against both Byzantine attacks (model-boosting attack, Gaussian noise attack, and local model amplification attack) and backdoor attacks (distributed backdoor attack, edge case attack, projected gradient descent attack) under three attacking strategies (base attack, with model-boosting, and with constrain-and-scaling). The experimental results show that FedTruth can reduce the impacts of poisoned model updates against both Byzantine and backdoor attacks. Moreover, FedTruth works well on both iid and non-iid datasets. 

    \item We further evaluate the efficiency of the FedTruth in terms of the number of iterations to reach FedTruth convergence and the time consumption for two deployments: FedTruth (with entire model updates as inputs) and FedTruth-Layer (deploy FedTruth in each layer of the model). The results show that our methods are efficient in large-scale FL systems. 
\end{itemize}

The remainder of this paper is organized as follows: 
Section~\ref{Sec:ProblemStatement} presents the problem statement in terms of the system model, threat model, and design goals. Then, we describe the technical overview of our proposed FedTruth, followed by the concrete formulation. Section~\ref{Sec:evaluation} shows the key experimental results against both Byzantine attacks and Backdoor attacks. 
Section~\ref{sec:relatedwork}, we describe the related work.
{Section~\ref{Sec:conclusion}} concludes the paper. In the appendices, we provide detailed model poisoning attacks, more experimental results against these attacks using ResNet-18 (CIFAR-10) and CNN (FMNIST) models, discussions on the distance function in FedTruth and the impact of non-iid data on FedTruth.

\section{Problem Statement}\label{Sec:ProblemStatement}

\textbf{Federated Learning}:
A general FL system consists of an aggregator and a set of clients $S$. Let $\mathcal{D}_k$ be the local dataset held by the client $k~(k\in S)$. The typical FL goal \cite{mcmahan2017communication} is to learn a model collaboratively without sharing local datasets by solving 
\begin{equation*}
    \min_{w} F(w)  = \sum_{k\in S} a_k\cdot F_k(w),~s.t.~\sum_{k\in S} a_k  = 1 ~(a_k\geq 0),
\end{equation*}
where 
\[
F_k(w) = \frac{1}{n_k}\sum_{j_k = 1}^{n_k} f_{j_k}(w; x^{(j_k)}, y^{(j_k)})
\]
is the local objective function for a client $k$ with $n_k = |\mathcal{D}_k|$ available samples. $a_k$ is the aggregation weights, which are usually set as $a_k = n_k/\sum_{k\in S} n_k$ (e.g., FedAvg \cite{mcmahan2017communication}). The FL training process usually contains multiple rounds, and a typical FL round consists of the following steps: 
\begin{enumerate}
    \item \textit{client selection and model update}: a subset of clients $S_t$ is selected, each of which retrieves the current global model $w_t$ from the aggregator.
    \item \textit{local training}: each client $k$ trains an updated model $w^{(k)}_t$ with the local dataset $\mathcal{D}_k$ and shares the model update $\Delta_t^{(k)} = w_t - w_t^{(k)}$ to the aggregator.
    \item \textit{model aggregation}: the aggregator computes the global model updates as $\Delta_t = \sum_{k\in S_t} a_k \Delta_t^{(k)}$ and update the global model as $w_{t+1} = w_t - \eta \Delta_t$, where $\eta$ is the server learning rate.
\end{enumerate}

{FedAvg ~\cite{mcmahan2017communication}} is the original aggregation rule, which 
averages all local model weights selected based on the number of samples the participants used. FedAvg has been shown to work well when all the participants are benign clients, but is vulnerable to model poisoning attacks. 

\begin{figure}[!t]
\centering
\includegraphics[width=0.9\textwidth]{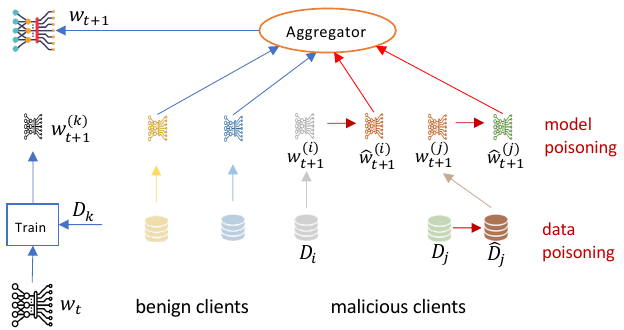}
\caption{System Model}
\label{fig:system_model}
\end{figure}

\textbf{System Model}: As shown in Fig. \ref{fig:system_model}, we consider a typical FL setting, which consists of two entities: 
\begin{itemize}
    \item \textbf{Clients}: FL clients are users who participate in the FL process with their end devices, e.g., mobile devices, computers, and vehicles. When selected in an FL round, the clients will train the model based on their local datasets and send local model updates to the aggregator. Due to personal schedules and device status, the group of clients will change dynamically in each FL round. For example, some clients may not be able to send model updates due to low battery or unstable network, and some clients may join the FL task in intermediate FL rounds. 
    \item \textbf{Aggregator}: The aggregator is an entity that runs the FL algorithm with the clients, including distributing the initial model to all the selected clients, aggregating local model updates, and sending the global model to the clients selected in a new round. 
\end{itemize}



\textbf{Threat Model}:
In this paper, we assume that the aggregator will aggregate all the local model updates honestly in each FL round. However, the clients may be 
compromised by adversaries and collude to launch Byzantine attacks and backdoor attacks. We assume that the adversaries cannot compromise more than half clients selected in each round. 
When launching an attack, the adversaries can directly modify their local models (model poisoning attack) and local datasets (data poisoning attack) while having full knowledge about the system (having direct access to any information shared through the system training). However, the adversaries cannot access the benign clients??? devices or data.  During a Byzantine attack, the adversarial goal is to degrade the global model or prevent it from convergence, while the backdoor attack aims to manipulate the global model by injecting it with a targeted backdoor. 

\textbf{Design Goals}: We aim to design a \textbf{generic solution} to defend against model poisoning attacks in FL with the following properties: 1) it does not have specific assumptions on benign or malicious data distribution or accessing to a benign root dataset; 2) it considers the potential contributions from all the benign clients (including those with under-representative data); and 3) it reduces the impacts of poisoned model updates from malicious clients.



\section{FedTruth: Truth of Global Model}\label{Sec:FedTruth}

\subsection{Technical Overview}

In FedAvg \cite{mcmahan2017communication}, the aggregation weight is determined by the size of the training dataset (i.e., $a_k = n_k/\sum_{k=1}^{m} n_k$) where $n_k = |\mathcal{D}_k|$. In some other works, such as FLAME \cite{nguyen2021flame}, equal aggregation weight is used regardless of the size of the training dataset (i.e., $a_k = 1/m$). However, neither FedAvg nor equal aggregation weights can reflect the performance of a client. 
In FLTrust \cite{cao2021fltrust}, the authors proposed using dynamic aggregation weights to calculate the global model. The aggregation weights are estimated based on the trust values, which are calculated based on the similarity between each model update with a \textit{ground-truth model update}. This \textit{ground-truth model update} is trained by the aggregator using a benign root dataset. However, this benign root dataset may not be practical in many applications. 

Without a benign root dataset, it is challenging to obtain the \textit{ground-truth model update} among all the local model updates in an FL round. We propose a new model aggregation algorithm, namely FedTruth, which enables the aggregator to uncover the truth among all the received local model updates. The basic idea of FedTruth is inspired by truth discovery mechanisms \cite{li2016conflicts, yin2008truth, li2015discovery}, which are developed to extract the truth among multiple conflicting pieces of data from different sources under the assumption that the source reliability is unknown a priori. Unlike FLTrust, in FedTruth, we do not obtain the \textit{ground-truth model update} and use it to calculate the aggregated weights. Instead, the \textit{ground-truth model update} is actually the aggregated global model update, which is computed as the weighted average of all the local model updates with dynamic aggregation weights for each round. The aggregation weights in FedTruth are dynamically calculated based on the distances between the estimated truth and local model updates, following the principle that higher weights will be assigned to more reliable clients. 

Although the truth discovery approach has been used in {RobustFed}~\cite{tahmasebian2022robustfed} and \textit{TDFL}~\cite{xu2022tdfl}, they simply apply the CRH truth discovery algorithm \cite{li2014resolving} which may still suffer from Byzantine attack or potentially magnifying the local model updates, see the related work for details. Here, we present a generic formulation with a coefficient function and a linear constraint, establishing the necessary conditions for the coefficient function to ensure the convexity and convergence of FedTruth. Furthermore, we demonstrate that our proposed FedTruth is also effective in defending against backdoor attacks, such as the Edge Case \cite{wang2020attack}, DBA \cite{xie2019dba}, and PGD \cite{wang2020attack} attacks.

\subsection{Formulation of FedTruth}


Suppose the aggregator receives $n_t$ different model updates $\Delta_{t}^{(1)}, \cdots$, $\Delta_{t}^{(n_t)}$ in FL round $t$. To find the global update $\Delta^*_{t}$, we 
formulate an optimization problem aiming at minimizing the total distance between all the model updates and the estimated global update: 
\begin{equation}\label{equ:fedtruth}
    \min_{\Delta^*_{t}, {\bf p_{t}}} D(\Delta^*_{t}, {\bf p_{t}}) = \sum_{k=1}^{n_t} g(p_{t}^{(k)})\cdot d(\Delta^*_{t}, \Delta_{t}^{(k)})
    ~~~ s.t.~~~\sum_{k=1}^{n_t} p_{t}^{(k)}  = 1
\end{equation}
where $d(\cdot)$ is the distance function and $g(\cdot)$ is a non-negative coefficient function. $p_{t}^{(k)}$ is the performance of the local model $\Delta_{t}^{(k)}$ which is calculated based on the distance. Note that, to better understand the performance of each client, our optimization problem is formulated based on the performance values $p_{t}^{(k)}$ rather than directly on the aggregation weights $a_t^{(k)}$. 
The aggregation weights can be easily calculated based on the performance value. 

There are many different choices of the distance function $d(\cdot)$, such as Euclidean distance ($ d(\Delta^*_{t}, \Delta_{t}^{(k)}) = ||\Delta^*_{t} - \Delta_{t}^{(k)}||$) and angular distance ($d(\Delta^*_{t}, \Delta_{t}^{(k)}) = 1- S_c(\Delta^*_{t} , \Delta_{t}^{(k)})$,
where $S_c$ is the cosine similarity. 

\subsection{Solving the optimization problem} 
We iteratively compute the estimated truth ${\Delta}^*_t$ and the performance values ${\bf p^t}$ using coordinate descent approach \cite{bertsekas1997nonlinear}. Specifically, given an initial global model update ${\Delta}^*_t$ (can be the result of FedAvg or simple average), the algorithm will update the performance values ${\bf p^t}$ to minimize the objective distance function. Then, it updates aggregation values $a_t^{(k)}$ and uses them to further estimate the new global model update ${\Delta}^*_t$. 
\begin{itemize}
    \item 
\textbf{\textit{Updating Aggregation Weights:}}
Once the truth $\Delta^*_{t}$ is fixed, we first calculate the performance of each model update $\{p_t^{(k)}\} (k=1,\cdots, n_t)$ as 
    $p_t^{(k)} = {d(\Delta^*_{t}, \Delta_{t}^{(k)})}/{\sum\limits_{k'=1}^{n_t} d(\Delta^*_{t}, \Delta_{t}^{(k')})}.$
Then, the aggregation weights can be updated as 
\begin{equation}
    a_t^{(k)} = {g(p_t^{(k)})}/{\sum_{k=1}^{n_t} g(p_t^{(k)})}.
\end{equation}

\item \textbf{\textit{Updating the Truth:}}
Based on the new aggregation weights $\{a_t^{(1)}, \cdots, a_t^{(n_t)}\}$, the truth of global update can be estimated as 
    $\Delta^*_{t} = \sum_{k=1}^{n_t} a_t^{(k)}\cdot  \Delta_{t}^{(k)}$
\end{itemize}
The global model update and aggregation weights will be updated iteratively until convergence criteria are met. It is easy to see that the longer the distance between the local model update and the estimated truth, the smaller the aggregation weight will be assigned in calculating the truth. This principle can eliminate the impacts of malicious model updates and keep certain contributions from a benign outlier model update.

\subsection{Convergence Guarantee of FedTruth}
We use the Lagrange multipliers to solve the optimization problem.
Under the linear constraint $\sum_{k=1}^{n_t} p_{t}^{(k)}  = 1$, we can define the Lagrangian function of Eq. \ref{equ:fedtruth} as
 \begin{equation*}
 L(\{p^{(k)}_{t}\}_{k=1}^{n_t},\lambda) = \sum_{k=1}^{n_t} g(p^{(k)}_{t})\cdot d(\Delta^*_{t}, \Delta_{t}^{(k)})+\lambda(\sum_{k=1}^{n_t} p_{t}^{(k)}  - 1), \\
 \end{equation*}
    where $\lambda$ is a Lagrange multiplier.
To solve the optimization problem, we set the partial derivative with $p^{(k)}_{t}$ to zero:
 \begin{equation}\label{equ_gw}
 g'(p^{(k)}_{t})\cdot d(\Delta^*_{t}, \Delta_{t}^{(k)})+\lambda=0
 \end{equation}
Then, the Eq. \ref{equ_gw} can be reformulated as:
 \begin{equation}\label{equ_weight}
p^{(k)}_{t} =g'^{-1}({-\lambda}/{d(\Delta^*_{t}, \Delta_{t}^{(k)})})
 \end{equation}
Since the linear constraint is $\sum_{k=1}^{n_t}p^{(k)}_{t}=1$, the $\lambda$ and $p^{(k)}_{t}$ can be derived from Eq. \ref{equ_weight}.

We can see that $g(\cdot)$ must be monotonous and differentiable in the aggregation weight domain in order to guarantee the existence of $g'^{-1}(\cdot)$. Moreover, according to the principle of truth discovery, $g(\cdot)$ should be a decreasing function. Some simple but effective coefficient functions are as follows: 
\begin{equation}
     g(p_t^{(k)}) = 1/p_t^{(k)}~~\text{or}~~ g(p_t^{(k)}) = - \log(p_t^{(k)}).
\end{equation}

Therefore, we say that as long as the coefficient function $g(\cdot)$ is monotonous, decreasing, and differentiable in the aggregation weight domain, the convexity and convergence of FedTruth can be guaranteed. 
From our experiments, we find that after 5 to 40 iterations of coordinate descent, the estimated truth is close to the converged value.

\subsection{Proof of Byzantine-Resilience}
In \cite{farhadkhani2022byzantine}, the authors proposed a formal definition of Byzantine-resilience of aggregation algorithm, namely $(f, \lambda)$-Resilient Averaging. 

\begin{mydef}[$(f, \lambda)$-Resilient Averaging\cite{farhadkhani2022byzantine}]
    For $f < n$ and real value $\lambda \leq 0$, an aggregation rule $F$ is $(f, \lambda)$-Resilient Averaging if for any collection of $n$ vectors $x_1, 
    \cdots, x_n$, and any set $S\subset \{1, \cdots, n\}$ of size $n-f$, the following condition holds
    \[
        ||F(x_1, \cdots, x_n) - \sum_{i\in S} \frac{1}{n-f} x_i||\leq \lambda \cdot \max_{i, j\in S} ||x_i - x_j||.
    \]
\end{mydef}

Under this definition, we show the Byzantine-resilience of FedTruth as in the following theorem:
\begin{mytheorem}
    FedTruth is $(f, 1)$-resilient averaging, where $f < n/2$. 
\end{mytheorem}
\begin{proof}
    In FedTruth, the aggregated global model (i,e., the truth) is calculated as the weighted average of all the model updates: 
    \[
        FedTruth(x_1, \cdots, x_n) = \sum_{j\in [1, n]} a_j x_j 
    \]
    where the aggregation weights $a_j$ are dynamically calculated and $\sum_{j\in [1, n]} a_j = 1$.

    For an arbitrary set $S \in \{1, \cdots, n\}$ of size $n-f$, we can rewrite the average of those weights in the set $S$ as 
    \[
        \sum_{i\in S} \frac{1}{n-f} x_i = \sum_{i \in S} b_i x_i
    ~~\text{where}~~ 
    b_i = \frac{1}{n-f} 
    ~~~\text{and}~~~ \sum_{i \in S} b_i = 1.
    \]
    
    Then, we can obtain the difference between the truth and the average of set $S$ as 
    \[
    \begin{split}
        & ||F(x_1, \cdots, x_n) - \sum_{i\in S} \frac{1}{n-f} x_i|| = ||\sum_{j\in [1, n]} a_j x_j - \sum_{i \in S} b_i x_i|| \\
        & = ||\sum_{j\in [1, n]} a_j (x_j - \sum_{i \in S} b_i x_i) ||
         = ||\sum_{j\in [1, n]} a_j \big(\sum_{i \in S} b_i(x_j -  x_i)\big)|| \\
        & \leq \sum_{j\in [1, n]} a_j \big(\sum_{i \in S} b_i||x_j -  x_i||\big) \leq \sum_{j\in [1, n]} a_j \big(\sum_{i \in S} b_i \max_{i\in S, j \in [1, n]}||x_j -  x_i||\big) 
    \end{split}
    \]
    BEGCuse $S$ is an arbitrary set of size $n-f$, we say that 
    \[
    \sum_{i \in S} b_i \max_{i\in S, j \in [1, n]}||x_j -  x_i|| \leq \sum_{i \in S} b_i \max_{i,j\in S}||x_j -  x_i||
    \]
    Then, we have 
    \[
    ||F(x_1, \cdots, x_n) - \sum_{i\in S} \frac{1}{n-f} x_i|| \leq \sum_{j\in [1, n]} a_j \sum_{i \in S} b_i \max_{i,j\in S}||x_j -  x_i|| = \max_{i,j\in S}||x_j -  x_i||
    \]
    \qed
\end{proof}

\subsection{Resisting against Adaptive Attack on FedTruth}\label{Sec:adaptive}


In an adaptive attack targeting the Euclidean distance metric, an attacker might be capable of designing an alternative local model, denoted as $w^*$, such that its Euclidean distance from the baseline model (for instance, the ground truth $G$) is identical to the Euclidean distance between the actual local model $w$ and the baseline model $G$. This scenario is feasible if the baseline model remains static and is accessible to the attacker. However, in the FedTruth framework, the baseline model is not a constant; instead, it evolves and is progressively estimated over multiple iterations.

The effectiveness of FedTruth relies on the assumption that the majority of clients are reliable and diverse. If an adversary compromises more than 50\% of the clients, they can dominate the results of FedTruth. 
From our experimental results, we find that when an adversary compromises 40\% (4 out of 10) clients in each round, FedTruth can still prevent Byzantine attacks, as seen in Figure~\ref{fig:model-boosting-attacks-mnist}. However, when the non-iid degree is further increased to 95\%, as shown in Figure~\ref{fig:model-boosting-noniid}, the accuracy drops and convergence speed becomes slow bEGCuse some uncompromised clients may perform poorly with highly non-iid training data, leading to a bad estimation of the ground truth by FedTruth. However, our results outperformed all other aggregation algorithms during this experiment, excluding FLTrust. 

To counter this, we propose strategies like filtering out inputs from historically unreliable clients, thereby reducing malicious influence. Although FedTruth and FedTruth-Layer aim to consider the contributions of all clients, it may be necessary to exclude inputs from clients who have a bad reputation or low reliability in previous FL tasks. To evaluate the reputation or reliability of clients, we need to assess the contributions of each client in an FL task. This motivated us to formulate FedTruth with linear constraints.
In practice, we can trim the clients' inputs who have been identified as untrustworthy or unreliable based on their past contributions to FL tasks. By doing so, we can further improve the accuracy and robustness of the global model by preventing the contributions of bad actors from affecting the overall performance. 
We should also be aware that trimming inputs from clients may have unintended consequences, such as reducing the diversity of the training data and reducing the number of participating clients, potentially leading to overfitting and decreased overall performance. Therefore, we need to carefully evaluate the trade-offs between trimming inputs and maintaining the diversity of the training data. 
Additionally, leveraging clustering algorithms to categorize model updates before aggregation can help FedTruth remain effective even when faced with a majority of malicious clients.

\subsection{Deploying FedTruth in Each Layer?} 
One major challenge in truth discovery is data heterogeneity, which may include non-structured data and missing values. However, this challenge is not applicable to FedTruth bEGCuse all the local model updates are in the same structure. For example, in deep neural networks, the model updates can be represented as multiple-layer tensors. 
FedTruth can be run for just one time by the aggregator to compute the truth of model updates by feeding all the local model updates into the FedTruth algorithm. This deployment treats the local model update as an observed value in the truth discovery algorithms. 
Also, we can deploy FedTruth on each layer to estimate the truth of that single layer, which means that the weights allocated to all the clients may vary on different layers. We denote this deployment as FedTruth-Layer. Such layer-wise deployment seems reasonable, it also brings the computation overhead which is linear to the number of layers. We compare the efficiency between FedTruth and FedTruth-Layer in the Section \ref{sec:efficiency}.


\section{Experimental Results}\label{Sec:evaluation}
We compare the performance of FedTruth with the state-of-the-art aggregation algorithms: FedAvg \cite{mcmahan2017communication}, Krum~\cite{blanchard2017machine}, Trimmed mean~\cite{yin2018byzantine}, Median~\cite{yin2018byzantine}, FLTrust~\cite{cao2021fltrust}  and FLAME~\cite{nguyen2021flame}. The Gaussian noise and backdoor attacks are implemented with three attacking strategies: 
\begin{enumerate}
\item \textit{base attack}: During the base attacks, the attacker will not boost the poisoning model or model updates. 

\item \textit{combine with model-boosting attack}: The poisoning model or model updates will be boosted with a boosting factor. Similar to \cite{wang2020attack}, we set the boosting factor as $x = C_{t}/C_{adv, t}$, where $C_t$ denotes the total number of clients selected in $t$-th round, and $C_{adv, t}$ denotes the number of adversarial clients in this round. In our experiment, we have 10 clients selected in each FL round, and the default number of adversarial clients is 3 in this section. So, here the default number of boosting factor is $x = 10/3$ for all the figures in this section.

\item \textit{combine with constrain-and-scaling attack}: We implement this attack by letting each adversarial client train a benign local model $w_{t}^{j, b}$ first like a benign client. Then, the adversarial client produces a poisoning model $w_{t}^{j, p}$ according to the attack. A smoothed poisoning model will be calculated as $w_{t}^{j} = \alpha w_{t}^{j, b} + (1-\alpha) w_t^{j, p}$. In our experiment, the default value of $\alpha$ is 0.5. Finally, this value will be scaled or boosted by a boosting factor similar to the \textit{model-boosting attack. }
\end{enumerate}   


\subsection{Experimental Settings}\label{Sec:experimentalSetup}

The experimental settings are as follows: 

\textbf{\textit{Datasets}}:
We conduct the experiments with MNIST~\cite{deng2012mnist}, FMNIST~\cite{xiao2017fashionmnist}, and CIFAR-10~\cite{krizhevsky2009learning} datasets. FMNIST and MNIST are comprised of 60,000 black-and-white labeled images of size (28×28). MNIST contains handwritten digits, and FMNIST consists of images of clothing items. CIFAR-10 consists of 60000 color images of size (32x32). 
During the \textit{edge-case attack} experiment, we used the \textit{Arkiv Digital Sweden (ARDIS)~\cite{kusetogullari2020ardis}} dataset as the adversarial backdoor images. The ARDIS dataset consists of handwritten digits originating from Swedish church records. This dataset is suitable for targeted images when inserting backdoors into MNIST, as ARDIS entirely consists of naturally occurring edge cases.


\textbf{\textit{Clients}}: When crafting the clients' local datasets, we draw their datapoints randomly from a \textit{non-iid} distribution. We use a \textit{non-iid} distribution bEGCuse it better represents clients' data in practice than an \textit{iid} distribution. 
The client's local data is generated a \textit{non-iid} distribution, where the \textit{bias} parameter default is $0.8$. In addition, we evaluate the impact of \textit{non-iid} degree using the \textit{model-boosting}  attacks, as seen in {Section~\ref{Sec:noniid}. In each FL round, we randomly select 10 clients and choose a subset of these selected clients as adversarial clients. 

\textbf{\textit{Models}}: We constructed a Convolutional Neural Network (CNN) classifier for all the experiments considered in this work. It includes an input layer (28x28x1), two convolutional layers with ReLU activation (20x5x1 and 4x4x50), two max pooling layers (2x2), a fully connected layer with ReLU (500 units), and a final fully connected layer with Softmax (10 units).
The ResNet-18 model was used when running experiments using the CIFAR-10 dataset. We ran both the MNIST and CIFAR10 experiments on an AWS (g6.xlarge) EC2 instance. When running the FedTruth and FedTruth-Layer experiments, we set our convergence threshold to $1\mathrm{e}{-6}$.

\subsection{Byzantine Attacks}

This section presents experimental results for two attacks: the \textit{model-boosting} and \textit{Gaussian noise attacks}, conducted on various FL frameworks. In these attacks, only the model updates (the difference between the newly trained local model and the global model in the previous round) are communicated. {Appendix~\ref{app:moremodelboosting} and Appendix~\ref{app:moregaussiannoise}} contain findings from Byzantine experiments performed on the FMNIST and CIFAR-10 datasets. 

\textbf{Model-boosting Attack}: The model-boosting attack seeks to degrade the model's performance by boosting the adversary's local updates by a multiple of 10. The subset of compromised clients are randomly selected each round. We conducted experiments with varying percentages of compromised clients in each round to evaluate the robustness of different aggregation algorithms under different attacking scenarios.

\begin{figure*}[!h]
    \captionsetup{font={tiny}} 
    \centering
    \begin{subfigure}{\textwidth}
        \centering
        \caption{Model Boosting Attack (MNIST, 0 Adversaries)}
        \label{fig:model-boosting-attacks-mnist-a}
        \includegraphics[clip,trim={0 3.9cm 0 0}, width=0.9\textwidth]{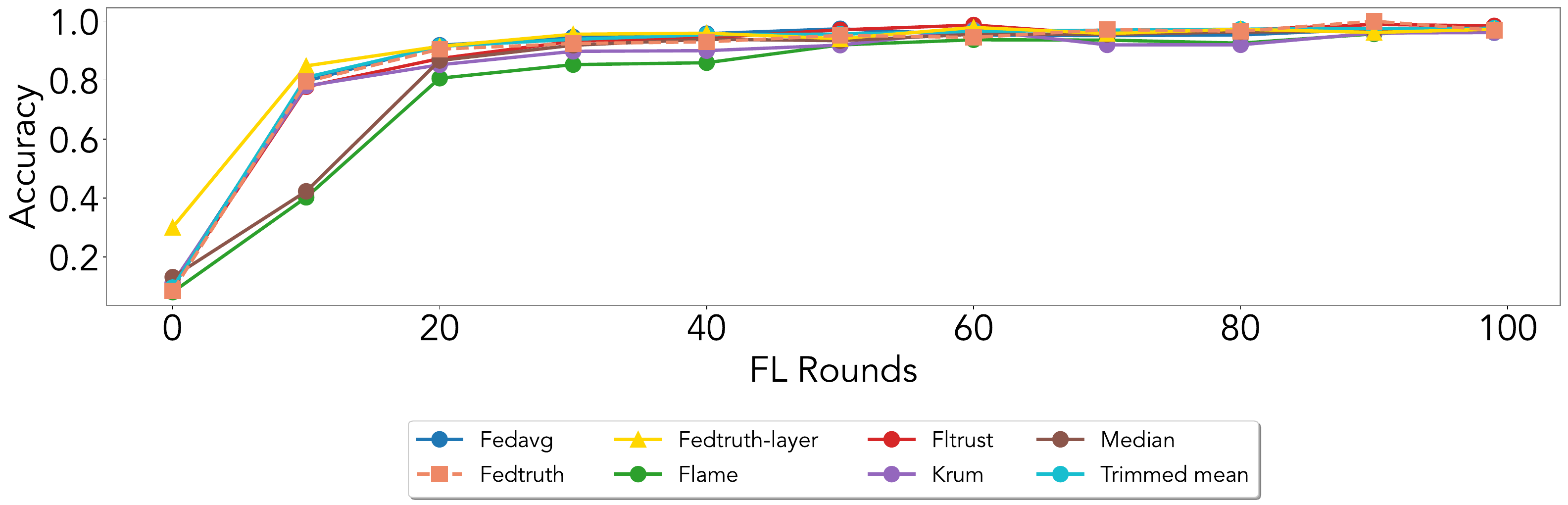}
    \end{subfigure}
    \begin{subfigure}{\textwidth}
        \centering
        \caption{Model Boosting Attack (MNIST, 1 Adversaries)}
        \label{fig:model-boosting-attacks-mnist-b}
        \includegraphics[clip,trim={0 3.9cm 0 0},width=0.9\textwidth]{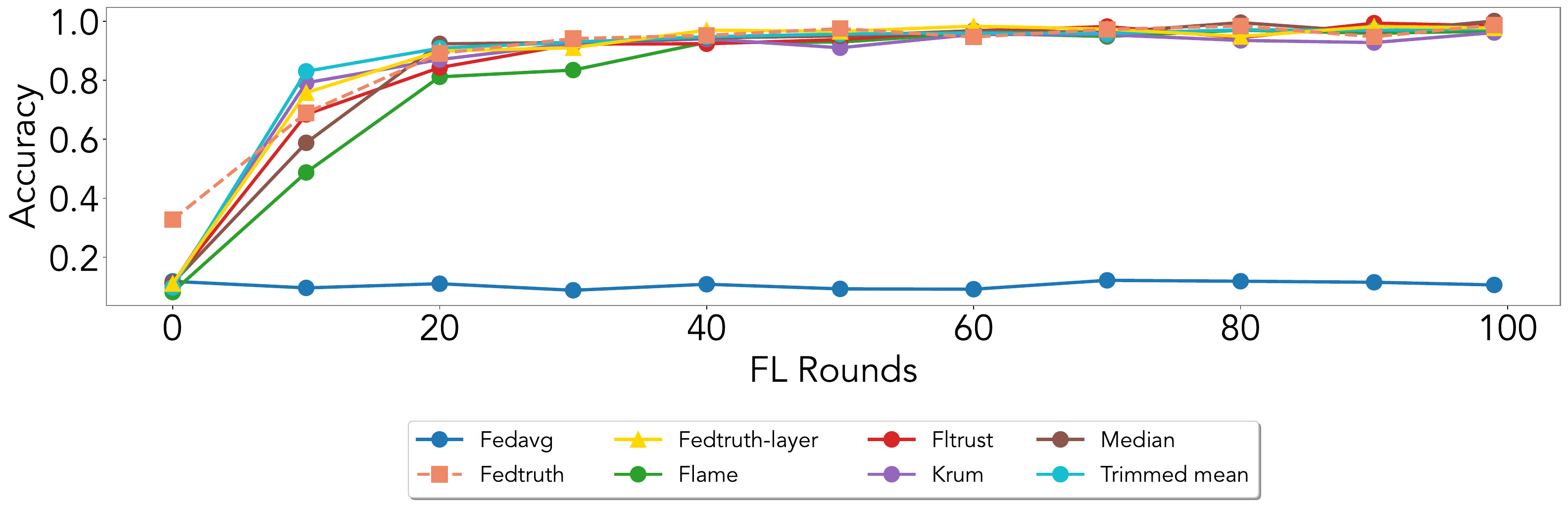}
    \end{subfigure}
    \begin{subfigure}{\textwidth}
        \centering
        \caption{Model Boosting Attack (MNIST, 3 Adversaries)}
        \label{fig:model-boosting-attacks-mnist-c}
        \includegraphics[clip,trim={0 3.9cm 0 0},width=0.9\textwidth]{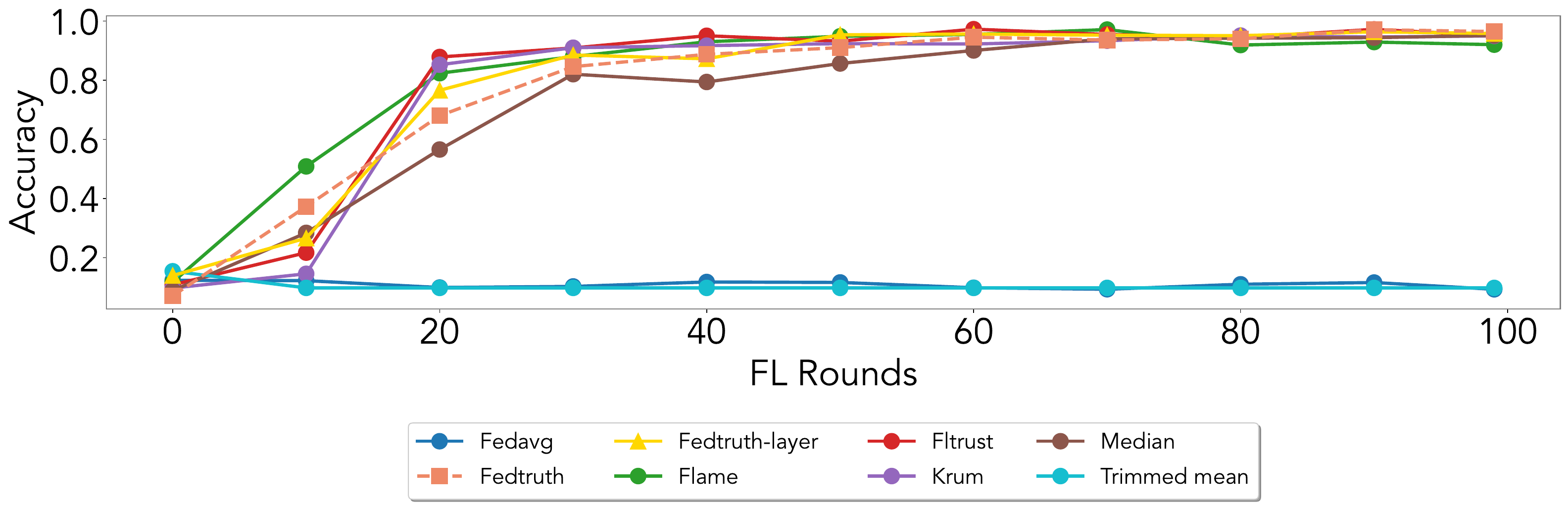}
    \end{subfigure}
    \begin{subfigure}{\textwidth}
        \centering
        \caption{Model Boosting Attack (MNIST, 4 Adversaries)}
        \label{fig:model-boosting-attacks-mnist-d}
        \includegraphics[clip,trim={0 0cm 0 0},width=0.9\textwidth]{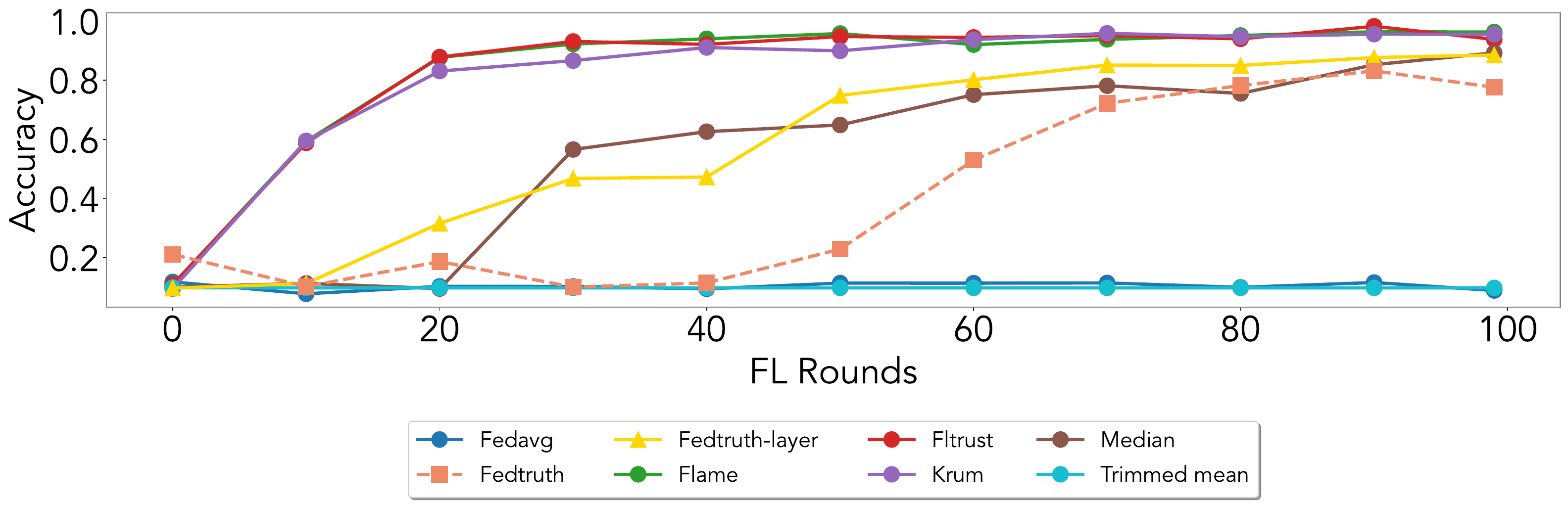}
    \end{subfigure}
    
    \captionsetup{font={normal}} 
    \caption{\textbf{Model Boosting Attack} (MNIST, $\times 10$ boosting factor)}
    \label{fig:model-boosting-attacks-mnist}
\end{figure*}

{Figure~\ref{fig:model-boosting-attacks-mnist-a}} shows how all of the aggregation algorithms perform when there are no adversaries present. Figure~\ref{fig:model-boosting-attacks-mnist-b} presents the results when \(10\%\) of the clients are compromised in each round. We observe that all of the aggregation algorithms performed well except for FedAvg. However, when the percentage of compromised clients increases to 30\%, as shown in Figure~\ref{fig:model-boosting-attacks-mnist-c}, the FedTruth methods remain unaffected by the attack. However, Trimmed-mean, similar to FedAvg, is significantly impacted at this stage and does not reach convergence.


The results in {Figure~\ref{fig:model-boosting-attacks-mnist-d}} show that increasing the number of adversaries per iteration to 40\% slows the convergence rate for the FedTruth, FedTruth-Layer, and Median aggregation algorithms. However, they are still able to reach an accuracy of 80\% after the 100th iteration. The FedAvg and Trimmed-mean algorithms were compromised during this version of the experiment as well, preventing them both from reaching any convergence when at least \(20\%\) of the clients are adversarial. In contrast, the remaining algorithms (FLTrust, Krum, FLAME) were not affected during this attack, regardless of the number of adversaries we selected.


\textbf{Gaussian Noise Attack}: 
The Gaussian noise attack aims to degrade the performance of the global model by adding arbitrary noise to the model. The noise is drawn from a multivariate Gaussian distribution $N(0,\sigma^{2}I)$~\cite{cao2021fltrust, blanchard2017machine} and is added directly to the adversaries' model before sending it to the aggregator.

\begin{figure}[!t]
    \captionsetup{font={tiny}} 
    \centering
    \begin{subfigure}{\textwidth}
        \centering
        \caption{Gaussian Noise Attack (base attack)}
        \label{fig:gaussian-attacks-mnist-a}
        \includegraphics[clip,trim={0 3.9cm 0 0}, width=0.9\textwidth]{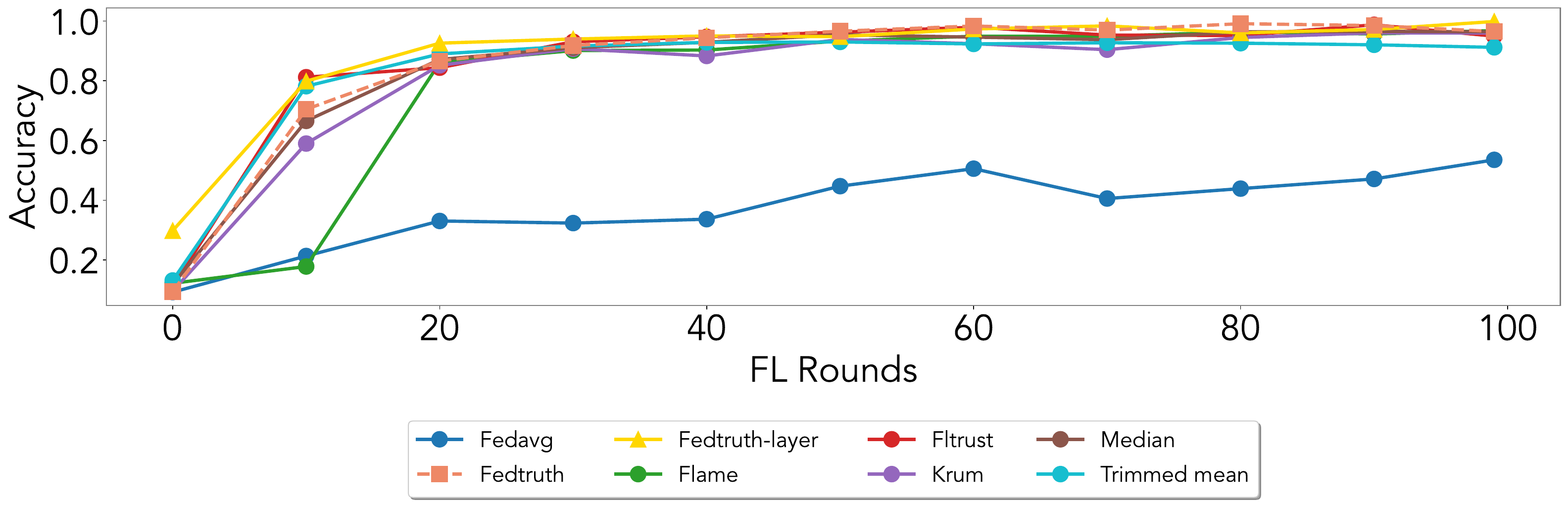}
    \end{subfigure}
    \begin{subfigure}{\textwidth}
        \centering
        \caption{Gaussian Noise Attack (model-boosting attack)}
        \label{fig:gaussian-attacks-mnist-b}
        \includegraphics[clip,trim={0 0 0 0},width=0.9\textwidth]{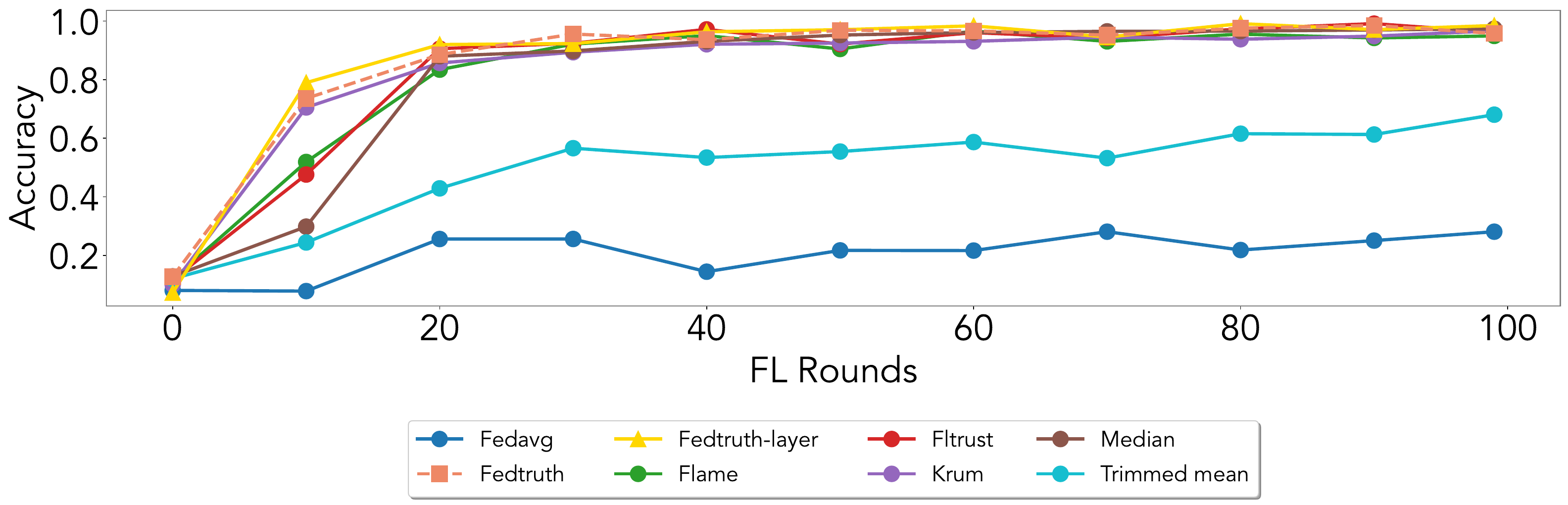}
    \end{subfigure}
    \captionsetup{font={normal}} 
    \caption{\textbf{Gaussian Noise Attack} (MNIST, 3 adversaries)}
    \label{fig:gaussian-attacks-mnist}
\end{figure}

In Figure~\ref{fig:gaussian-attacks-mnist-a}, we show the accuracy and convergence speed of the model for all the aggregation algorithms against the Gaussian noise attack, where three adversaries launch this attack per round. Our findings are as follows: 1) FedTruth and FedTruth-Layer can defend against the Gaussian noise attack without significantly slowing down the convergence speed; 2) FedTruth and FedTruth-Layer can achieve the same model accuracy as FLTrust, which requires a benign dataset, showing that our proposed algorithm can actually find the ground truth of the model updates; 
and 3) FedAvg cannot converge within 100 rounds under the Gaussian noise attack.

In Figure~\ref{fig:gaussian-attacks-mnist-b}, we combine the Gaussian noise attack with the model-boosting attack, which also does not degrade the performance of FedTruth or FedTruth-Layer. However, FedAvg and Trimmed-mean do not perform well against this attack.

\begin{figure}[!t]
    \centering
    \begin{subfigure}{\textwidth}
        \captionsetup{font={tiny}}
        \centering
        \caption{DBA Attack (base attack) - Main Task Accuracy}
        \label{fig:appx-dba-attacks-mnist-base-ma}
        \includegraphics[clip,trim={0 4cm 0 0}, width=0.9\textwidth]{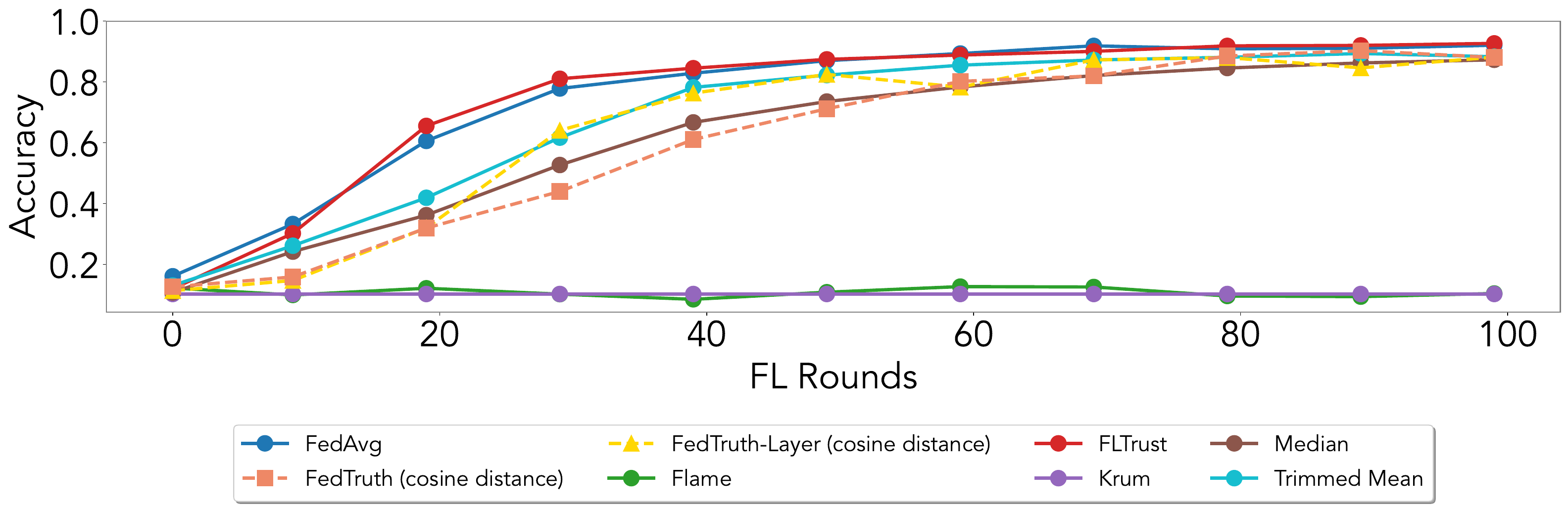}
    \end{subfigure}
    \hfill 
    \begin{subfigure}{\textwidth}
        \captionsetup{font={tiny}} 
        \centering
        \caption{DBA Attack (base attack) - Backdoor Accuracy}
        \label{fig:appx-dba-attacks-mnist-base-ba}
        \includegraphics[clip,trim={0 0 0 0}, width=0.9\textwidth]{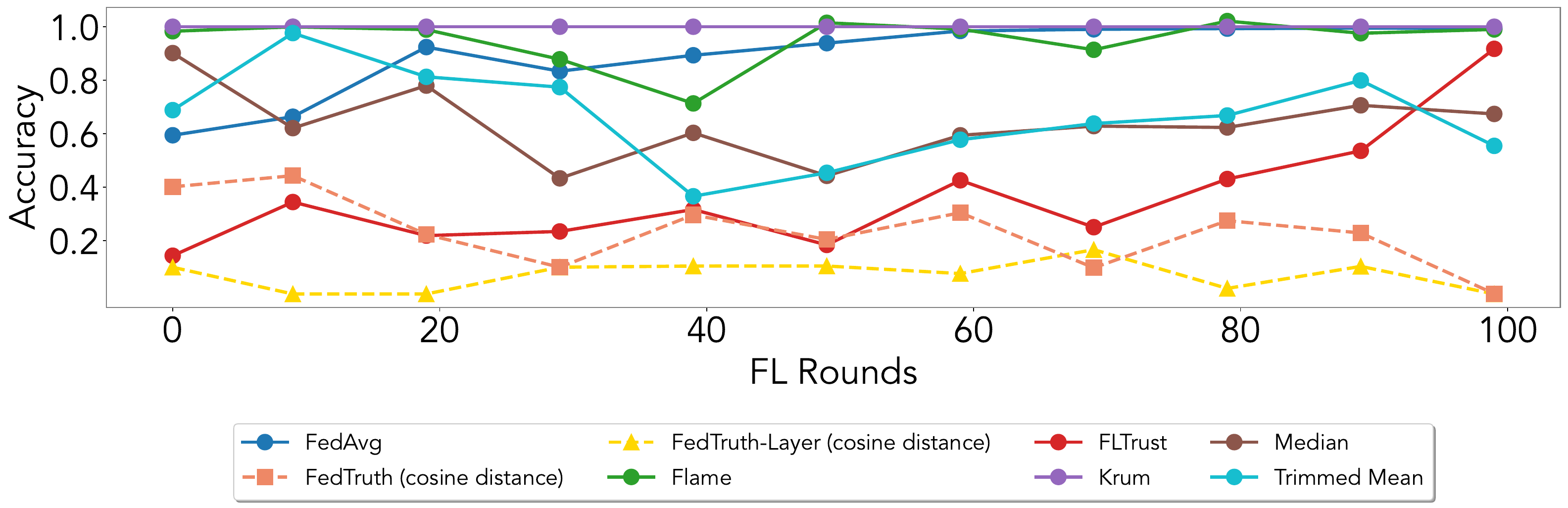}
    \end{subfigure}
  \caption{\normalsize \textbf{Distributed Backdoor Attack (base attack)} (MNIST, 3 adversaries)}
  \label{fig:appx-dba-attacks-mnist} 
\end{figure}

\subsection{Backdoor Attacks} \label{sec:backdoorDBA}

\begin{figure}[!t]
    \centering

    \begin{subfigure}{\textwidth}
        \captionsetup{font={tiny}}
        \centering
        \caption{DBA Attack (model-boosting attack) - Main Task Accuracy}
        \label{fig:appx-dba-attacks-mnist-mb-ma}
        \includegraphics[clip,trim={0 4cm 0 0}, width=0.9\textwidth]{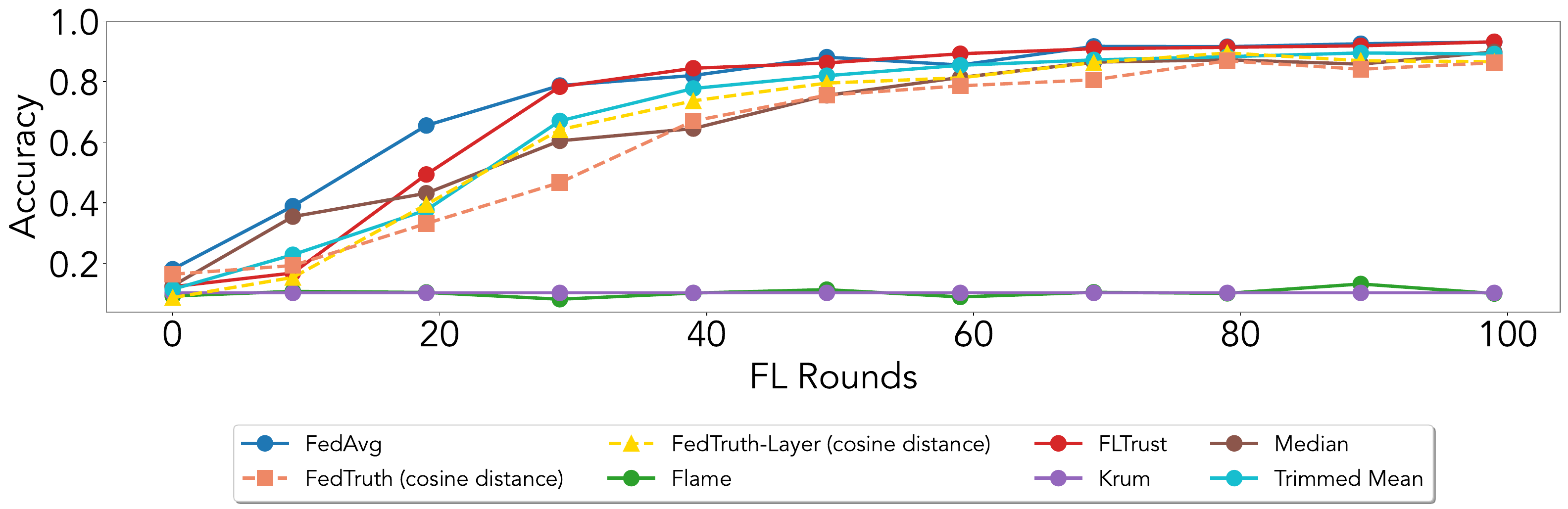}
    \end{subfigure}
    \hfill 
    \begin{subfigure}{\textwidth}
        \captionsetup{font={tiny}}
        \centering
        \caption{DBA Attack (model-boosting attack) - Backdoor Accuracy}
        \label{fig:appx-dba-attacks-mnist-mb-ba}
        \includegraphics[clip,trim={0 0 0 0}, width=0.9\textwidth]{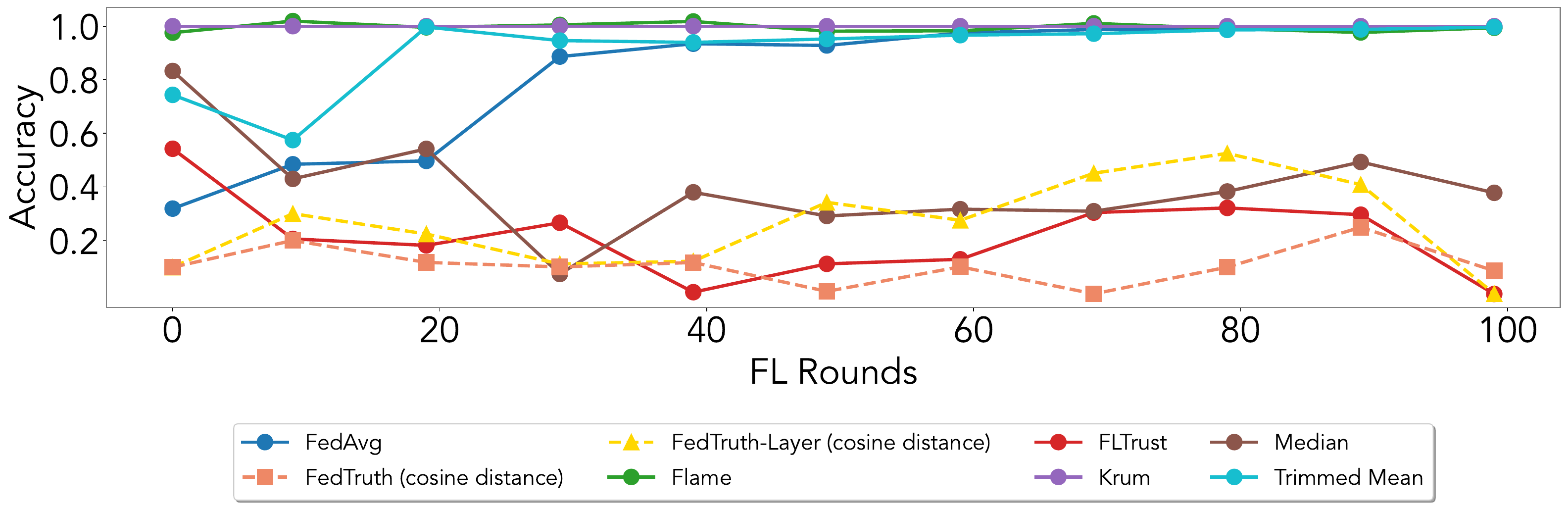}
    \end{subfigure}

  \caption{\normalsize \textbf{Distributed Backdoor Attack (combined with Model-Boosting Attack)} (MNIST, 3 adversaries)}
  \label{fig:appx-dba-attacks-mnist} 
\end{figure}

\begin{figure}[!t]
    \centering

    \begin{subfigure}{\textwidth}
        \captionsetup{font={tiny}}
        \centering
        \caption{DBA Attack (constrain-and-scale attack) - Main Task Accuracy}
        \label{fig:appx-dba-attacks-mnist-cs-ma}
        \includegraphics[clip,trim={0 4cm 0 0}, width=0.9\textwidth]{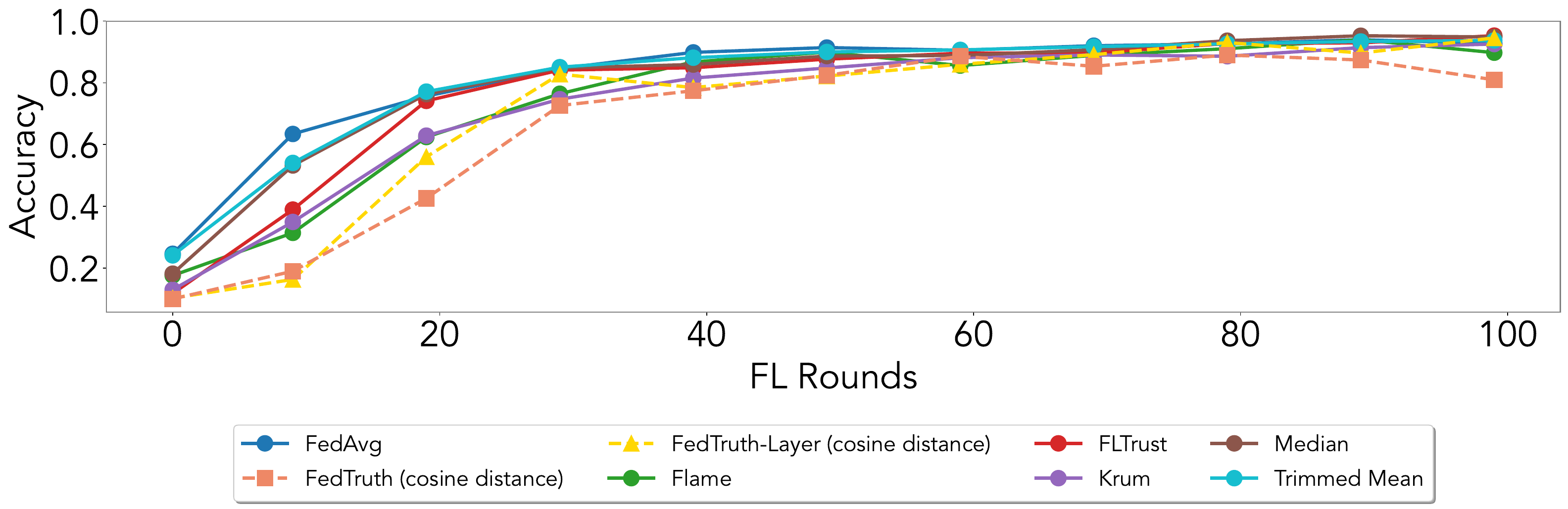}
    \end{subfigure}
    \hfill
    \begin{subfigure}{\textwidth}
        \captionsetup{font={tiny}}
        \centering
        \caption{DBA Attack (constrain-and-scale attack) - Backdoor Accuracy}
        \label{fig:appx-dba-attacks-mnist-cs-ba}
        \includegraphics[width=0.9\textwidth]{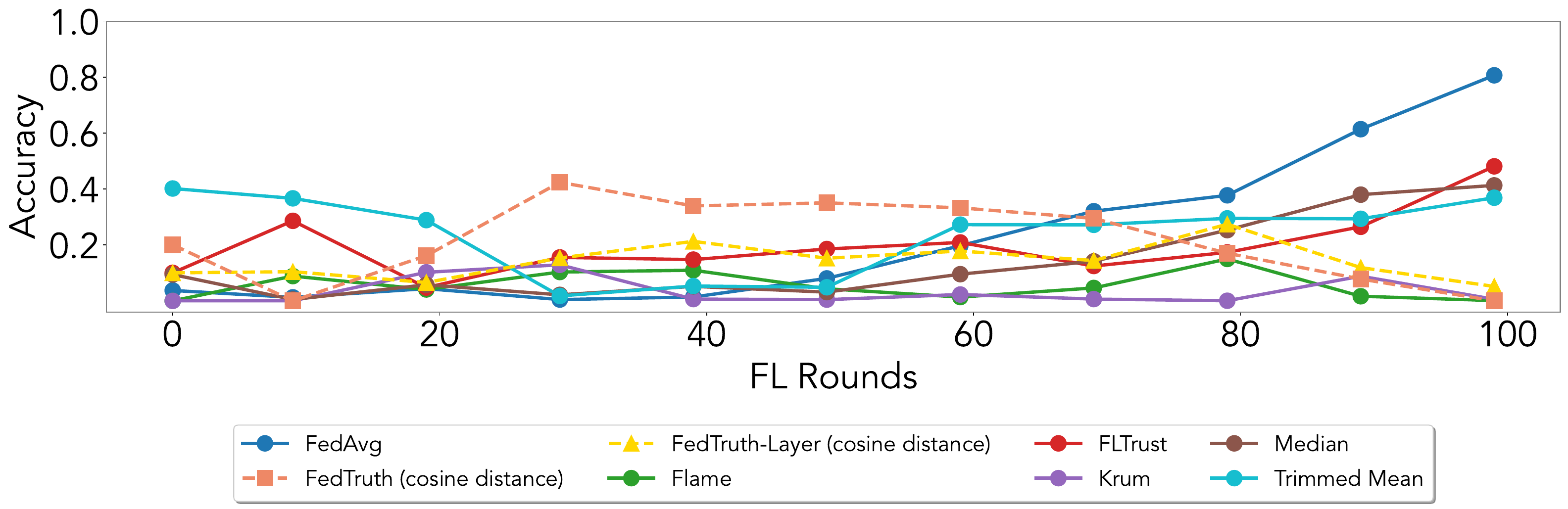}
    \end{subfigure}
  \caption{\normalsize \textbf{Distributed Backdoor Attack (combine with constraint-and-scalew)} (MNIST, 3 adversaries)}
  \label{fig:appx-dba-attacks-mnist} 
\end{figure}

In this section, we present our findings for \textit{target task accuracy} (accuracy on an adversarial backdoor dataset) and the \textit{main task accuracy} (accuracy on a benign dataset) for the distributed backdoor attack (DBA)  during the \textit{base},\textit{model-boosting}, and \textit{constrain-and-scale} attacks. 
During these attacks, we used the cosine distance metric, as it presented the best results during the majority of backdoor attack configurations. We provide results for the projected gradient descent and edge-case attacks with the same configurations in Appendix~\ref{app:more-backdoor-attacks} and analysis on the effects of different distance metrics in Section~\ref{sec:distance-functions} and Appendix~\ref{app:more-distance-functions}. 

Figures~\ref{fig:appx-dba-attacks-mnist-base-ma}, \ref{fig:appx-dba-attacks-mnist-mb-ma}, and \ref{fig:appx-dba-attacks-mnist-cs-ma} show the \textit{main task} accuracy for all of the versions of the DBA attack. Here, we see that both Krum and Flame are unable to train the main task during the \textit{base} attack and when combined with the \textit{model-boosting}. We suspect these results could be caused by the large number of adversaries colluding during this experiment or the imbalanced sample data. The remaining algorithms are able to reach convergence on the \textit{main task} during these attacks. When the DBA is combined with the \textit{constrain-and-scale} attack, all algorithms are able to reach convergence, including Flame and Krum, which we suspect is due to the \textit{constrain-and-scale} attack reducing the amount that an adversarial model can diverge during each epoch. However, FedTruth and FedTruth-Layer experience a slower convergence rate during the DBA with \textit{model-boosting} and \textit{constrain-and-scale} attacks.

Figures~\ref{fig:appx-dba-attacks-mnist-base-ba}, \ref{fig:appx-dba-attacks-mnist-mb-ba}, and \ref{fig:appx-dba-attacks-mnist-cs-ba} present our findings on the \textit{targeted task accuracy} (i.e., the backdoor accuracy) during the DBA experiments. The adversarial goal during this attack is to add the targeted artifact to the global model without being detected and without affecting the model's performance on the main task. Therefore, we will not be considering the Krum and FLAME algorithms during the \textit{base} and \textit{model-boosting} attacks, as they were unable to converge when training the main task, as seen in {Figure~\ref{fig:appx-dba-attacks-mnist-base-ma} and \ref{fig:appx-dba-attacks-mnist-mb-ba}}. 

During the base attack ({Figure~\ref{fig:appx-dba-attacks-mnist-base-ba}}), FedTruth and FedTruth-Layer are the only aggregation algorithms that are able to remove the backdoor, finishing with a backdoor accuracy below 5\%. 
Furthermore, during the DBA with \textit{model-boosting} attack, we observe similar results with FedTruth and FedTruth-Layer being able to remove the backdoor and reach a final backdoor accuracy below 5\%. However, FLTrust is now able to reach a similar convergence rate, and the Median algorithm improves as well, with a final accuracy below 40\%. 
This is a result of a tradeoff between the stealthiness of the base DBA attack being diminished when increasing the amplitude of the adversarial updates vector in hopes of replacing the global model with the adversarial model. 
Our results also show that the DBA with \textit{constrain-and-scale} attack is effective at adding the targeted task into FedAvg, FLTrust, Median, and Trimmed mean, reaching a backdoor accuracy above 40\% after 100 iterations. However, FedTruth, FedTruth-Layer, Krum, and FLAME are able to remove the adversarial artifact finishing with a backdoor accuracy below 5\%.

\subsection{The Impact of non-iid on FedTruth}\label{Sec:noniid}

To perform our non-iid experiments, we used label skew, where each of the clients had an equal number of data points. However, each client has a primary label from which the majority of their data points will come. By changing how many data points come from a client's primary, we are able to change the degree to which their data is non-iid.


\begin{figure*}[!t]
    \captionsetup{font={tiny}} 
    \centering
    \begin{subfigure}{\textwidth}
        \centering
        \caption{noniid = 10\%}
        \label{fig:model-boosting-noniid-mnist-a}
        \includegraphics[clip,trim={0 4cm 0 0}, width=0.9\textwidth]{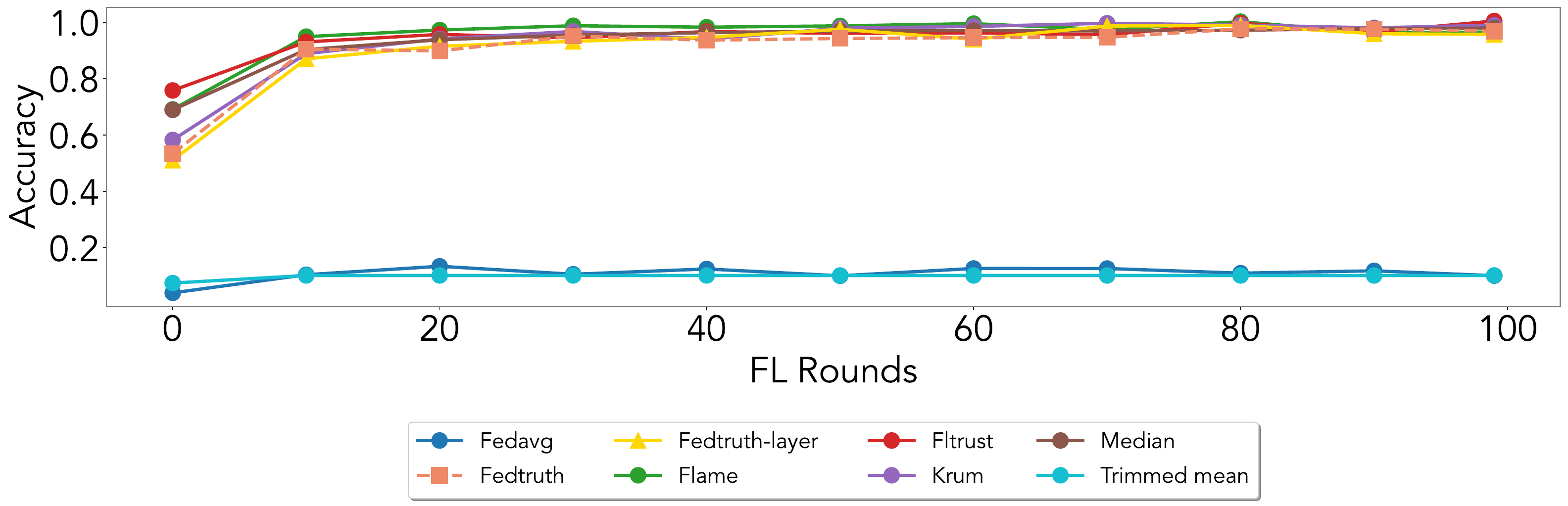}
    \end{subfigure}
    \begin{subfigure}{\textwidth}
        \centering
        \caption{noniid = 50\%}
        \label{fig:model-boosting-noniid-mnist-b}
        \includegraphics[clip,trim={0 4cm 0 0},width=0.9\textwidth]{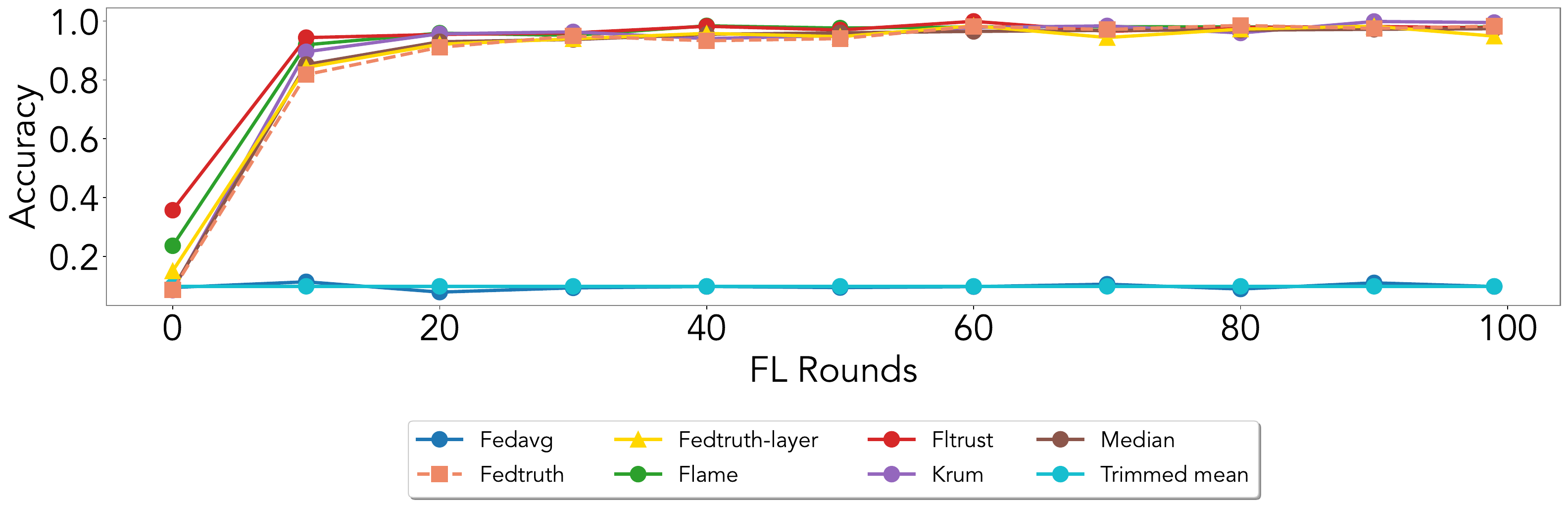}
    \end{subfigure}
    \begin{subfigure}{\textwidth}
        \centering
        \caption{noniid = 80\%}
        \label{fig:model-boosting-noniid-mnist-c}
        \includegraphics[clip,trim={0 4cm 0 0},width=0.9\textwidth]{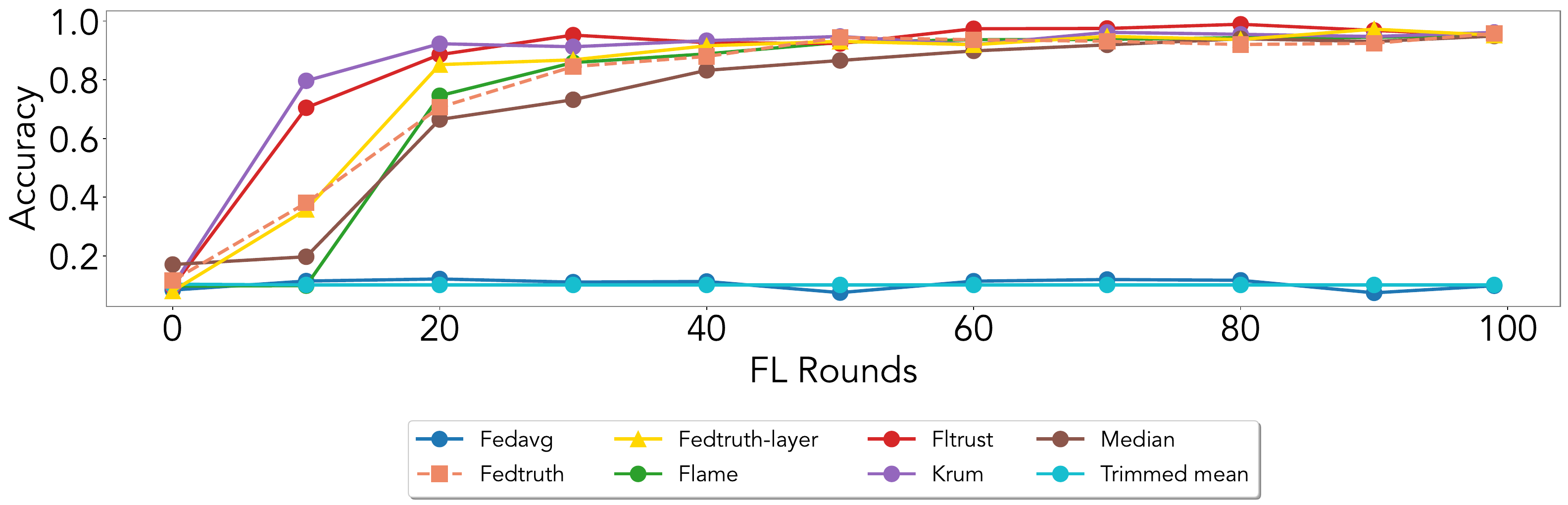}
    \end{subfigure}
    \begin{subfigure}{\textwidth}
        \centering
        \caption{noniid = 95\%}
        \label{fig:model-boosting-noniid-mnist-d}
        \includegraphics[clip,trim={0 0 0 0},width=0.9\textwidth]{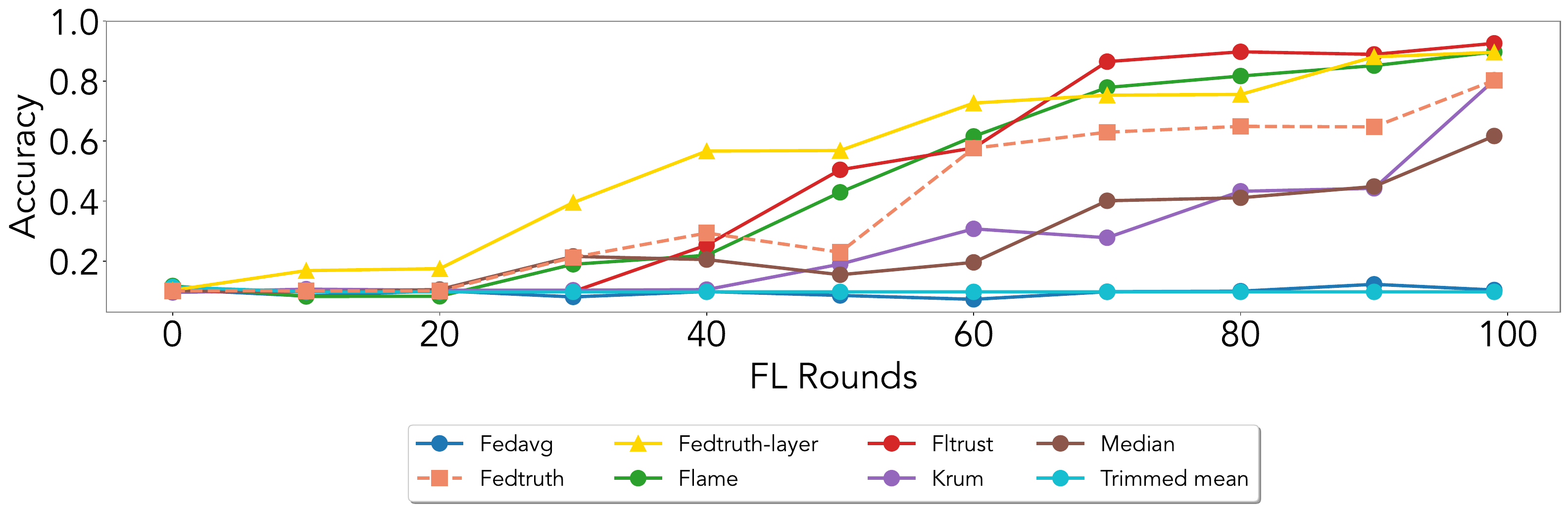}
    \end{subfigure}
    \captionsetup{font={normal}} 
    \caption{\textbf{Non-iid Impact on Model Boosting Attack} (MNIST, 3 adversaries, $\times 10$  boosting factor)}
    \label{fig:model-boosting-noniid}
\end{figure*}

Figure~\ref{fig:model-boosting-noniid} shows how various non-iid bias parameters affect the experiments when adversaries apply the \textit{model-boosting} attack.  During these experiments, three adversaries were selected in each round. 
Our methodology for sampling non-iid data was specifically engineered to replicate varying degrees of label bias, thus enabling an in-depth analysis of its influence on federated learning model efficacy. We manipulated a bias parameter to adjust the label proportions within each client's local dataset. For instance, setting the bias parameter to 0.9 indicated that 90\% of a client's dataset contained instances of their primary label, with the remaining 10\% consisting of instances from other labels, allocated based on a Gaussian distribution.
For this experiment, we set the bias parameters as $0.1$, $0.5$, $0.8$, and $0.95$.

The results of the experiments, as seen in {Figure~\ref{fig:model-boosting-noniid}}, suggest that FedTruth can mitigate the impacts of the boosted model regardless of the non-iid degree of the datasets. The FedTruth and FedTruth-Layer algorithms do experience some performance degradation as the non-iid degree increases, which is to be expected. However, as seen in {Figure~\ref{fig:model-boosting-noniid}}, after 100 FL iterations, both algorithms reach a top accuracy regardless of the non-iid bias degree.

\subsection{Distance Function in FedTruth}\label{sec:distance-functions}

\begin{figure}[!h]
    \centering
    \begin{subfigure}{\textwidth}
        \centering
        \caption{Model Boosting Attack}
        \label{fig:distance-functions-a}
        \includegraphics[clip,trim={0 5cm 0 0}, width=0.9\textwidth]{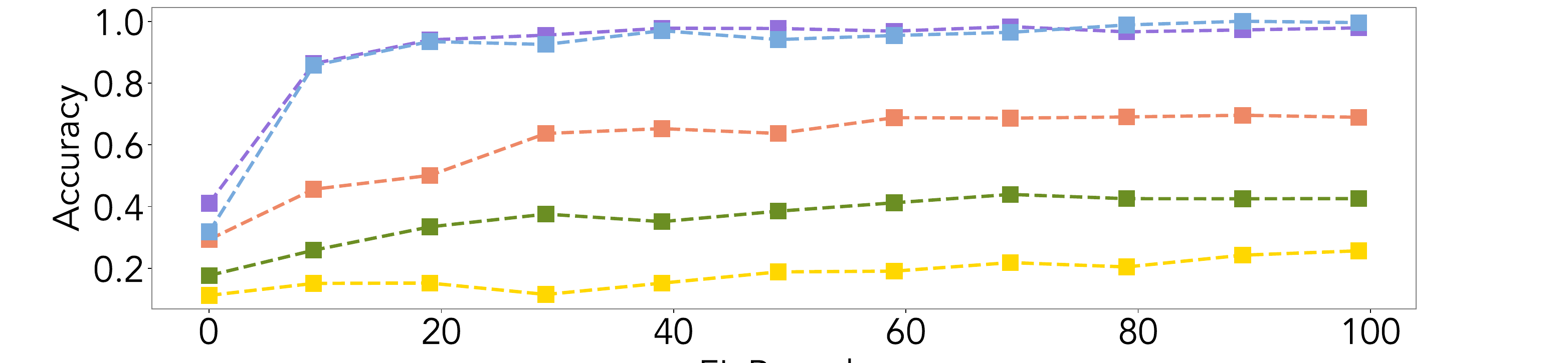}
    \end{subfigure}
    \begin{subfigure}{\textwidth}
        \centering
        \caption{Gaussian Noise Attack}
        \label{fig:distance-functions-b}
        \includegraphics[clip,trim={0 5cm 0 0},width=0.9\textwidth]{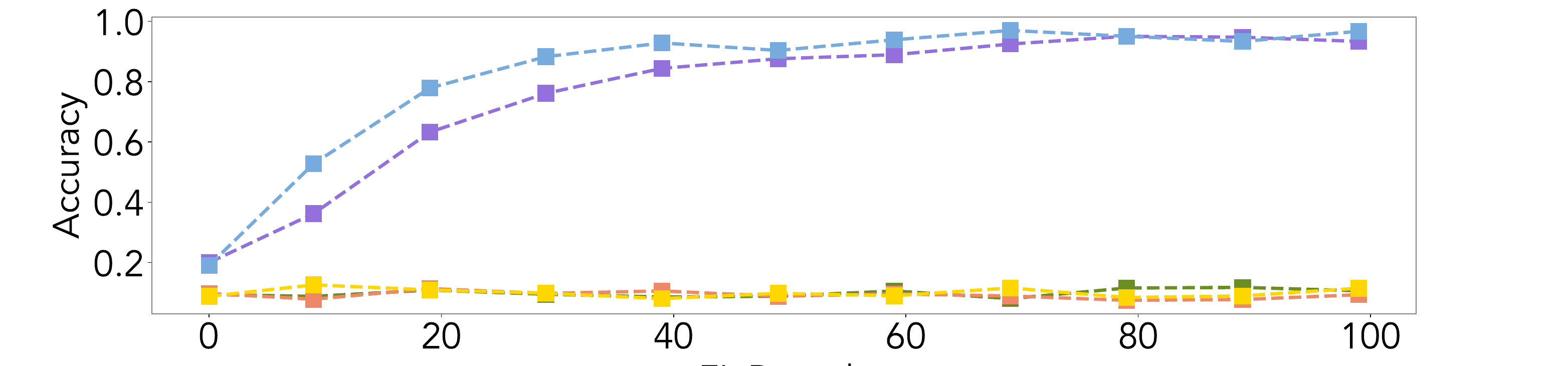}
    \label{fig:byzantine-attacks-mnist-distance}
    \end{subfigure}

    \begin{subfigure}{\textwidth}
        \centering
        \caption{DBA Main Accuracy}
        \label{fig:distance-functions-c}
        \includegraphics[clip,trim={0 5cm 0 0}, width=0.9\textwidth]{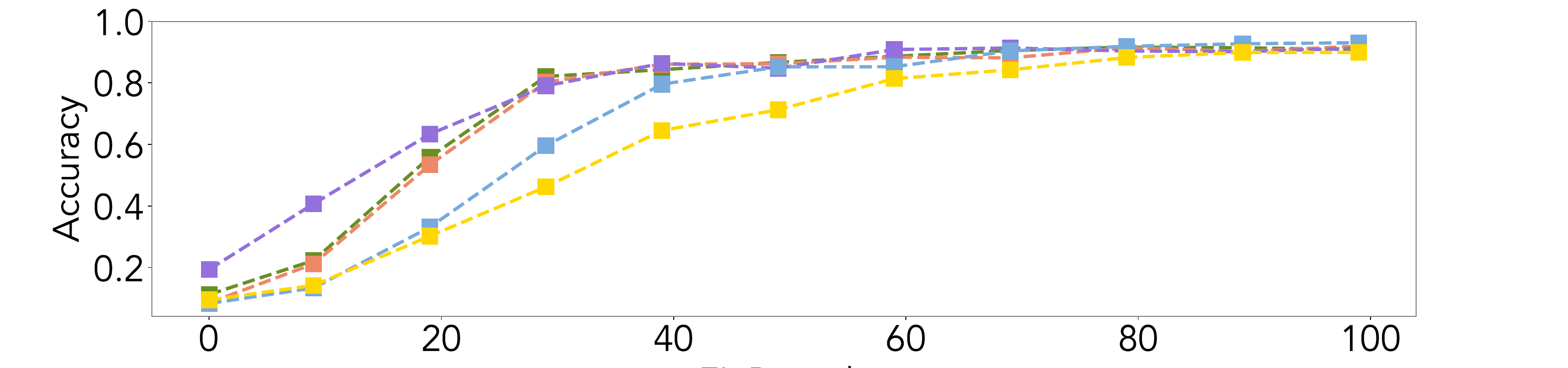}
    \end{subfigure}
    \begin{subfigure}{\textwidth}
        \centering
        \caption{DBA Backdoor Accuracy}
        \label{fig:distance-functions-d}
        \includegraphics[clip,trim={0 0 0 0},width=0.9\textwidth]{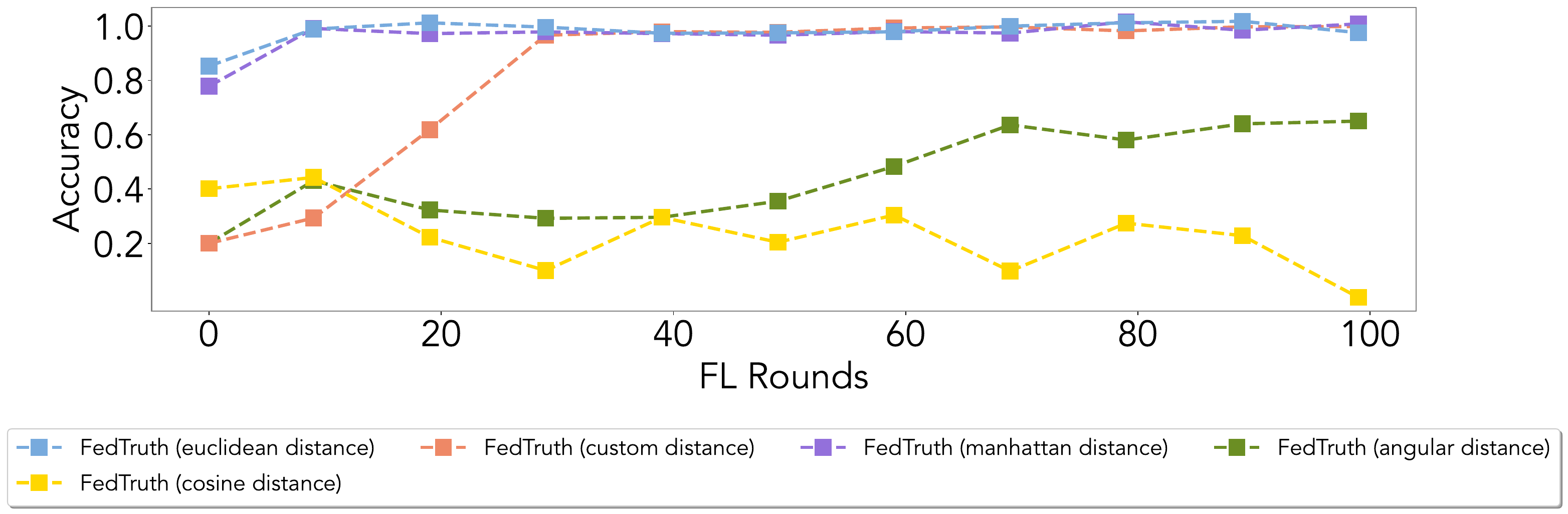}
    \end{subfigure}
    \captionsetup{font={normal}} 
    \caption{\textbf{Comparison of Distance Functions} (MNIST, 3 adversaries)}
    \label{fig:distance-functions}
\end{figure}


From the FedTruth formulation (i.e., Equ. \ref{equ:fedtruth}), we can see that the distance function plays a significant role in separating benign and malicious model updates. This section discusses how FedTruth performs with different distance functions. 
More results on different distance metrics of FedTruth will be shown in the Appendix~\ref{app:more-distance-functions}. 


We evaluate the performance of FedTruth against both Byzantine and backdoor attacks using the following 
distance functions: 
1) two metrics that compute the \textit{difference between the angles} of two vectors (angular distance = arccos(cosine similarity)/${\pi}$ and cosine distance = 1 - cosine similarity);
2) two metrics that determine the \textit{difference between two points} (Euclidean and Manhattan distances); and
3) one custom distance that \textit{combines the angular distance and the Euclidean distance}, which we combine half and half in our results. 


Figures~\ref{fig:distance-functions-a} and \ref{fig:distance-functions-b} show how the choice of distance metric affects FedTruth during Byzantine attacks. As expected, the two metrics that solely measure the difference between two points (Euclidean and Manhattan distances) performed the best, reaching convergence in both cases with a final accuracy of 100\%. This is because these approaches can easily identify the adversarial model furthest from the benign models, as these are either boosted or have additional random noise inserted into them. During the Model Boosting attack (Figure~\ref{fig:distance-functions-a}), we see that the custom distance metric was able to reach sub-optimal performance, finishing with an accuracy of 60\%. However, since the angular distance and cosine distance metrics do not consider a model's magnitude, they are ineffective during the model boosting attack (Figure~\ref{fig:distance-functions-a}). During the Gaussian noise attack (Figure~\ref{fig:distance-functions-b}), the angular, cosine, and custom distances were also not effective. This is due to the slight modification of the angle during the Gaussian noise attack, causing the adversarial updates to be less effective. This is also why the custom distance, which partially relies on the Euclidean distance, is not sufficient for removing the adversarial updates.

Figures~\ref{fig:distance-functions-c} and~\ref{fig:distance-functions-d} present our results for all distance metrics during a backdoor (DBA) attack. These results indicate that the cosine distance metrics perform the best, as it is able to remove any backdoors injected into the model. Interestingly, while the angular distance finished with a final backdoor accuracy of 60\%, while the cosine distance metric was able to reduce the backdoor accuracy to below 5\%. The models focused on point-to-point distance measurement (Euclidean, Manhattan, and custom distance) did not perform well, with all of their final backdoor accuracies reaching 100\%.



\subsection{Efficiency Evaluation of FedTruth and FedTruth-Layer}\label{sec:efficiency}
From our experimental results, we observe that both FedTruth and FedTruth-Layer perform similarly in terms of model accuracy and robustness. To evaluate their efficiency, we present the average time consumption for each aggregation algorithm in {Table~\ref{table:time_comparison}}. We measure each client's average aggregation time and average training time based on training a CNN model on MNIST and CIFAR10 for 100 rounds with three adversaries in each round.

\begin{table}[!h]
\vspace{-20pt}
  \centering
  \caption{Aggregation and training time for different FL algorithms (Model Boosting Attack, 3 Adversaries, 10 clients/round)}
\resizebox{\columnwidth}{!}{
\begin{tabular}{l||c|c|c|c}
\hline\hline
\multirow{3}{*}{\textbf{Algorithm}} &
  \multicolumn{2}{c|}{\textbf{Average Aggregation Time (s)}} &
  \multicolumn{2}{c}{\textbf{Average Training Time (s)}} \\ 
  \cline{2-5} 
  & \textit{MNIST} & \textit{CIFAR-10}& \textit{MNIST} & \textit{CIFAR-10} \\ \hline
  FedAvg         & 0.0034
& 0.0175
& 0.0767 & 0.8528 \\ \hline
  FedTruth       & 0.1054
& 0.1476
& 0.0773 & 0.8617 \\ \hline
  FedTruth-Layer & 0.7299
& 3.2935
& 0.0778 & 0.8528  \\ \hline
  Flame          & 0.1573
& 1.7956
& 0.0800 & 0.8508 \\ \hline
  FLTrust        & 0.0192
& 0.0947
& 0.0808 & 0.8555 \\ \hline
  Krum           & 0.0749
& 0.4000
& 0.0782 & 0.8579  \\ \hline
  Median         & 0.0008
& 0.0022
& 0.0804 & 0.8578 \\ \hline
  Trimmed mean   & 0.0020& 0.0053& 0.0792 & 0.8580 \\\hline
\end{tabular} \label{table:time_comparison}
}\vspace{-15pt}
\end{table}

We find that during the MNIST experiments, the Median algorithm is the most efficient, while Trimmed mean, FedAvg, and FLTrust take less than 0.02 seconds. FLAME takes about 0.16 seconds to filter and clip, while Krum takes around 0.08 seconds. Interestingly, FedTruth has a slightly faster aggregation time than FLAME, and FedTruth-Layer is the slowest aggregation algorithm with an average aggregation time of 0.7 seconds. 

We observed similar results when we ran the experiment using the ResNet-18 model on the CIFAR-10 dataset, with FedTruth-Layer still being the slowest algorithm. However, we also notice significant increases in aggregation time for FedTruth-Layer, Krum, and FLAME, with a marginal increase in FedTruth's aggregation time. The reason is that FedTruth-Layer has a high number of total iterations, as shown in  Table~\ref{table:fedtruth_comparison}.


\begin{table}[!h]
\vspace{-25pt}
  \centering
\caption{Number of Iterations for FedTruth and FedTruth-Layer (Model Boosting Attack, 3 Adversaries, 10 clients)}
\resizebox{\columnwidth}{!}{
  \begin{tabular}{c||c|c}
      \hline\hline
    \multirow{3}{*}{\textbf{Algorithm}}
     & \multicolumn{2}{c}{\textbf{Average Number of Iterations until FedTruth Convergence}}\\  
     \cline{2-3}
     & \textit{MNIST} & \textit{CIFAR-10} \\ \hline
    FedTruth & 5.17& 4.71
\\ \hline
    FedTruth-Layer Total & 33.73
& 148.54
\\   \hline
    FedTruth-Layer L1 & 4.13
& 4.3
\\ \hline
    FedTruth-Layer L2 & 3.62
& 3.48
\\\hline
    FedTruth-Layer L3 & 4.34
& 3.38
\\\hline
    FedTruth-Layer L4 & 3.71
& 4.22
\\\hline
    FedTruth-Layer L5 & 4.83
& 3.32
\\\hline
    FedTruth-Layer L6 & 4.04
& 3.38
\\\hline
    FedTruth-Layer L7 & 5.06
& 4.31
\\\hline
    FedTruth-Layer L8 & 4.6& 3.3\\\hline
    \hline
  \end{tabular}
  }
  \vspace{-15pt}
  \label{table:fedtruth_comparison}
\end{table}
We count the number of iterations required for both algorithms to reach convergence and present the results in Table~\ref{table:fedtruth_comparison}. For the CNN (8 layers) model on the MNIST dataset, we find that FedTruth requires an average of 5.17 iterations to estimate the ground-truth model update, while FedTruth-Layer requires an average of 33.73 iterations (six times more than FedTruth) to reach convergence, despite having eight layers in the CNN model. Moreover, FedTruth on each layer has a smaller input size and requires less computation time on the distance compared to FedTruth with the entire model update as input. In conclusion, we find that both FedTruth and FedTruth-Layer have similar performance on the MNIST CNN (8 layers) model regarding accuracy and robustness. 
However, FedTruth is much more efficient than FedTruth-Layer when the number of layers increases, which can be observed in the CIFAR-10 (ResNet-18) model.

We further evaluate the average aggregation time for 10, 100, and 1000 clients in a single FL round in Table~\ref{tab:aggtime}, where the aggregation time is calculated as the average of 100 FL rounds. We can see that FedTruth and FedTruth-Layer are as efficient as FLTrust (which requires a benign dataset) and much more efficient than FLAME (which does not require a benign dataset). FLAME becomes very slow when there are 1000 clients in each round.

\begin{table}[!h]
\vspace{-20pt}
    \centering
    \caption{Comparison of Average Aggregation Time (Model Boosting Attack, 3 Adversaries, MNIST)}
    \resizebox{\columnwidth}{!}{
    \begin{tabular}{c||c|c|c|c|c|c|c|c} \hline \hline
       \textbf{Clients}  & \multicolumn{8}{c}{\textbf{Average Aggregation Time (s)}} \\ \cline{2-9}
       \textbf{per}& \multirow{2}{*}{\textbf{FedAvg}} & \multirow{2}{*}{\textbf{FedTruth}} & \textbf{FedTruth-} & \multirow{2}{*}{\textbf{FLAME}} & \multirow{2}{*}{\textbf{FLTrust}} & \multirow{2}{*}{\textbf{Krum}} & \multirow{2}{*}{\textbf{Median}} & \textbf{Trimmed} \\
      \textbf{round} & & & \textbf{Layer}&&&&& \textbf{Mean} \\ \hline
       10 & 0.006 & 0.107& 0.777& 0.194 & 0.033 & 4.06 & 0.041 & 0.027 \\ \hline
       100  & 0.047 &  0.518& 3.498& 8.813 & 0.313 & 824.803 & 0.173 & 0.414\\ \hline 
       1000  & 1.554 & 4.903& 29.88& \textbf{824.549} & 3.468 & 83172.343 & 2.849 & 6.547\\ \hline \hline
    \end{tabular}
    }
    \label{tab:aggtime}
\end{table}


\section{Related Work}\label{sec:relatedwork}

Defending against model poisoning attacks in federated learning has been an area of active research, with many efforts focusing on designing robust aggregation rules.
One approach to identifying and removing malicious model updates involves clustering methods (e.g., Krum \cite{blanchard2017machine}, AFA \cite{munoz2019byzantine}, FoolsGold \cite{fung2020limitations}, and Auror \cite{shen2016auror}). Although effective, these methods rely on specific assumptions about the underlying data distribution among clients. For example, Krum and Auror assume that benign clients' data are independent and identically distributed (iid), whereas FoolsGold and AFA assume non-iid benign data. Additionally, these defenses may be ineffective against stealthy attacks, such as constraint-and-scale attacks \cite{bagdasaryan2020backdoor}, or adaptive attacks, such as the Krum attack \cite{fang2020local}.

Another approach aims to reduce the impact of poisoned model updates on the global model by clipping individual weights to a certain threshold and adding random noise \cite{bagdasaryan2020backdoor, nguyen2021flame}. For instance, FLAME \cite{nguyen2021flame} combines clustering with adaptive clipping and noising to mitigate poisoning attacks. However, this technique may unintentionally suppress contributions from benign clients, particularly those with underrepresented datasets.

Other methods find the mean or median of model update weights by excluding values based on thresholds (e.g., trimmed mean or median \cite{yin2018byzantine}) or frequency of occurrence (FreqFed \cite{fereidooni2024freqfed}). Despite their robustness, these approaches are vulnerable to adaptive attacks, such as the Trim attack \cite{fang2020local}, which exploit the limitations of these methods.

Some defenses adjust aggregation weights based on the distance between model updates and a benign root dataset \cite{cao2021fltrust}.  {FLTrust} assumes that there is a benign root dataset available to the aggregation server, who will also train and output a server model in each FL round. Upon receiving all the local model updates from clients, the server calculates a \textit{Trust Score} using the ReLU-clipped cosine similarity between each local model update and the server model update. The global model update is computed as the average of the normalized local model updates weighted by the \textit{trust scores}.


In \cite{tahmasebian2022robustfed}, the authors proposed \textit{RobustFed} that applies the truth discovery approach to estimate the reliability of clients in each round. Then, the estimated reliability is used to compute the next round aggregated model. This method suffers from the following two drawbacks. 
1) RobustFed applies truth discovery to calculate the reliability $r_{c_i}^t$ of each client $c_i$ in round $t$, and uses it to aggregate the global model for round $t+1$ (see Eq.11 in RobustFed, $w^{t+1}_G = w^t_G + \sum_{i\in K} r^t_{c_i} \cdot \alpha_i \cdot \delta^{t+1}$). In this case, an attacker can behave honestly to obtain a high reliability score in round $t$, and launch the Byzantine attack in the next round $t+1$;
 and 2) Even revising the method to calculate the reliability in the same round, the reliability cannot be directly added to the FedAvg in RobustFed. The reliability is defined by a negative logarithm function of the difference between its local model updates and the truths (ranges between 0 and 1). So, the reliability is a real number ranging between 0 and $+\infty$. The global model aggregation in RobustFed directly adds the reliability on top of the FedAvg, potentially magnifying the local model updates if the reliability is a large number.

TDFL~\cite{xu2022tdfl} also relies on Truth Discovery to aggregate the global model but, it mainly focuses on applying clustering and clipping filters as shown in FLAME \cite{nguyen2021flame} before the truth discovery procedure to defend against Byzantine attacks. However, akin to RobustFed, it simply uses the negative exponential regulation function as detailed in the CRH truth discovery \cite{li2014resolving}. 

Recently, several works \cite{kusetogullari2020ardis, gorbunov2023variance, farhadkhani2022byzantine} have been proposed to achieve provable Byzantine robustness by integrating variance-reduced algorithms and byzantine-resilient aggregation algorithms. However, they require prior knowledge of the
variance of the gradients \cite{kusetogullari2020ardis, gorbunov2023variance} or only focus on existing byzantine-resilient aggregation algorithms. 
In this paper, we propose a generic and robust model aggregation algorithm by computing the aggregation weight dynamically, which is also effective in defending against backdoor attacks, such as DBA \cite{xie2019dba} and PGD \cite{wang2020attack}.

\section{Conclusion}\label{Sec:conclusion}
In this paper, we developed FedTruth and FedTruth-Layer, a generic solution to defend against model poisoning attacks in FL. Compared with existing solutions, FedTruth eliminates the assumptions of benign or malicious data distribution and the need to access a benign root dataset. Specifically, a new approach was proposed to estimate the \textit{ground-truth model update} (i.e., the global model update) among all the model updates with dynamic aggregation weights in each round, following the principle that higher weights will be assigned to more reliable clients. The experimental results show that FedTruth and FedTruth-Layer can efficiently reduce poisoned model updates' impacts against Byzantine and backdoor attacks. Moreover, FedTruth works well on both iid and non-iid datasets.


\bibliographystyle{splncs04}
\bibliography{Reference}


\newpage
\appendix

\section{Details of Model Poisoning Attacks}
A malicious client or an adversary who compromises a set of clients can influence the global model by changing local datasets (data poisoning attack, e.g., label-flipping attack \cite{tolpegin2020data}) or directly manipulating local model updates (model poisoning attack \cite{blanchard2017machine, chen2017distributed, bhagoji2019analyzing, bagdasaryan2020backdoor, xie2019dba}). 
Specifically, the adversary can change both \textit{direction (angle)} and \textit{magnitude (length)} of the model updates to launch model poisoning attacks, including: 

\textbf{\textit{Byzantine attacks}}: The goal of the Byzantine attack is to make the global model converge to a sub-optimal model \cite{blanchard2017machine, chen2017distributed}. Some Byzantine attacks are: 
\begin{itemize}
    \item \textit{Model-boosting Attack}: Basic aggregation algorithms like FedAvg can remove the artifact during each iteration, making it challenging to impact the final global model. Similar to the amplification attack, the model-boosting attack~\cite{bhagoji2019analyzing} refers to explicitly boost the local model updates ($\Delta_t^{(k)} = w_t - w_t^{(k)}$) rather than the local models ($w_t^{(k)}$).

    \item \textit{Gaussian Noise Attack}: In~\cite{blanchard2017machine}, Byzantine clients randomly draw the local model from a Gaussian Distribution, which is referred to as a Gaussian Byzantine attack. When only local model updates ($\Delta_t^{(k)} = w_t - w_t^{(k)}$) are communicated, Gaussian Byzantine attack aims to add noise to the adversaries' local model. The adversarial noise used to degrade the model performance is drawn from a Gaussian distribution.

    \item \textit{Constraint-and-scaling Attack}. Simply boosting the model can be easily detected by anomaly detection algorithms. The \textit{constraint-and-scaling attack} \cite{bagdasaryan2020backdoor} does the model boosting attack while taking the anomaly detection constraints into the crafting of the adversarial model. 
\end{itemize}

\textbf{\textit{Backdoor attacks}}: A backdoor attack aims to manipulate local model updates to cause the final model to misclassify certain inputs with high confidence \cite{bhagoji2019analyzing, bagdasaryan2020backdoor, xie2019dba, wang2020attack}. Some backdoor attacks are: 
\begin{itemize}
    \item  \textit{Distributed Backdoor Attack (\textit{DBA})}~\cite{xie2019dba}: The \textit{DBA} attack compromises the global model by cropping the adversarial data into multiple segments based on the number of adversaries colluding during a given FL iteration. Therefore, when the server aggregates the selected models, the backdoor artifact is inserted into the model. 

    \item \textit{Edge Case Attack}~\cite{wang2020attack}:
The \textit{edge-case attacks} takes advantage of a systematic weakness that ML models face when a subset of labeled data is drawn from a minority subset of training data. 

    \item \textit{Projected Gradient Descent Attack (\textit{PGD})}~\cite{wang2020attack}: 
    In PGD attacks, adversaries periodically project their local models on a small ball, centered around the global model of the previous round. 
\end{itemize}
These backdoor attacks may also be combined with the  \textit{model-boosting attack} or \textit{constrain-and-scaling attack} to increase the impact on the final global model.

\section{More Results on Model-boosting Attack}\label{app:moremodelboosting}


\begin{figure*}[!h]
\centering
\begin{subfigure}{\textwidth}\centering \caption{\scriptsize \textbf{Model Boosting Attack} (Adversaries = 1)}  
\includegraphics[clip,trim={0 3cm 0 0}, width=0.8\linewidth]{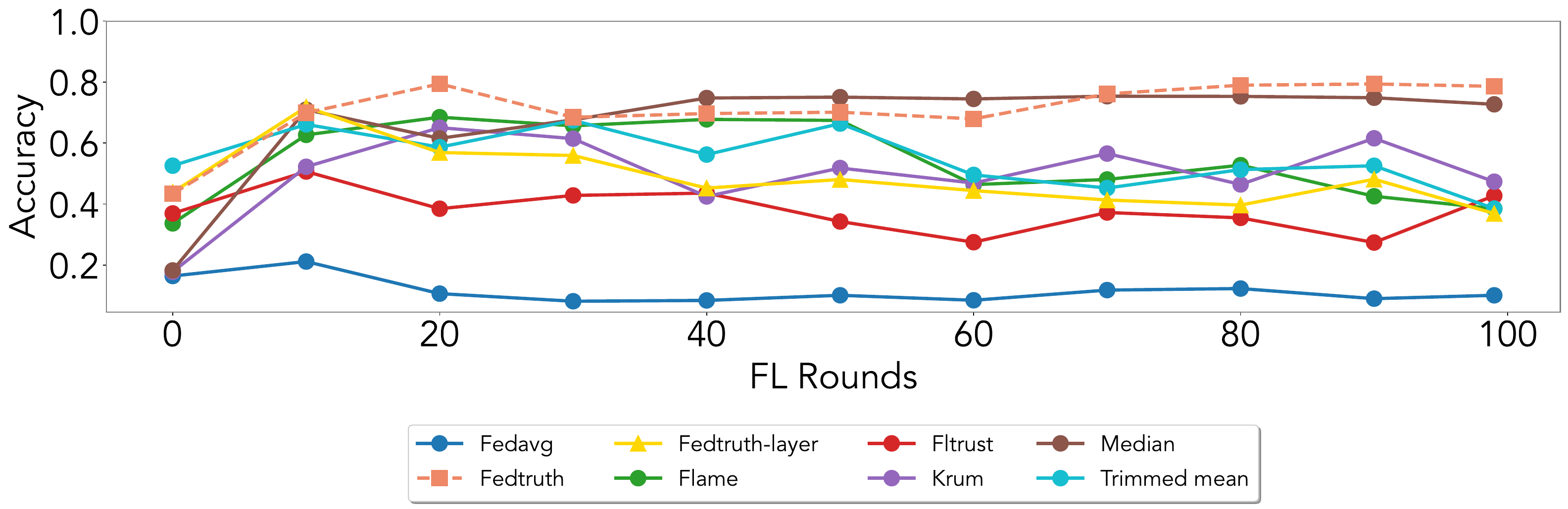}
\label{fig:model-boosting-attacks-FMNIST-1ADV}
\end{subfigure}
\begin{subfigure}{\textwidth}\centering \caption{\scriptsize \textbf{Model Boosting Attack} (Adversaries = 2)} \includegraphics[clip,trim={0 3cm 0 0}, width=0.8\linewidth]{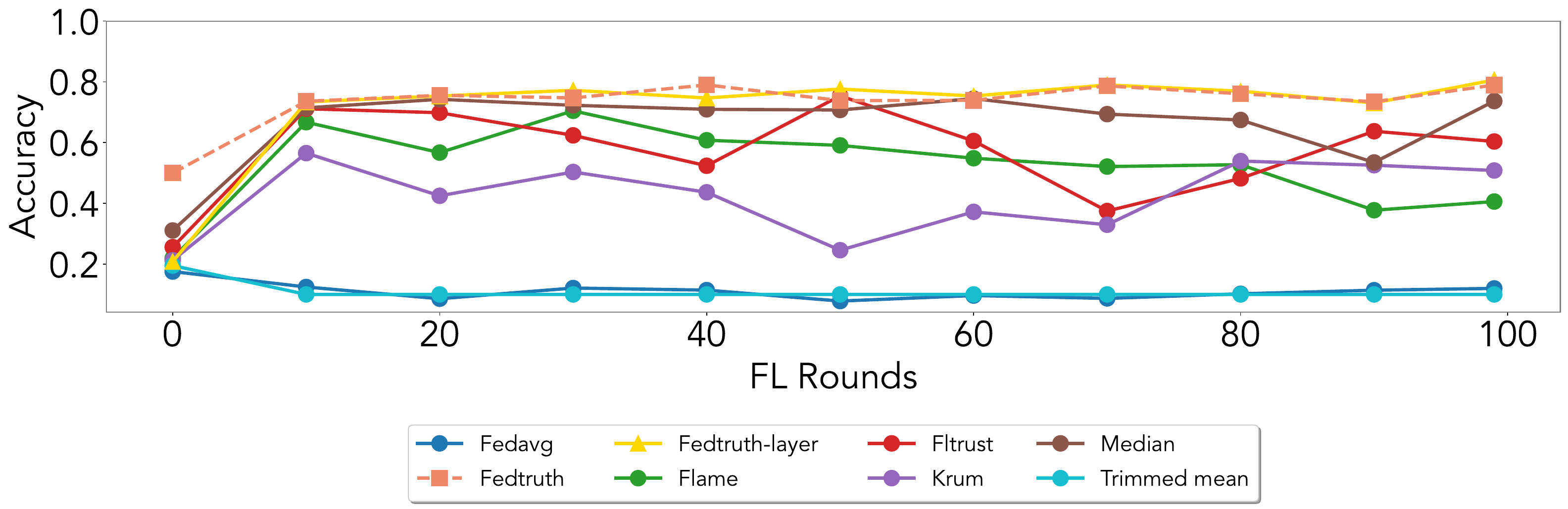}
\label{fig:model-boosting-attacks-FMNIST-2ADV}
\end{subfigure}
\begin{subfigure}{\textwidth}\centering \caption{\scriptsize \textbf{Model Boosting Attack} (Adversaries = 3)} \includegraphics[clip,trim={0 3cm 0 0}, width=0.8\linewidth]{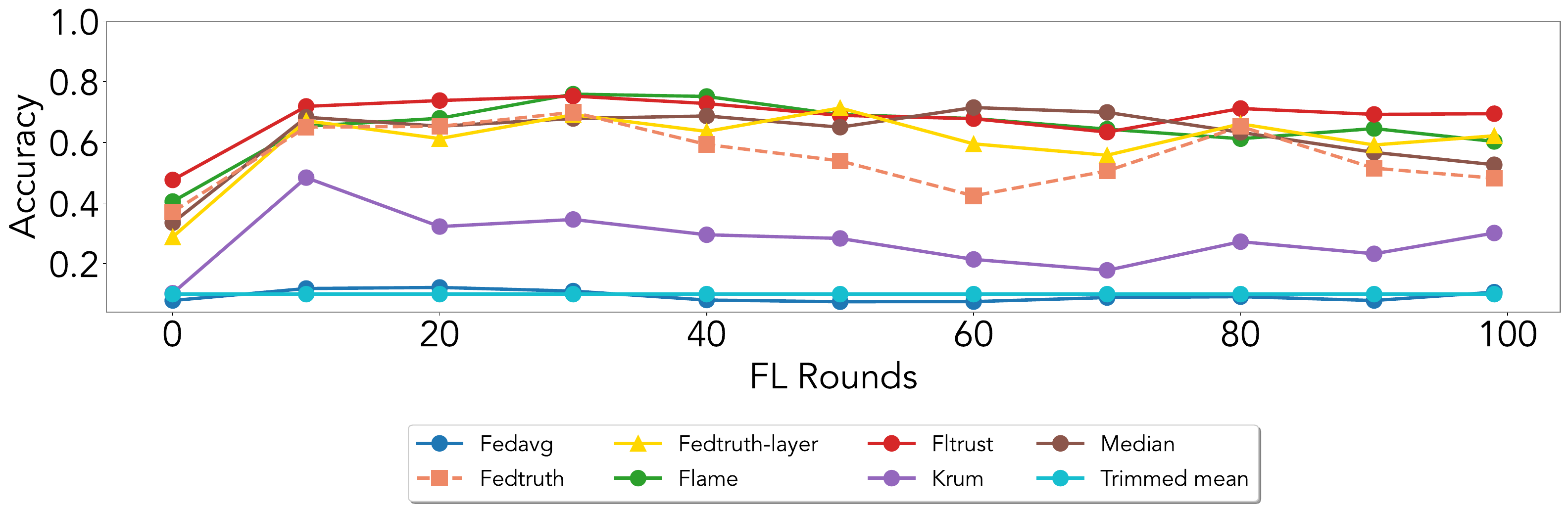}
\label{fig:model-boosting-attacks-FMNIST-3ADV}
\end{subfigure}
\begin{subfigure}{\textwidth}\centering \caption{\scriptsize \textbf{Model Boosting Attack} (Adversaries = 4)} 
\includegraphics[clip,trim={0 0 0 0}, width=0.8\linewidth]{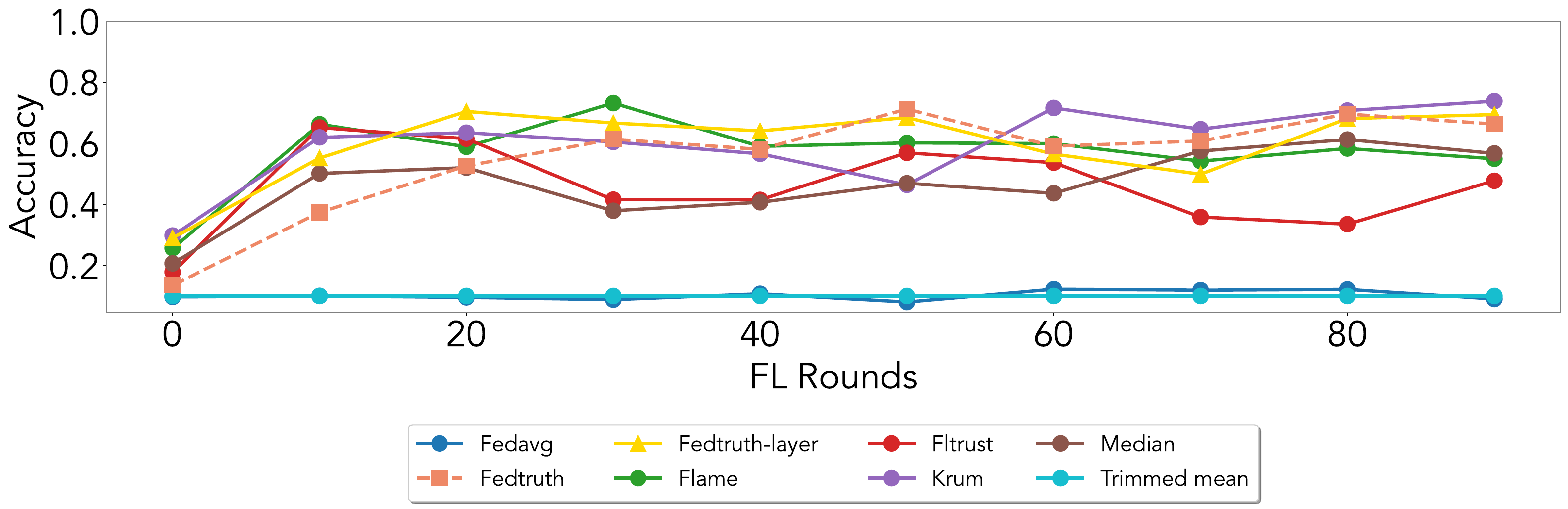}
\label{fig:model-boosting-attacks-FMNIST-4ADV}
\end{subfigure}
\caption{\scriptsize \textbf{Model Boosting Attack (FMNIST, ??10 boosting factor)}}\label{fig:model-boosting-attacks-fmnist}
\label{fig:model-boosting-attacks-FMNIST}
\end{figure*}

\textbf{Model-boosting Attack on FMNIST}: 
Figure~\ref{fig:model-boosting-attacks-fmnist} presents the results of the model-boosting attack on the FMNIST dataset, evaluating the impact of varying the number of adversaries per round from $1$ to $4$.


Figure~\ref{fig:model-boosting-attacks-FMNIST-1ADV} illustrates that when one adversarial client is selected per round, FedAvg is prevented from reaching convergence. This is likely due to FedAvg’s proportional distribution of adversarial updates, which, under non-iid data conditions, can amplify the effect of even a single adversary. As a result, FedAvg’s accuracy rapidly declines, stabilizing at an accuracy of 0\%. 
A subset of algorithms, specifically FedTruth-Layer, FLAME, Trimmed Mean, FLTrust, and Krum, also exhibit noteworthy degradation, with their performance reaching a final accuracy below 50\%. This trend suggests that the combination of non-iid data (80\%), the attack configuration, and the presence of even a single adversary (10\% of clients) can undermine the robustness of these aggregation strategies over time. However, FedTruth and Median offer the best results, reaching a convergence rate above 75\%.


Figures~\ref{fig:model-boosting-attacks-FMNIST-2ADV}-\ref{fig:model-boosting-attacks-FMNIST-4ADV} show that with 2, 3, or 4 adversaries, neither Median nor FedAvg converges. Furthermore, Krum reaches a convergence rate of 35\% when 3 adversaries are present, as seen in Figure~\ref{fig:model-boosting-attacks-FMNIST-3ADV}. The remaining algorithms during these experiments reach a minimum convergence rate of above 40\%. Interestingly, when 4 adversaries are present (Figure~\ref{fig:model-boosting-attacks-FMNIST-4ADV}), FLTrust is the worst performing algorithm that reached convergence. This is potentially due to FLTrust’s reliance on a representative root dataset, which may have been suboptimal. In contrast, FedTruth and FedTruth-Layer, which do not depend on a representative dataset and instead use adaptive aggregation, consistently outperform all other methods regardless of the number of adversaries.


We observe that FedTruth and FedTruth-Layer display similar robustness patterns as the number of adversaries increases with one or both reaches an optimal convergence rate. However, FedTruth achieves higher final accuracy than FedTruth-Layer when more adversaries are present (specifically, 2 - 4). This aligns with our hypothesis that FedTruth-Layer, designed for heightened sensitivity to subtle (stealthy) attacks, excels with fewer adversaries and, while FedTruth is more resilient as attack intensity increases.


\begin{figure*}[!t]
\centering
\begin{subfigure}{\textwidth}\centering \caption{\scriptsize \textbf{Model Boosting Attack} (Adversaries = 1)}  
\includegraphics[clip,trim={0 3cm 0 0}, width=0.8\linewidth]{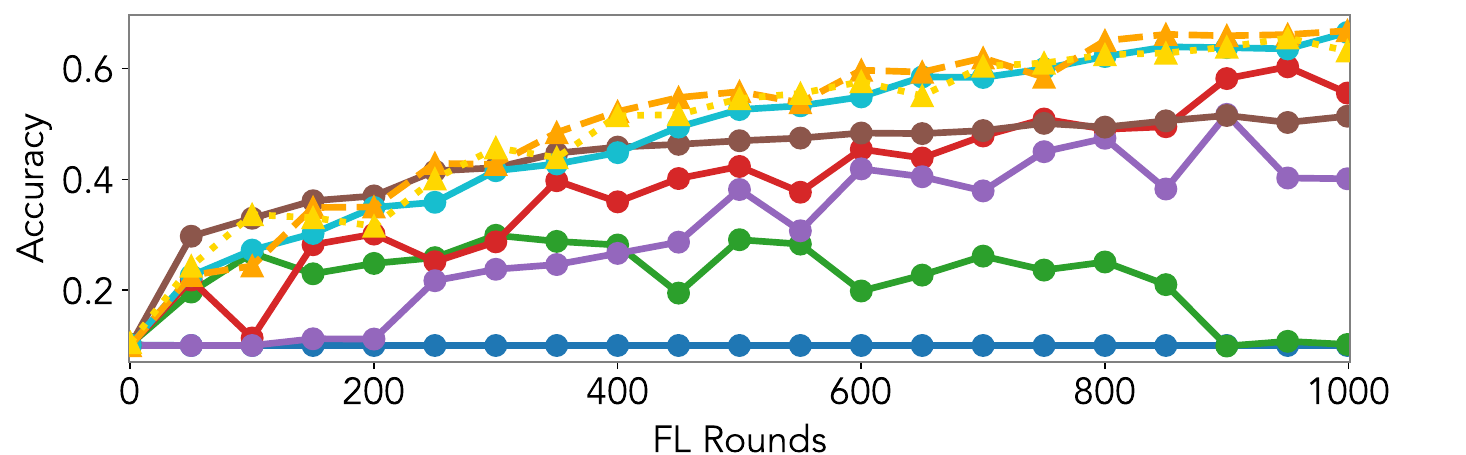}
\label{fig:model-boosting-attacks-CIFAR10-1ADV}
\end{subfigure}
\begin{subfigure}{\textwidth}\centering \caption{\scriptsize \textbf{Model Boosting Attack} (Adversaries = 2)}  

\includegraphics[clip,trim={0 3cm 0 0}, width=0.8\linewidth]{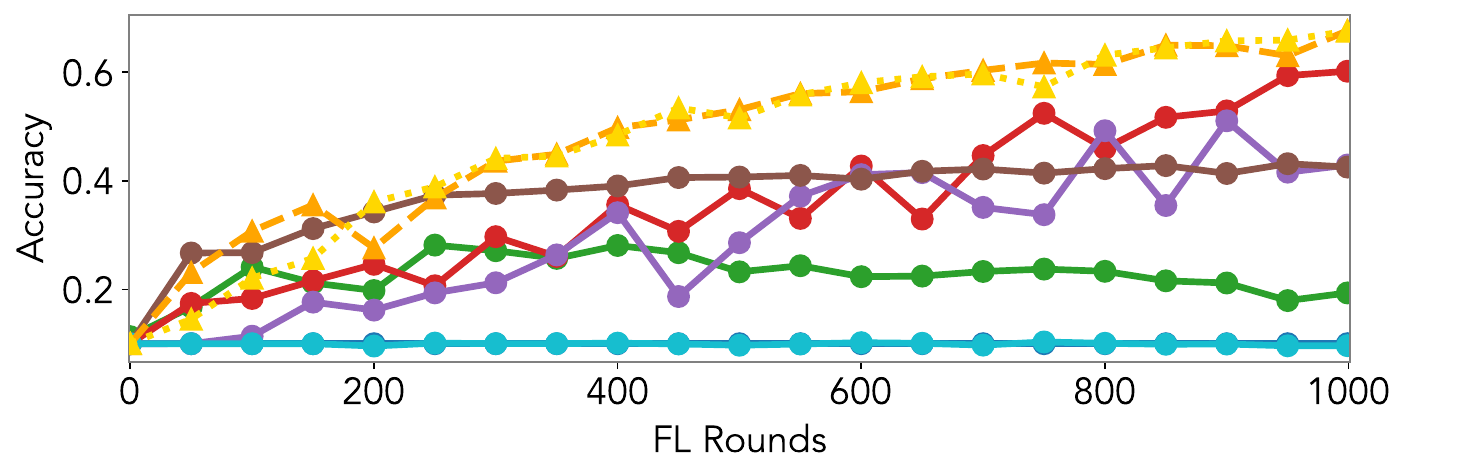}
\label{fig:model-boosting-attacks-CIFAR10-2ADV}
\end{subfigure}

\begin{subfigure}{\textwidth}\centering \caption{\scriptsize \textbf{Model Boosting Attack} (Adversaries = 3)}  
\includegraphics[clip,trim={0 3cm 0 0}, width=0.8\linewidth]{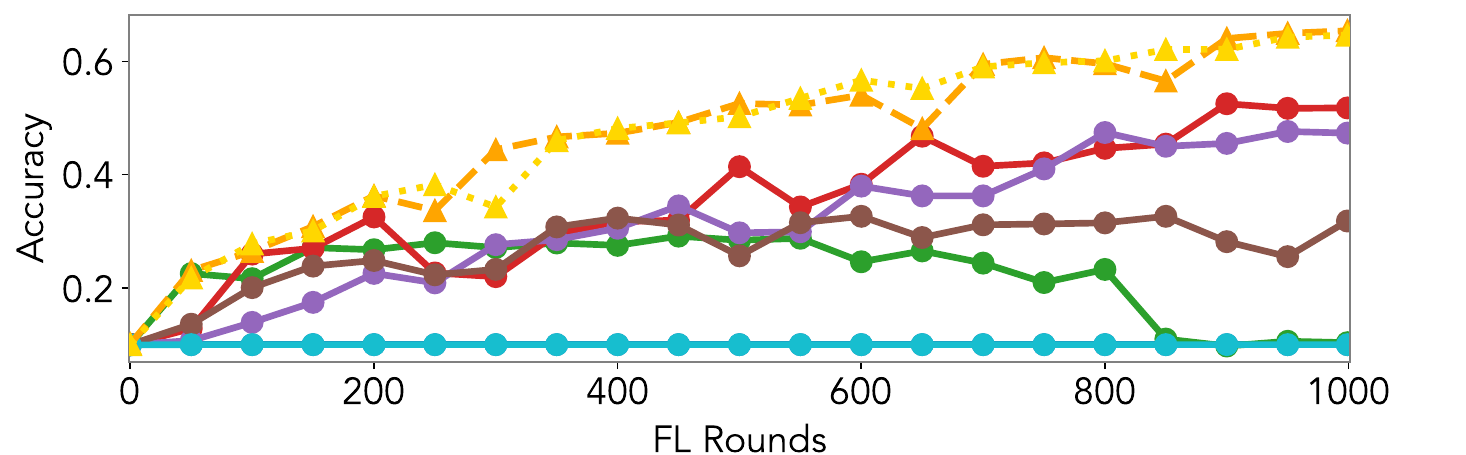}
\label{fig:model-boosting-attacks-CIFAR10-3ADV}
\end{subfigure}
\begin{subfigure}{\textwidth}\centering \caption{\scriptsize \textbf{Model Boosting Attack} (Adversaries = 4)}  
\includegraphics[clip,trim={0 0 0 0}, width=0.8\linewidth]{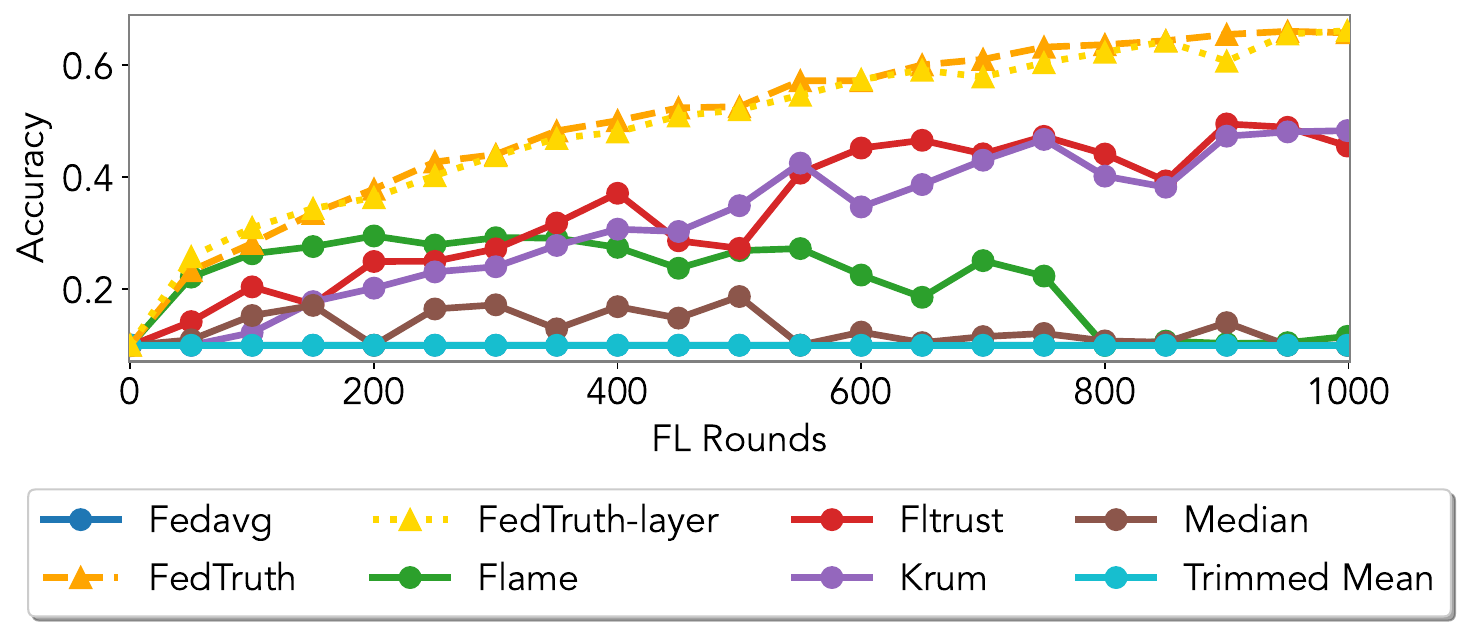}
\label{fig:model-boosting-attacks-CIFAR10-4ADV}
\end{subfigure}
\caption{\scriptsize \textbf{Model Boosting Attack (CIFAR-10, ??10 boosting factor)}}\label{fig:model-boosting-attacks-cifar10}
\label{fig:model-boosting-attacks-CIFAR10}
\end{figure*}

\textbf{Model-boosting Attack on CIFAR-10}: 
\textit{Figure~\ref{fig:model-boosting-attacks-CIFAR10}} presents our results for the \textit{model-boosting} attack, in which between $1$ and $4$ adversaries per iteration boost their local updates by a factor of $10$ on the CIFAR-10 dataset. We increased the number of iterations to 1,000, compared to the 100 iterations used for the MNIST and FMNIST experiments, to accommodate the slower convergence rate observed under our hyperparameter settings and non-iid sampling constraints. We used the \textit{ResNet-18} model for all CIFAR-10 experiments.


\textit{Figure~\ref{fig:model-boosting-attacks-CIFAR10-1ADV}} presents results for the model-boosting attack on the CIFAR-10 dataset when 1 adversary is present. These results indicate that all algorithms except FedAvg and FLAME converge, with both algorithms' accuracy falling below 1\% after 1,000 iterations. Krum's accuracy reaches approximately 40\%, lower than the other algorithms which reached convergence. Trimmed mean, FedTruth, and FedTruth-Layer's accuracy exceed 60\%, while the remaining algorithms finish above 50\%.

\textit{Figures~\ref{fig:model-boosting-attacks-CIFAR10-2ADV} and~\ref{fig:model-boosting-attacks-CIFAR10-3ADV}} show our results with 2 and 3 adversaries, respectively, revealing similar trends to Figure~\ref{fig:model-boosting-attacks-CIFAR10-1ADV}. Notably Trimmed mean fails to converge in both cases. When there are 2 adversaries, the Median algorithm's accuracy falls to just below 40\%, and with 3 adversaries, it falls further to around 30\%. In contrast, FedTruth and FedTruth-Layer maintain high final accuracy, unaffected by the increased number of adversaries. FLTrust converges to about 50\% accuracy after 1,000 iterations in the presence of 3 adversaries.

\textit{Figure~\ref{fig:model-boosting-attacks-CIFAR10-4ADV} }reports our results when there are 4 adversaries during each aggregation round. In this experiment, the Median algorithm’s accuracy collapsed, plateauing at 1\%, significantly below our proposed algorithms’ 60\% final accuracy. FLTrust’s accuracy drops below Krum’s, with both ending slightly below 40\%. In contrast, FedTruth and FedTruth-Layer maintain a robust convergence rate.

Overall, FedTruth and FedTruth-Layer consistently achieve final accuracies above 60\% across all attack configurations. As adversaries increase, their performance advantage over FLTrust, Median, and Krum becomes more pronounced, with divergences observable after iteration 250 in Figure~\ref{fig:model-boosting-attacks-CIFAR10-2ADV}, 225 in Figure~\ref{fig:model-boosting-attacks-CIFAR10-3ADV}, and 200 in Figure~\ref{fig:model-boosting-attacks-CIFAR10-4ADV}.
The diminished performance of Trimmed Mean, Krum, and Median can be attributed to their higher likelihood of incorporating adversarial or non-representative updates, which destabilizes aggregation. Krum, in particular, may overfit to dominant features early on due to non-iid sampling, yielding a suboptimal global model. FLTrust appears susceptible to bias toward server-side features when benign client contributions are limited. In contrast, FedTruth and FedTruth-Layer employ adaptive weighted averaging, enabling the integration of new features and minimizing the influence of any single label, regardless of the number of adversaries present.


\section{More Results on Gaussian Noise Attacks}\label{app:moregaussiannoise}

Figure~\ref{fig:gaussian-attacks-fmnist} presents our results for the Gaussian noise attack with and without model-boosting. We did not combine this attack with the constrain-and-scale method, whose design trains both a benign and adversarial model each epoch using separate loss functions for each dataset. Therefore, incorporating constrain-and-scale would not be possible with the Gaussian noise attack, since adversaries aim to degrade model performance by directly modifying model weights and thus do not rely on an adversarial dataset.


\textbf{Gaussian Noise Attacks on FMNIST}:
%
Figure~\ref{fig:gaussian-attacks-fmnist-base} offers our results for the \textit{Gaussian-noise} attack on the \textit{FMNIST} dataset when 3 adversaries are present. The top performers among them were \textit{FedTruth}, \textit{FedTruth-Layer}, and \textit{FLTrust} with their final accuracy above $65\%$. The second-best clustering of algorithms, consisting of the Krum, FLAME, FedAvg, and Median algorithms, yielded final accuracies ranging from $45\%$ to $60\%$. The lowest performing algorithm during this experiment was Trimmed mean with a final accuracy of $40\%$.


We first examine the lowest-performing algorithm, Trimmed mean, which we hypothesize underperforms due to its tendency to select adversarial models during aggregation. This issue likely arises as a result of the added adversarial noise being constrained, which creates smaller differences between benign and adversarial updates, increasing the probability of the algorithm inadvertently selecting a malicious update during aggregation. Furthermore, compared to similar algorithms like FedAvg, Trimmed mean’s vulnerability to adversarial selection is exacerbated by its use of a subset of the models during aggregation. This limited subset of models heightens the attack’s effectiveness, as even minor adversarial influences disproportionately impact the final global model.


The cluster of models with the second-best performance (specifically Krum, FLAME, FedAvg, and Median) showed peculiar behavior, initially performing similarly to the highest-performing algorithms (reaching accuracies ranging from 60\% to 70\%); however, their final accuracy declined to between 45\% and 60\%. This decline persisted throughout the experiment, indicating a significant decrease in performance for these algorithms. We hypothesize two main reasons for this behavior. First, the constrained amount of noise often fails to differ substantially from the updates of a benign model during certain rounds, causing adversarial updates to be selected inadvertently. This occurs because the differences between adversarial and benign models are initially more pronounced due to the non-iid nature of the data. However, as the global model starts to represent the overall data distribution better, these differences become subtler, gradually leading to performance degradation. Second, the limited adversarial noise introduced into the models is diluted through the averaging process over all updates (as in FLAME and FedAvg), reducing the adversarial update’s overall impact; however, as this continues, it gradually reduces the performance of the global model during the later iterations.

\begin{figure*}[!t]
    \begin{subfigure}{\textwidth}
        \captionsetup{font={tiny}}
        \centering
        \caption{Gaussian Noise Attack (base attack)}
        \label{fig:fig1}
        \includegraphics[clip,trim={0 4cm 0 0}, width=0.9\textwidth]{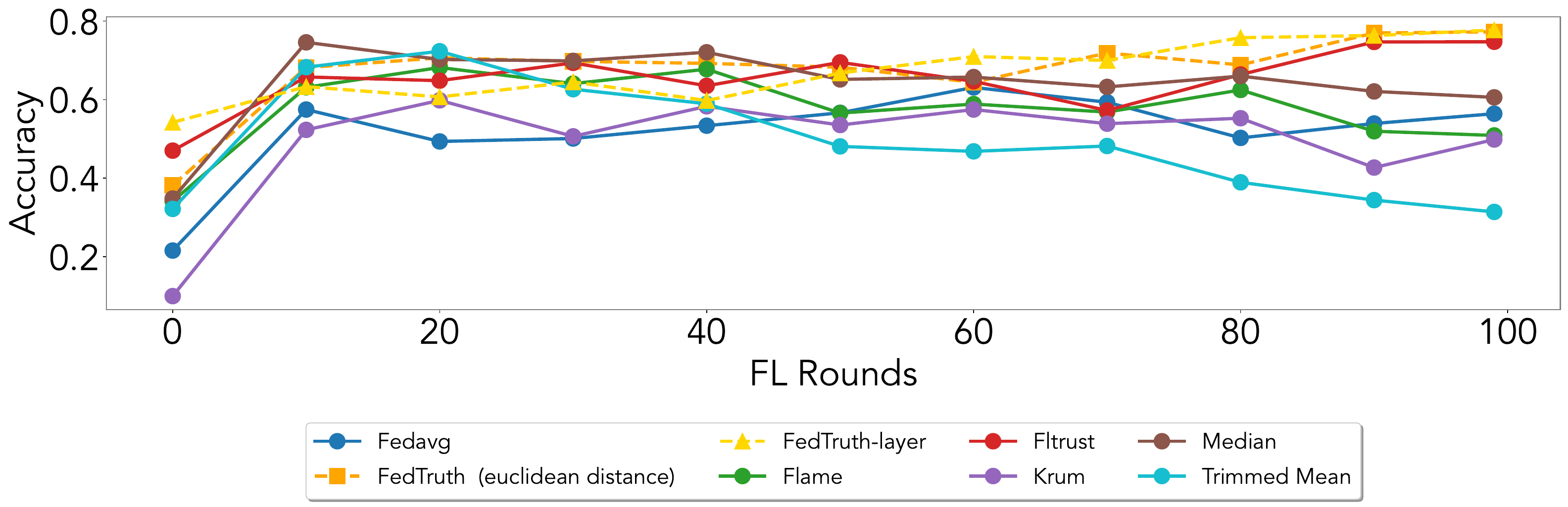}
    \end{subfigure}
    \hfill 

    \begin{subfigure}{\textwidth}
        \captionsetup{font={tiny}}
        \centering
        \caption{Gaussian Noise Attack (model-boosting attack)}
        \includegraphics[clip,trim={0 4cm 0 0}, width=0.9\textwidth]{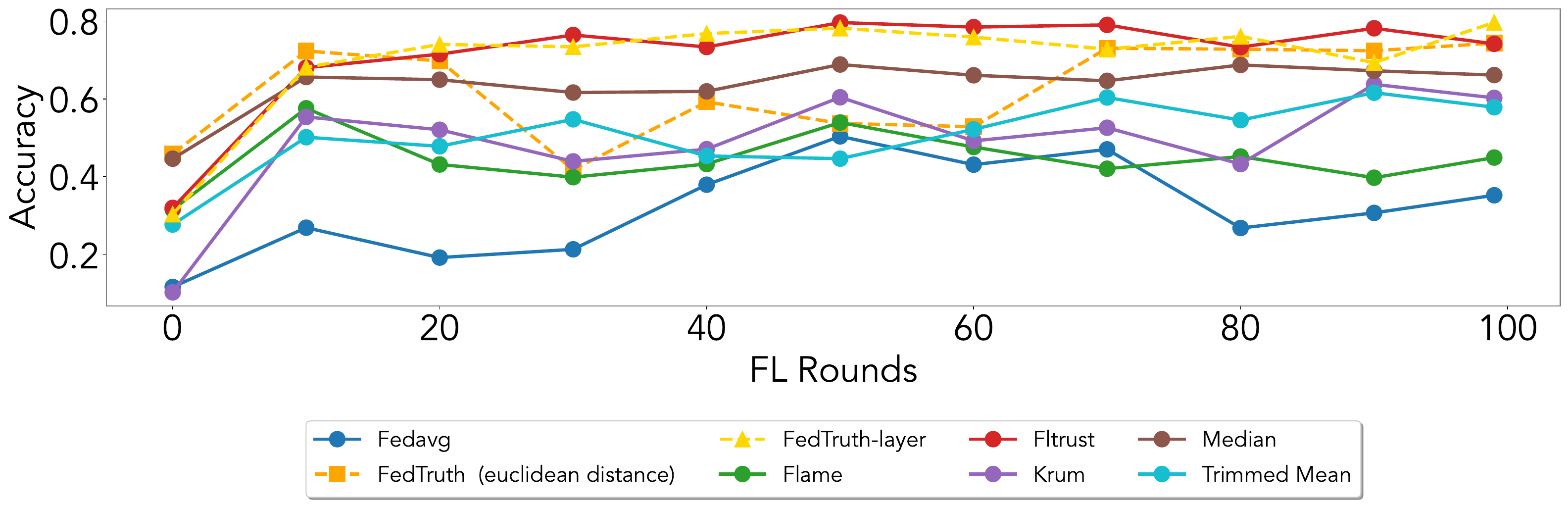}

    \end{subfigure}
  \caption{\textbf{Gaussian Noise Attack} (FMNIST, 3 Adversaries)}
    \label{fig:gaussian-attacks-fmnist}

\end{figure*}

\begin{figure*}[!t]

    \begin{subfigure}{\textwidth}
        \captionsetup{font={tiny}}
        \centering
        \caption{Gaussian Noise Attack (base attack)}
        \label{fig:fig2}
        \includegraphics[clip,trim={0 4cm 0 0}, width=0.9\textwidth]{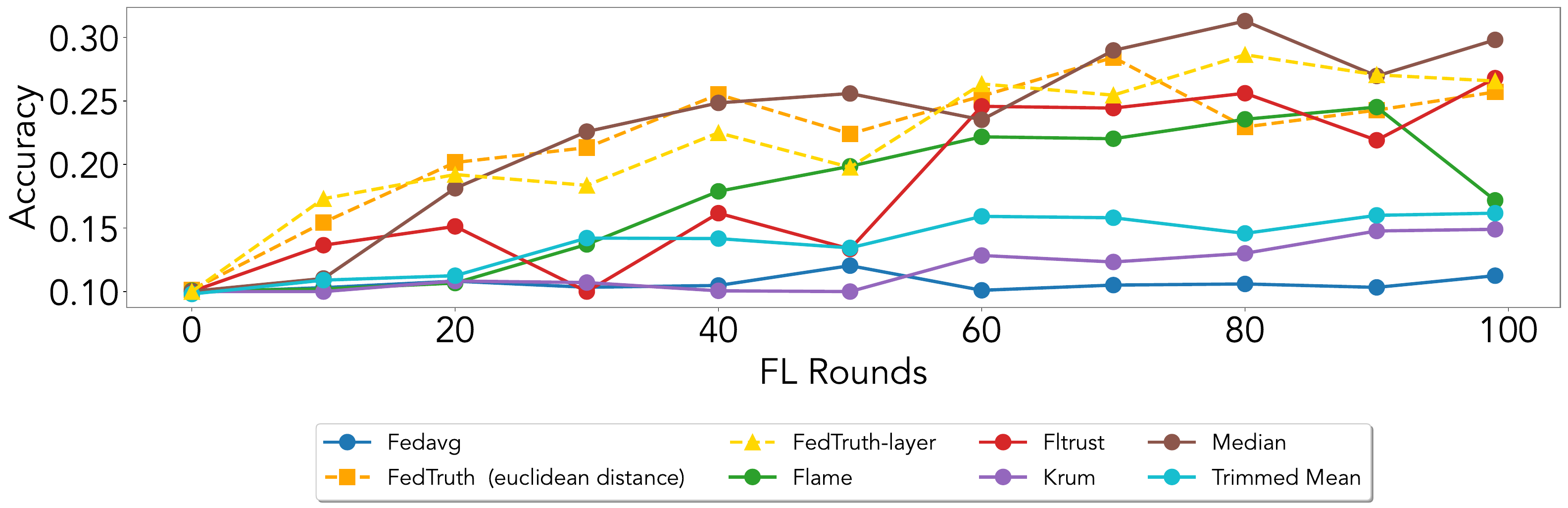}
    \end{subfigure}
    \hfill
    \begin{subfigure}{\textwidth}
        \captionsetup{font={tiny}}
        \centering
        \caption{Gaussian Noise Attack (model-boosting attack)}
        \label{fig:fig6}
        \includegraphics[width=0.9\textwidth]{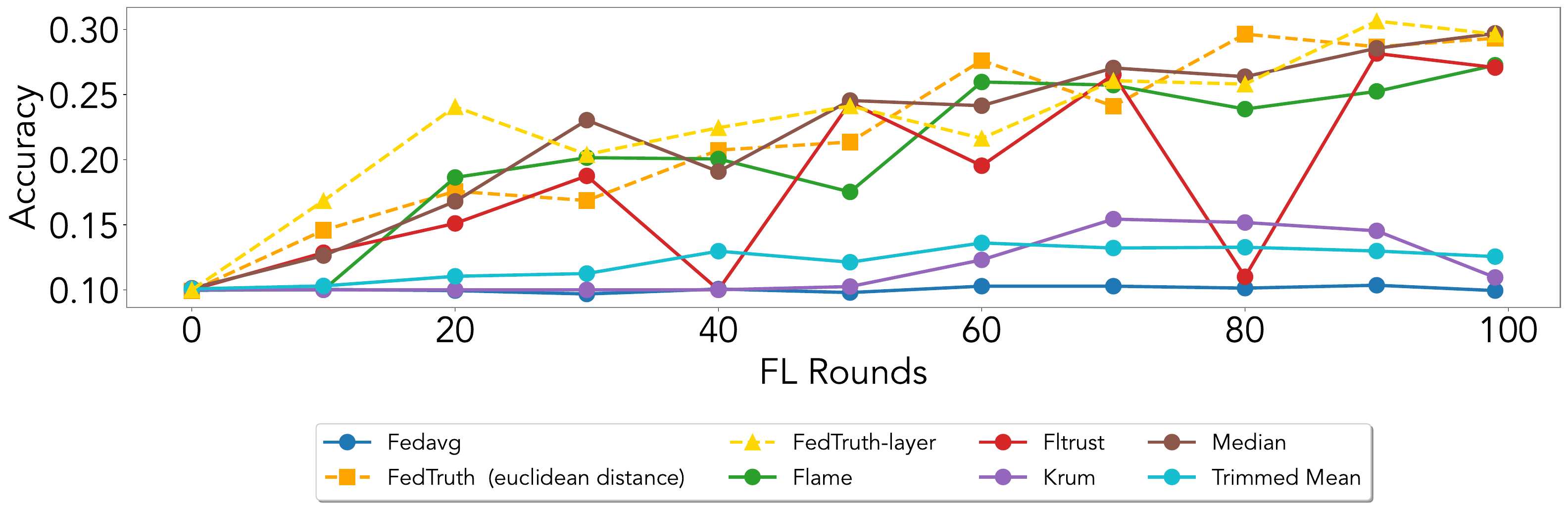}
    
    \end{subfigure}
  \caption{\textbf{Gaussian Noise Attack} (CIFAR-10, 3 Adversaries)}
  \label{fig:gaussian-attacks-cifar10}

\end{figure*}

The high-performing aggregation algorithms (\textit{FedTruth}, \textit{FedTruth-Layer}, and \textit{FLTrust}), used \textit{truth discovery} to calculate weights for each adversarial model. However, we noticed that FLTrust experiences slight decreases in accuracy around the 70th iterations. This may be due to the non-iid degree of the data sampling, causing the server model to not accurately represent the client's model. As a result, the weight assignment may overweight the adversarial models. Thankfully, the {adaptive} approach used by FedTruth and FedTruth-Layer solves this problem and avoids any dips in accuracy.



\textit{Figure~\ref{fig:gaussian-attacks-fmnist-model-boosting}} presents our results for the Gaussian Noise Attack when combined with the \textit{model-boosting} attack.
However, during this experiment, we observed results similar to those of the base attack (Figure~\ref{fig:gaussian-attacks-fmnist-base}), with the top-performing cluster still being FedTruth, FedTruth-Layer, and FLTrust. Between the 30th and 60th iterations, FedTruth experiences a decrease in accuracy, reaching 40\%. This dip does not persist in the final accuracy.
The second-best cluster of algorithms, achieving final accuracies ranging from 50\% to 60\%, now consists of Median, Krum, and Trimmed Mean, all of which perform better during this attack than during the base attack. This improvement was anticipated, as boosting the noised updates makes the adversarial updates more apparent, allowing these aggregation methods to more reliably remove adversarial updates before selecting a representative model (Median and Krum) or averaging a subset of models (Trimmed Mean).
FLAME and FedAvg become the worst-performing algorithms, finishing with accuracies of 38\% and 30\%, respectively. This outcome is expected, as the model-boosting attack can counteract the effects of adversarial weight delusion present during the base attack, making the attack more effective.

\textbf{Gaussian Noise Attacks on CIFAR-10}: {Figure~\ref{fig:gaussian-attacks-cifar10}} offers our results for the Gaussian Noise attack combined with the model-boosting attack (Figure~\ref{fig:gaussian-attacks-cifar10-model-boosting}) and without (Figure~\ref{fig:gaussian-attacks-cifar10-base}) on the CIFAR-10 dataset. During these attacks, it is apparent that \textit{FedTruth} and \textit{FedTruth-Layer} are among the best-performing algorithms. Additionally, these attack configurations significantly hinder FedAvg, Krum, and Trimmed mean. At the same time, FLAME and FLTrust experience a slower convergence rate. Additionally, during the base attack, FLAME has a backdoor accuracy of below 15\%, while FLTrust exhibits unstable accuracy, reaching a value below 10\% at the 40th and 80th iterations during the Gaussian noise attack with model-boosting.  Excluding FedTruth and FedTruth-Layer, all algorithms except for the Median algorithm suffered a slight drop in accuracy after 100 rounds when using the base or model-boosting attack combination. Furthermore, it is noteworthy that the Gaussian Noise attack combined with a model-boosting attack was more effective at degrading the performance of the CIFAR10 dataset than the FMNIST dataset. However, regardless of the attack combination or dataset selection, it is apparent that FedTruth and FedTruth-Layer are consistently among the top performers.

\begin{figure*}[!t]
    \centering
    \begin{subfigure}{\textwidth}
        \captionsetup{font={tiny}}
        \centering
        \caption{PGD Attack (base attack) - Main Task Accuracy}
          \label{fig:pgd-attacks-mnist-ma} 
        \includegraphics[clip,trim={0 4cm 0 0}, width=0.8\textwidth]{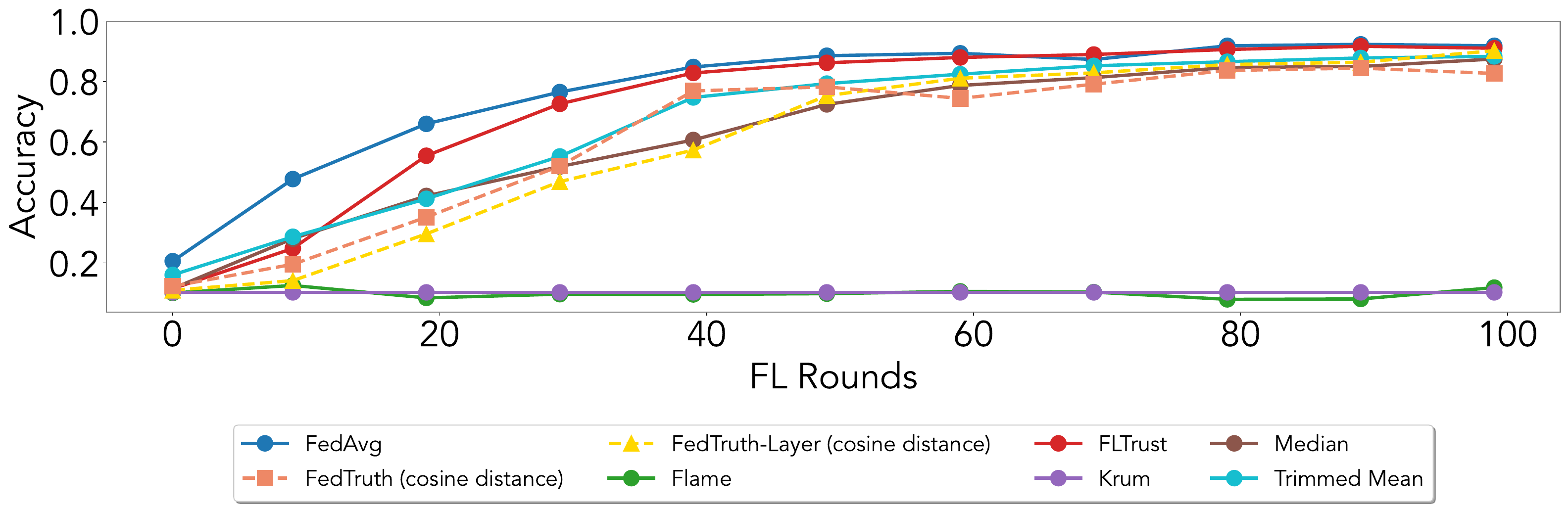}
    \end{subfigure}
    \hfill 
    \begin{subfigure}{\textwidth}
        \captionsetup{font={tiny}} 
        \centering
        \caption{PGD Attack (base attack) - Backdoor Accuracy}
          \label{fig:pgd-attacks-mnist-ba} 
        \includegraphics[clip,trim={0 0 0 0}, width=0.8\textwidth]{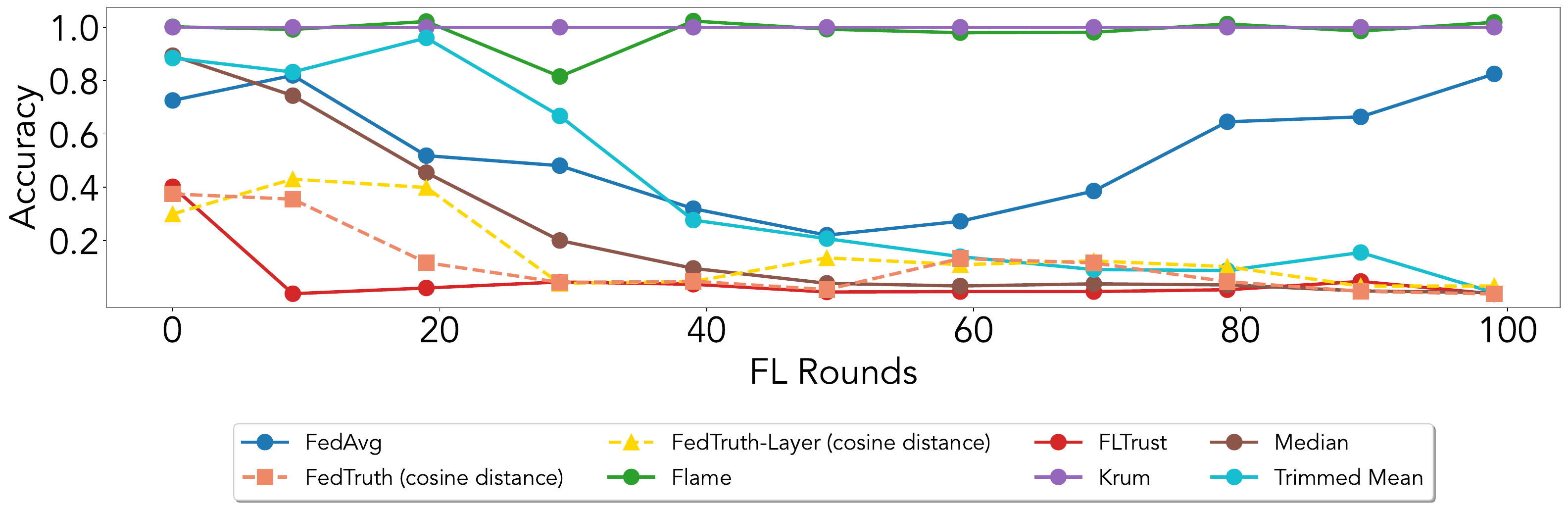}
    \end{subfigure}
  \caption{\normalsize \textbf{PGD Attack - base attack} (MNIST, 3 adversaries)}
  \label{fig:pgd-attacks-mnist} 
\end{figure*}

\begin{figure*}[!t]
    \centering
    \begin{subfigure}{\textwidth}
        \captionsetup{font={tiny}}
        \centering
        \caption{PGD Attack (model-boosting attack) - Main Task Accuracy}
          \label{fig:pgd-attacks-mnist-mb-ma} 
        \includegraphics[clip,trim={0 4cm 0 0}, width=0.8\textwidth]{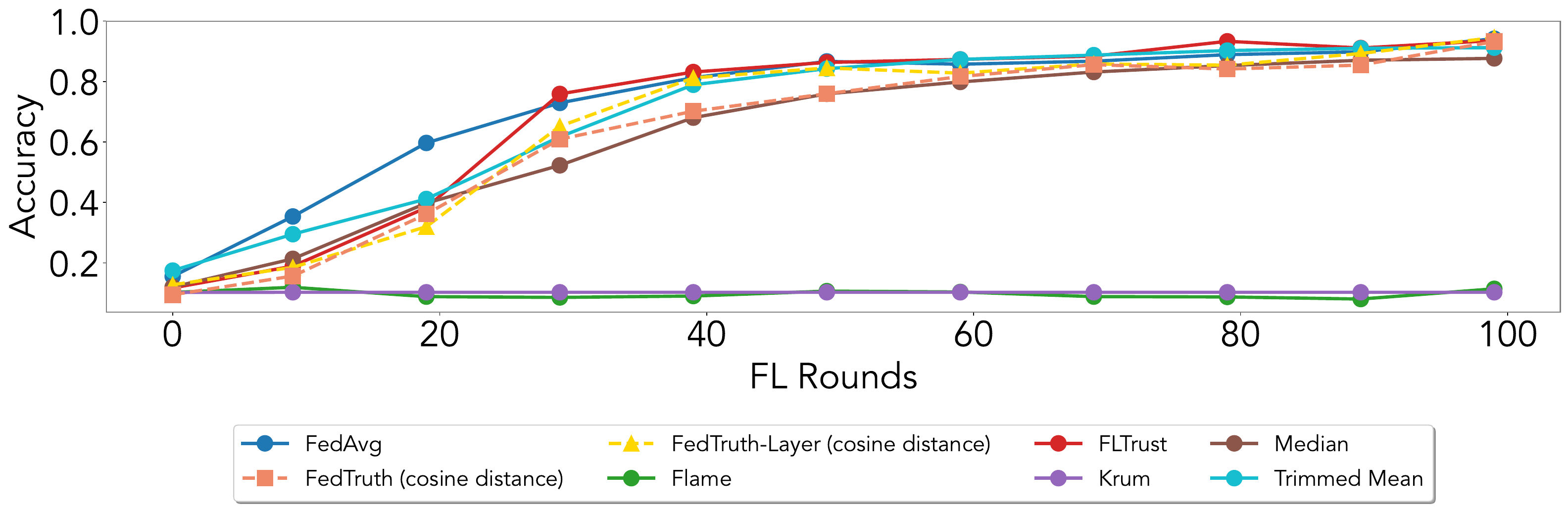}
    \end{subfigure}
    \hfill 
    \begin{subfigure}{\textwidth}
        \captionsetup{font={tiny}}
        \centering
        \caption{PGD Attack (model-boosting attack) - Backdoor Accuracy}
          \label{fig:pgd-attacks-mnist-mb-ba} 
        \includegraphics[clip,trim={0 0 0 0}, width=0.8\textwidth]{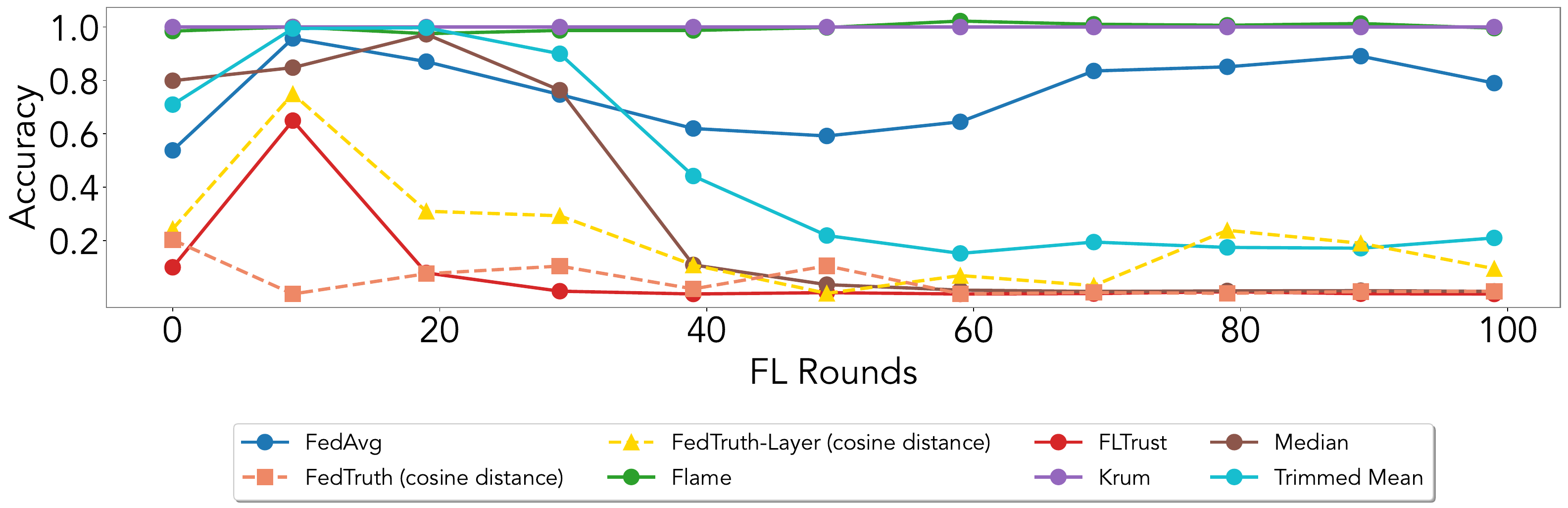}
    \end{subfigure}
  \caption{\normalsize \textbf{PGD Attack - combine with Model-Boosting Attack} (MNIST, 3 adversaries)}
  \label{fig:pgd-attacks-mnist-mb} 
\end{figure*}

\begin{figure*}[!t]
    \centering

    \begin{subfigure}{\textwidth}
        \captionsetup{font={tiny}}
        \centering
        \caption{PGD Attack (constrain-and-scale attack) - Main Task Accuracy}
          \label{fig:pgd-attacks-mnist-cs-ma} 
        \includegraphics[clip,trim={0 4cm 0 0}, width=0.8\textwidth]{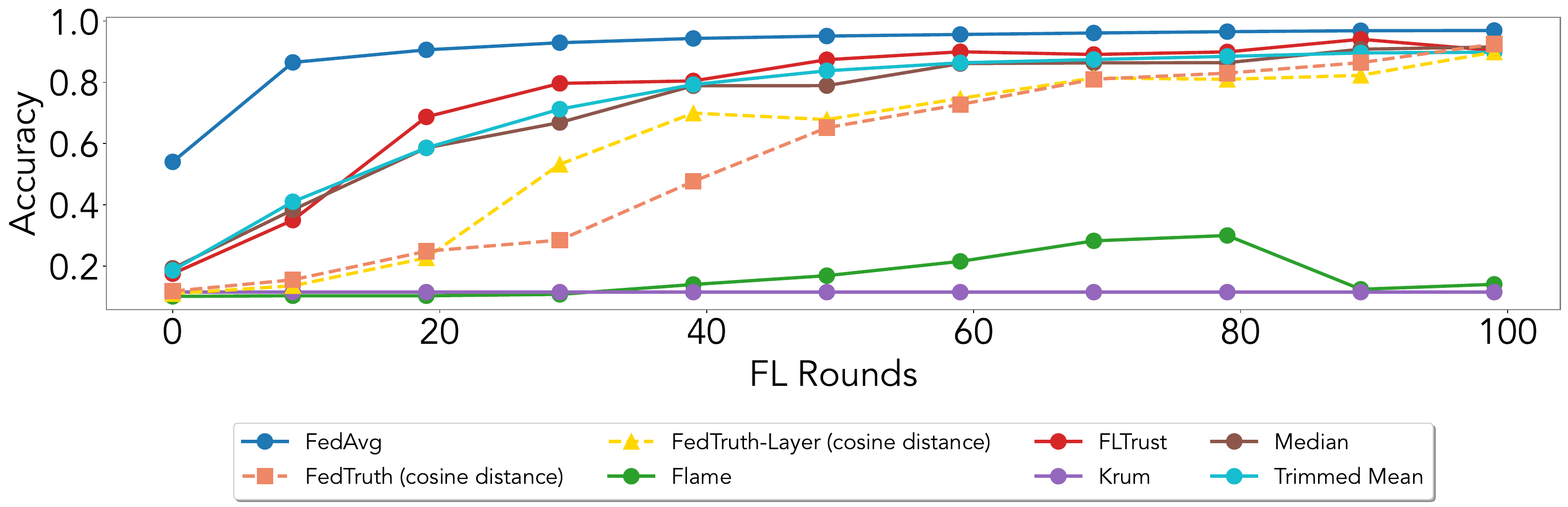}
    \end{subfigure}
    \hfill
    \begin{subfigure}{\textwidth}
        \captionsetup{font={tiny}}
        \centering
        \caption{PGD Attack (constrain-and-scale attack) - Backdoor Accuracy}
          \label{fig:pgd-attacks-mnist-cs-ba} 
        \includegraphics[width=0.8\textwidth]{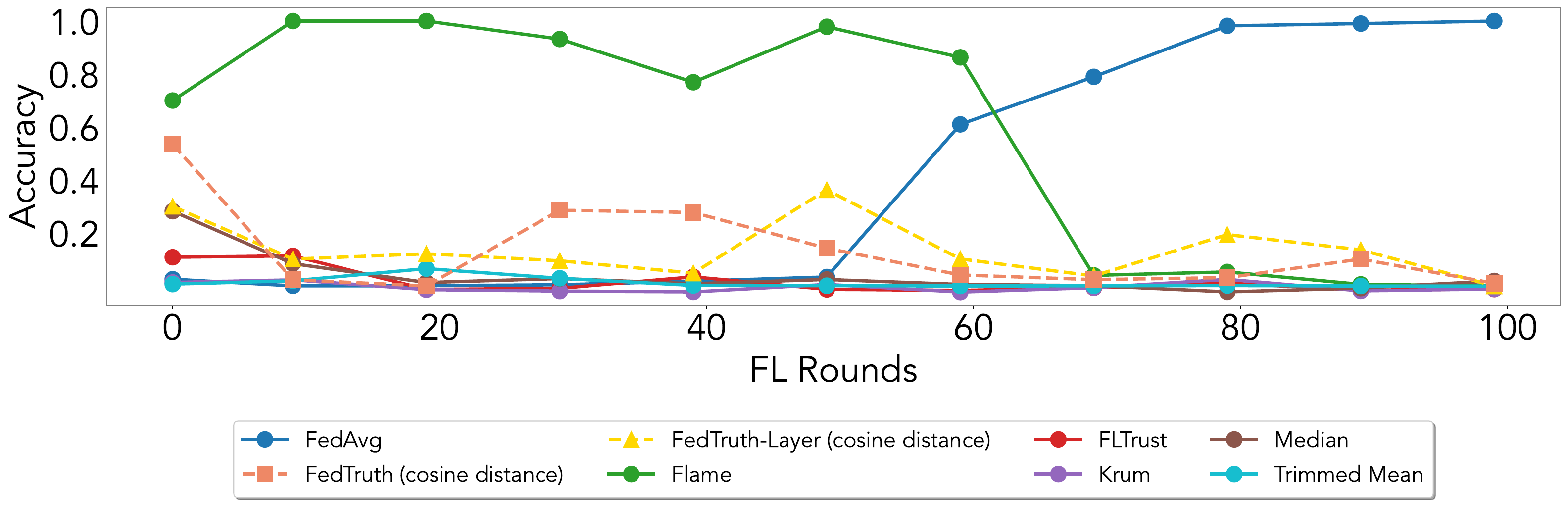}
    \end{subfigure}
  \caption{\normalsize \textbf{PGD Attack - combine with Constrain-and-Scale Attack} (MNIST, 3 adversaries)}
  \label{fig:pgd-attacks-mnist-cs} 
\end{figure*}

\section{More Results on Backdoor Attacks}\label{app:more-backdoor-attacks}

In this section, we present more findings for \textit{target task accuracy} (accuracy on an adversarial backdoor dataset) and the \textit{main task accuracy} (accuracy on a benign dataset) for the projected gradient descent and edge-case attacks observed during each round of aggregation. We used the cosine distance metric during these experiments when running the FedTruth and FedTruth-Layer algorithms.  

\textbf{Projected Gradient Descent Attack (\textit{PGD})}: 
We implemented the \textit{PGD} attacks with the \textit{Torch Attacks} \cite{kim2020torchattacks} library, which creates a generative model that takes as input an image and returns a perturbed version of the image. We set the max perturbation ($\epsilon=.3$), which is how the adversarial example knows how far an image can be noised while generating the adversarial model. Then we set the step size ($\ alpha=.03$) and the number of steps ($10$).

Figures~\ref{fig:pgd-attacks-mnist-ma},~\ref{fig:pgd-attacks-mnist-mb-ma}, and~\ref{fig:pgd-attacks-mnist-cs-ma} present the \textit{main task accuracy} for the \textit{PGD} attacks. During all of the attack, Krum and Flame do not reach convergence on the \textit{main task}. During the \textit{constrain-and-scale} (Figure~\ref{fig:pgd-attacks-mnist-cs-ma}) attack, FedTruth and FedTruth-Layer have a slower convergence rate than the other models that were able to reach convergence. However, these algorithms still reach a final accuracy above 80\%.


Figures~\ref{fig:pgd-attacks-mnist-ba},~\ref{fig:pgd-attacks-mnist-mb-ba}, and \ref{fig:pgd-attacks-mnist-cs-ba} show our results for the \textit{targeted task} accuracy during the attack combinations considered in this section. Krum and FLAME failed to converge on the main task and were excluded from the targeted task accuracy analysis. The results indicate that of the algorithms that demonstrated convergence on the main task, \textit{FedAvg} was the only algorithm to exhibit significant vulnerabilities to backdoor attacks, achieving a final backdoor accuracy above 80\% across all tested attack scenarios. 

\begin{figure*}[!t]
    \centering
    \begin{subfigure}{\textwidth}
        \captionsetup{font={tiny}}
        \centering
        \caption{Edge Case Attack (base attack) - Main Task Accuracy}
        \label{fig:egc-attacks-mnist-ma} 
        \includegraphics[clip,trim={0 4cm 0 0}, width=0.9\textwidth]{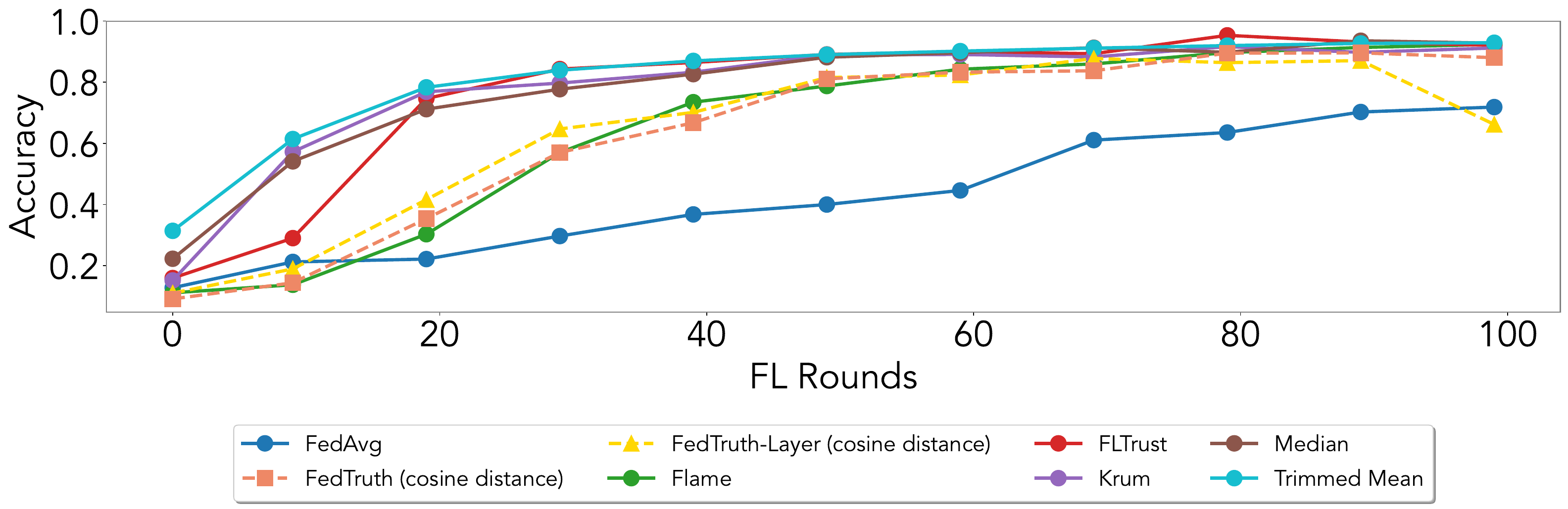}
    \end{subfigure}
    \hfill 
    \begin{subfigure}{\textwidth}
        \captionsetup{font={tiny}} 
        \centering
        \caption{Edge Case Attack (base attack) - Backdoor Accuracy}
        \label{fig:egc-attacks-mnist-ba} 
        \includegraphics[clip,trim={0 0 0 0}, width=0.9\textwidth]{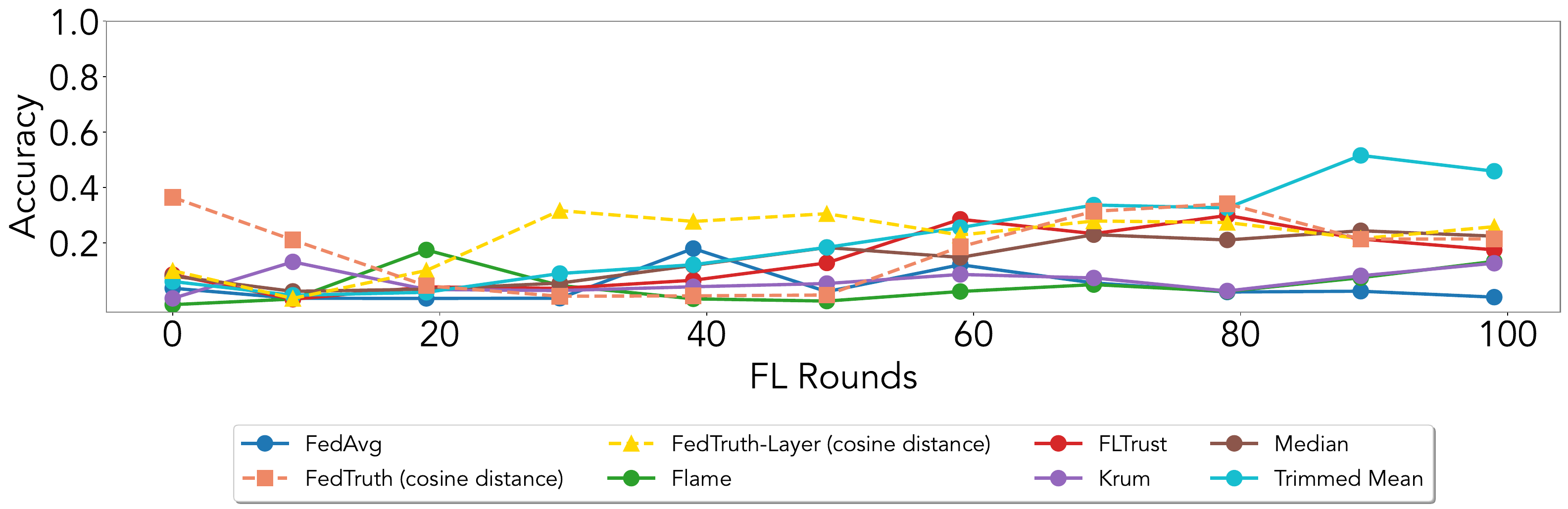}
    \end{subfigure}

  \caption{\normalsize \textbf{Edge Case Attack - base attack} (MNIST, 3 adversaries)}
  \label{fig:egc-attacks-mnist} 
\end{figure*}

\begin{figure*}[!t]
    \centering
    \begin{subfigure}{\textwidth}
        \captionsetup{font={tiny}}
        \centering
        \caption{Edge Case Attack (model-boosting attack) - Main Task Accuracy}
        \label{fig:egc-attacks-mnist-mb-ma} 
        \includegraphics[clip,trim={0 4cm 0 0}, width=0.9\textwidth]{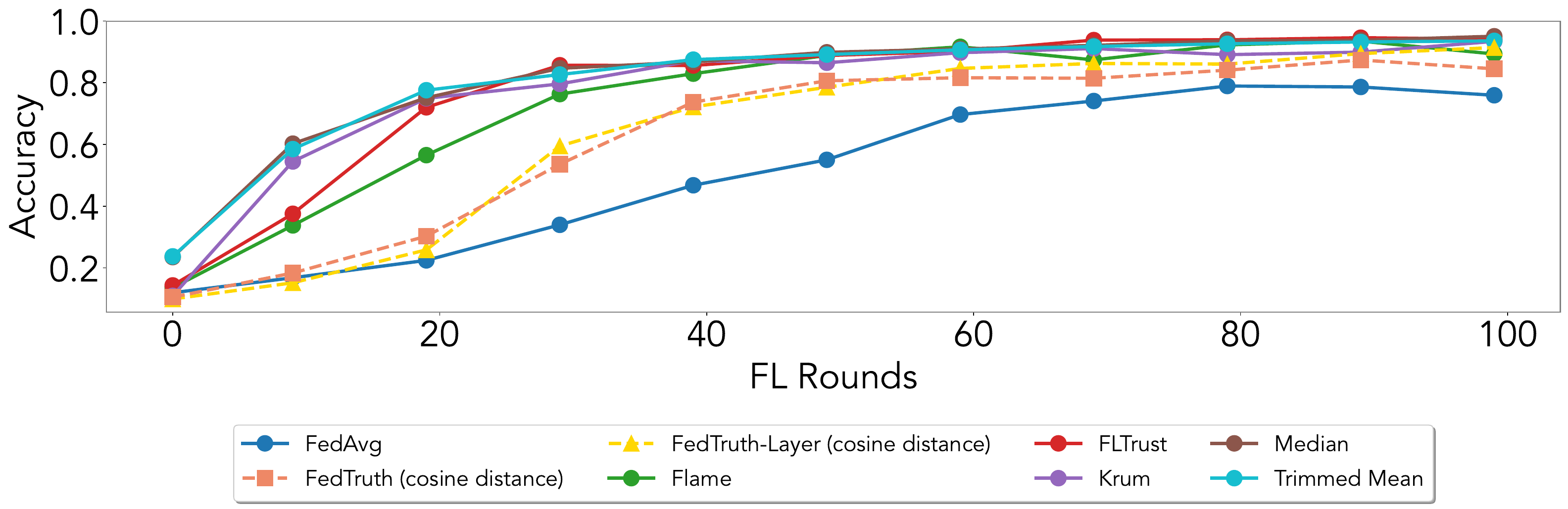}
    \end{subfigure}
    \hfill 
    \begin{subfigure}{\textwidth}
        \captionsetup{font={tiny}}
        \centering
        \caption{Edge Case Attack (model-boosting attack) - Backdoor Accuracy}
        \label{fig:egc-attacks-mnist-mb-ba} 
        \includegraphics[clip,trim={0 0 0 0}, width=0.9\textwidth]{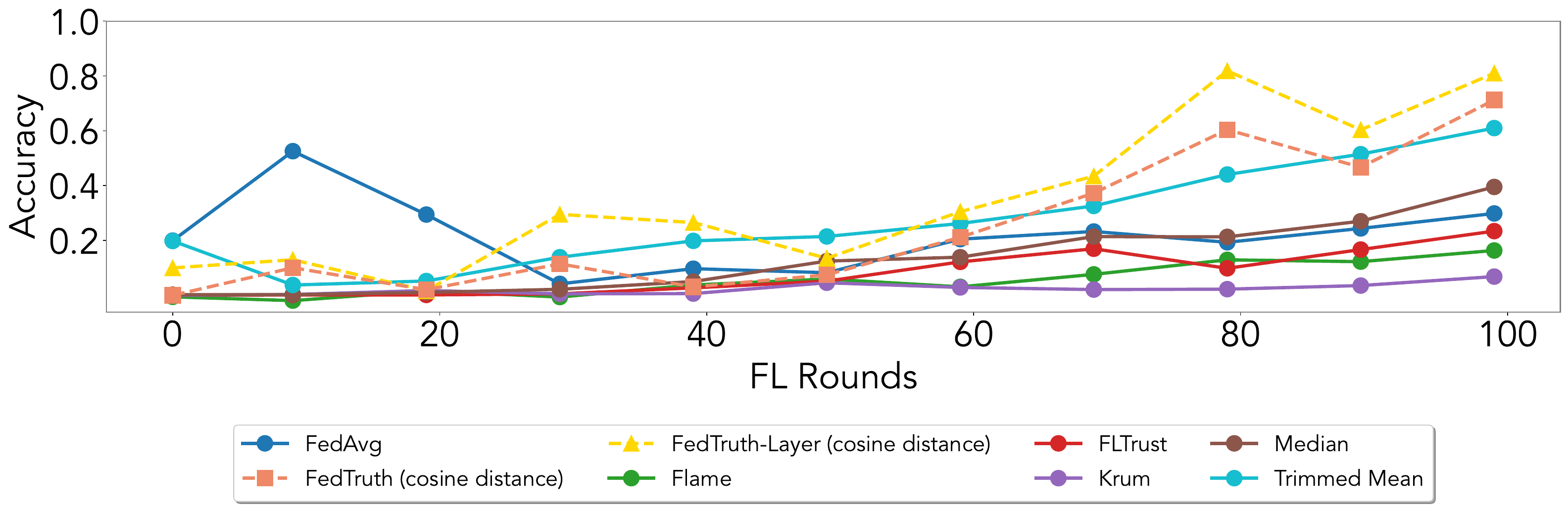}
    \end{subfigure}
    
  \caption{\normalsize \textbf{Edge Case Attack - combine with model-boosting attack} (MNIST, 3 adversaries)}
  \label{fig:egc-attacks-mnist-mb} 
\end{figure*}

\begin{figure*}[!t]
    \centering
    \begin{subfigure}{\textwidth}
        \captionsetup{font={tiny}}
        \centering
        \caption{Edge Case Attack (constrain-and-scale attack) - Main Task Accuracy}
        \includegraphics[clip,trim={0 4cm 0 0}, width=0.9\textwidth]{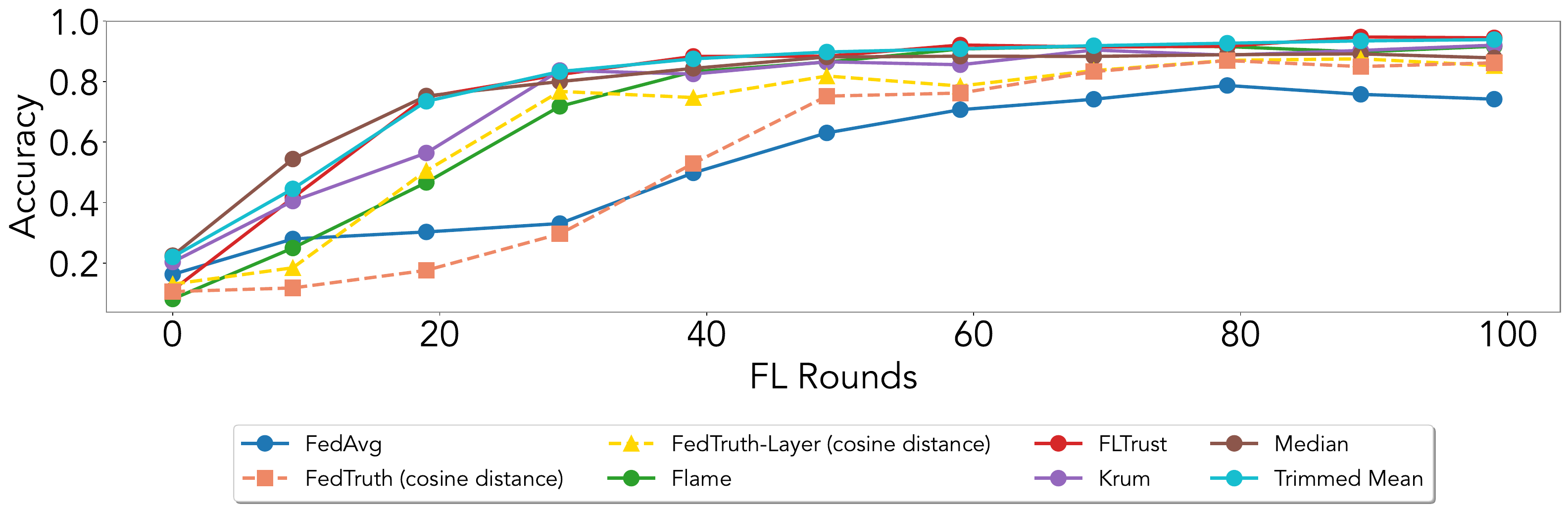}
        \label{fig:egc-attacks-mnist-cs-ma} 
    \end{subfigure}
    \hfill
    \begin{subfigure}{\textwidth}
        \captionsetup{font={tiny}}
        \centering
        \caption{Edge Case Attack (constrain-and-scale attack) - Backdoor Accuracy}
        \label{fig:egc-attacks-mnist-cs-ba} 
        \includegraphics[width=0.9\textwidth]{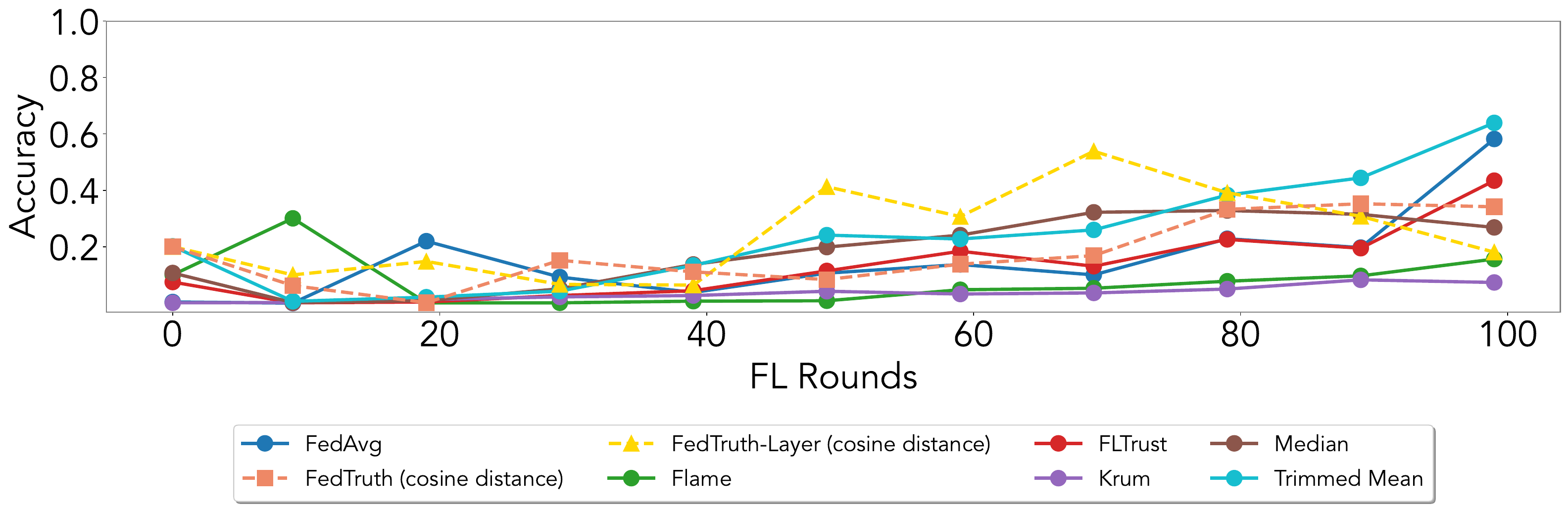}
    \end{subfigure}
  \caption{\normalsize \textbf{Edge Case Attack - combine with constrain-and-scale attack} (MNIST, 3 adversaries)}
  \label{fig:egc-attacks-mnist-cs} 
\end{figure*}

\textbf{Edge-case Attack}:
Based on the attacks presented in~\cite{wang2020attack}, we implemented the \textit{edge-case} attack for the MNIST dataset. During this attack, we used the \textit{Arkiv Digital Sweden (ARDIS)~\cite{kusetogullari2020ardis}} dataset as adversarial images (\textit{edge-cases}) being injected as backdoors into the models. During each training round, the adversaries examined their benign local data points to locate those corresponding to the targeted labels. Using this knowledge, each adversarial client added targeted edge-case data points to their training dataset. The number of these added data points was set to be equal to \(20\%\) of the matching benign data points.

Figures~\ref{fig:egc-attacks-mnist-ma},~\ref{fig:egc-attacks-mnist-mb-ma}, and~\ref{fig:egc-attacks-mnist-cs-ma} show our results for the \textit{main task} during the edge-case attacks. We observed that all the algorithms could reach convergence during the base attack. However, during the \textit{model-boosting} (Figure~\ref{fig:egc-attacks-mnist-mb-ma}) and \textit{constrain-and-scale} (Figure~\ref{fig:egc-attacks-mnist-cs-ma}) attacks, FedTruth and FedTruth-Layer had a slower convergence rate than the other algorithms, excluding FedAvg, which had the slowest convergence rate during all of the attack configurations.


Figures~\ref{fig:egc-attacks-mnist-ba},~\ref{fig:egc-attacks-mnist-mb-ba}, and~\ref{fig:egc-attacks-mnist-cs-ba} present our results for the \textit{targeted task} accuracy. During the base attack (Figure~\ref{fig:egc-attacks-mnist-ba}), we see that the least effective algorithm was Trimmed mean with a final backdoor accuracy around \(40\%\) after 100 iterations. The remaining algorithms, FedTruth-Layer, FedAvg, Krum, and FLTrust, mostly removed the backdoor, with a final backdoor accuracy below \(20\%\).

Figure~\ref{fig:egc-attacks-mnist-mb-ba} presents our results for backdoor accuracy during the \textit{edge-case with model-boosting} attack. We observed that FLTrust and Krum were the best-performing algorithms, both finishing with a backdoor accuracy below 20\%. The Median and FedAvg algorithms also performed well, finishing with a backdoor accuracy of slightly above 25\%. However, Trimmed Mean, FedTruth, and FedTruth-Layer did not perform as well, each finishing with a backdoor accuracy around 60\%. We suspect that the degraded performance of our algorithm is due to the small difference between the adversarial and benign datasets, which allows adversarial updates to receive too high a weight during aggregation. As a result, when model boosting is applied, the adversarial model can hijack the global model and insert the backdoor artifact. Nevertheless, when using a different distance metric that takes into consideration the magnitude of the client updates (i.e., Euclidean Distance), we observe better performance, as discussed in Appendix~\ref{app:more-distance-functions}.



Figure~\ref{fig:egc-attacks-mnist-cs-ba} shows the effect of the  \textit{edge-case attack with constrain-and-scale} where FedTruth, FLAME, and Krum are the most robust algorithms, finishing with a final accuracy below 10\%. FedTruth-Layer was slightly less effective during this attack, finishing with a targeted task accuracy of around \(30\%\). FLTrust, Krum, and Median produced similar results to FedTruth-Layer, finishing with a backdoor accuracy below \(40\%\). During this attack, Trimmed Mean and FedAvg were the most susceptible algorithms, finishing with a backdoor accuracy of \(60\%\).

\section{More Results on Distance Functions}\label{app:more-distance-functions}
 Figures \ref{fig:gaussian-attacks-mnist-distance}, \ref{fig:pgd-attacks-mnist-distance},  and \ref{fig:egc-attacks-mnist-distance} compare the effects that different distance metrics (Euclidean, Manhattan, angular, cosine, and custom distance) have on the FedTruth and FedTruth-Layer. We compare these results during the Gaussian Noise (Figure \ref{fig:gaussian-attacks-mnist-distance}), PGD (Figure \ref{fig:pgd-attacks-mnist-distance}), and edge-case attacks (Figure \ref{fig:egc-attacks-mnist-distance}). Additionally, combined the Byzantine (Gaussian noise) attack with model boosting and the backdoor (PGD and Edge-case) attack with \textit{model-boosting} and \textit{constrain-and-scale} attacks. 

\begin{figure}[htp]
    \captionsetup{font={tiny}} 
    \centering
    \begin{subfigure}{\textwidth}
        \centering
        \caption{Gaussian Noise Attack (base attack)}
        \label{fig:gaussian-attacks-mnist-base-distance}
        \includegraphics[clip,trim={0 5.3cm 0 0}, width=0.9\textwidth]{figures/results/Gaussian_Noise_Attack/Gaussian_Noise_Attack__MNIST,_3_Adversaries__-_Distances.pdf}
    \end{subfigure}
    \begin{subfigure}{\textwidth}
        \centering
        \caption{Gaussian Noise Attack (model-boosting attack)}
        \label{fig:gaussian-attacks-mnist-mb-distance}
        \includegraphics[clip,trim={0 0 0 0},width=0.9\textwidth]{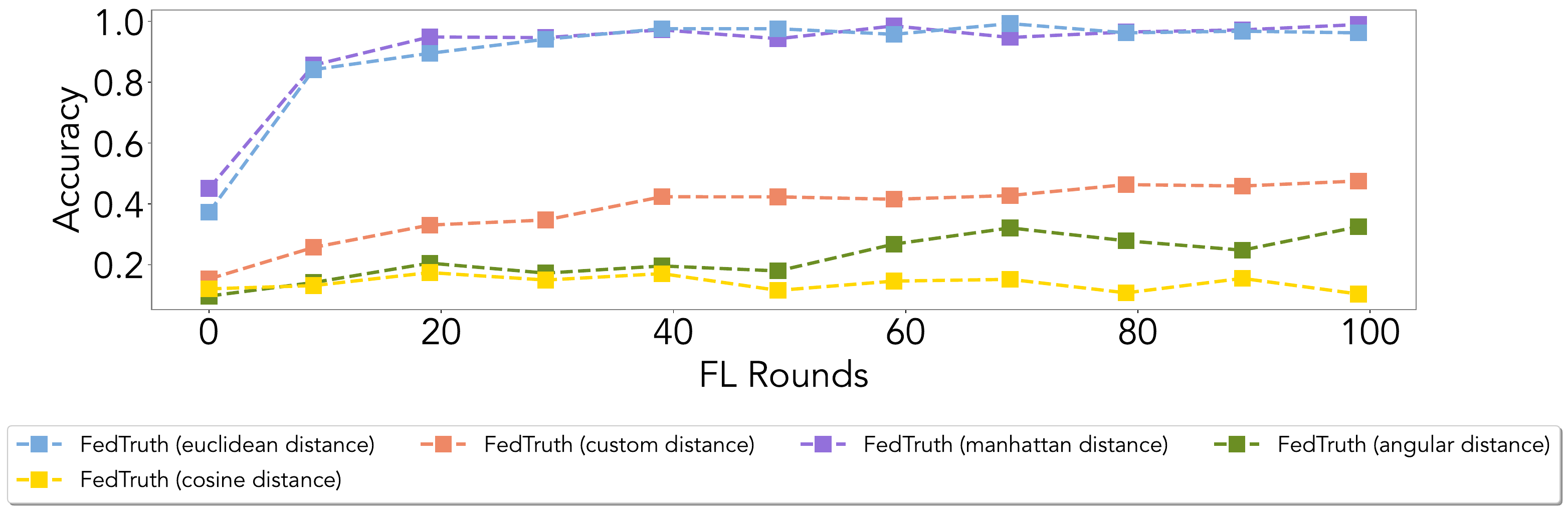}
    \end{subfigure}
    \captionsetup{font={normal}} 
    \caption{\textbf{Gaussian Noise Attack} (MNIST, 3 adversaries) - Comparison of Distance Metrics}
    \label{fig:gaussian-attacks-mnist-distance}
\end{figure}

\textbf{Gaussian Noise Attack - Comparison of Distance Functions}: 
Figure~\ref{fig:gaussian-attacks-mnist-distance} presents our results for the effectiveness of different distance metrics during the Gaussian noise attack with and without model-boosting. During both attacks, metrics based on \textit{vector magnitude differences} (Euclidean and Manhattan distances) consistently outperformed the other approaches, maintaining a final accuracy above $90\%$.

In contrast, metrics based on the vectors' direction, cosine, angular, and custom distances performed poorly during both experiments. Throughout the base attack, cosine, angular, and custom distance metrics maintained accuracies below $5\%$. During the model-boosting attack, the custom distance performed slightly better, with accuracy converging to around $40\%$, while angular distance converged to about $30\%$, and cosine distance remained low, converging to only around $5\%$.

This is consistent with our expectations, as adding arbitrary noise changes the magnitude of the adversarial updates, which is reflected in the strong performance of the Euclidean and Manhattan distances. However, these attacks do not significantly alter the angular difference when there are three adversaries present per round. We suspect that when the Gaussian noise attack is combined with model-boosting, the boosted updates make the custom distance function more effective at distinguishing between adversarial and benign updates, causing it to weight the adversarial updates lower. Nevertheless, the directional components in our custom metric ultimately limit its performance, causing the model to converge to a suboptimal global model.

\begin{figure*}[htp]
    \centering
    \begin{subfigure}{\textwidth}
        \captionsetup{font={tiny}}
        \centering
        \caption{PGD Attack (base attack) - Main Task Accuracy}
          \label{fig:pgd-attacks-mnist-base-ma-distance} 
        \includegraphics[clip,trim={0 5.4cm 0 0}, width=0.9\textwidth]{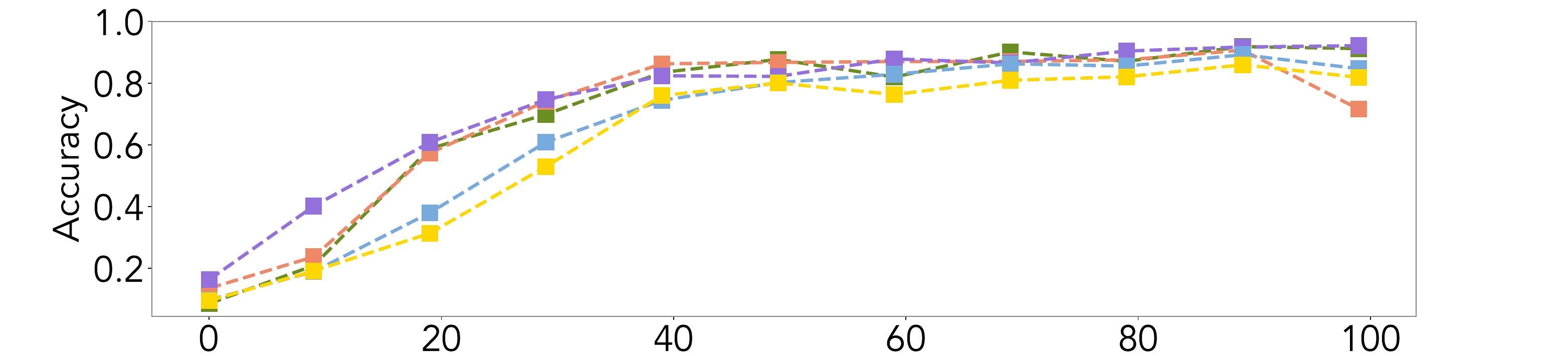}
    \end{subfigure}
    \hfill 
    \begin{subfigure}{\textwidth}
        \captionsetup{font={tiny}} 
        \centering
        \caption{PGD Attack (base attack) - Backdoor Accuracy}
          \label{fig:pgd-attacks-mnist-base-ba-distance} 
        \includegraphics[clip,trim={0 5.4cm 0 0}, width=0.9\textwidth]{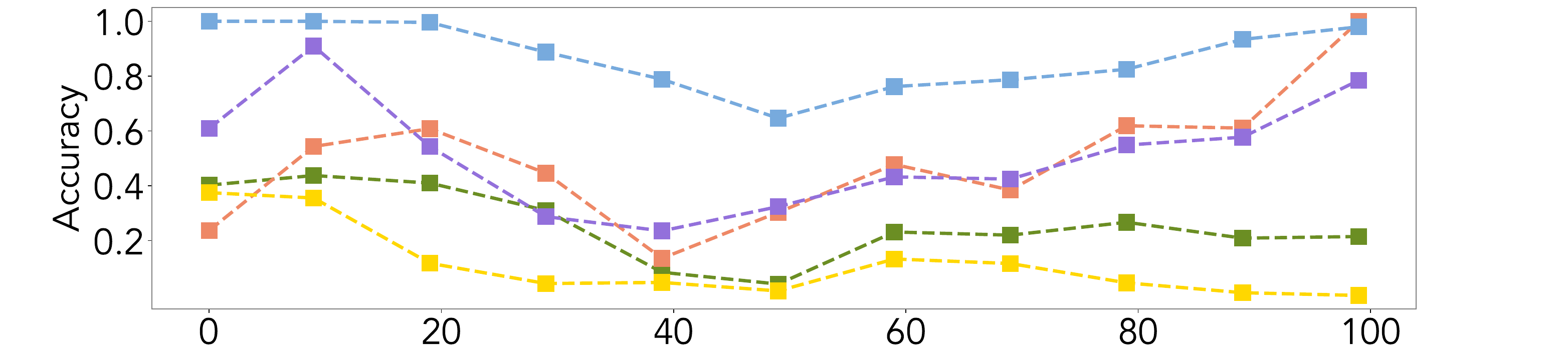}
    \end{subfigure}
    \begin{subfigure}{\textwidth}
        \captionsetup{font={tiny}}
        \centering
        \caption{PGD Attack (model-boosting attack) - Main Task Accuracy}
          \label{fig:pgd-attacks-mnist-mb-ma-distance} 
        \includegraphics[clip,trim={0 5.4cm 0 0}, width=0.9\textwidth]{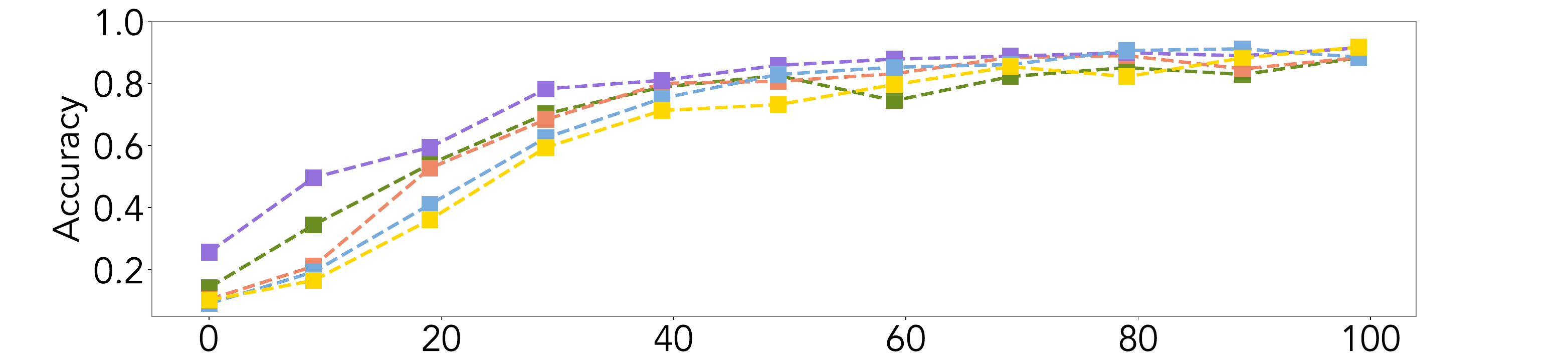}
    \end{subfigure}
    \hfill 
    \begin{subfigure}{\textwidth}
        \captionsetup{font={tiny}}
        \centering
        \caption{PGD Attack (model-boosting attack) - Backdoor Accuracy}
        \label{fig:pgd-attacks-mnist-mb-ba-distance} 
        \includegraphics[clip,trim={0 5.4cm 0 0}, width=0.9\textwidth]{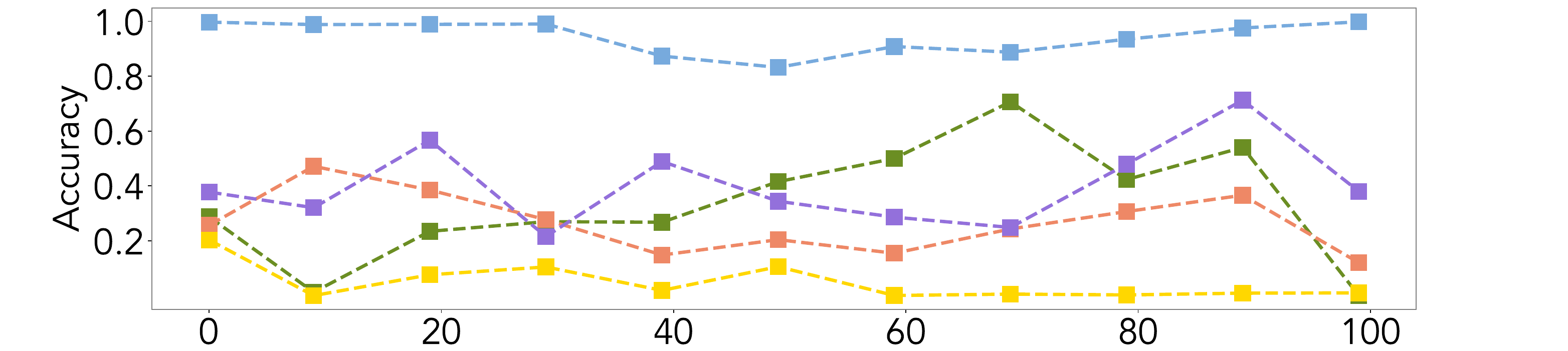}
    \end{subfigure}
    \begin{subfigure}{\textwidth}
        \captionsetup{font={tiny}}
        \centering
        \caption{PGD Attack (constrain-and-scale attack) - Main Task Accuracy}
        \label{fig:pgd-attacks-mnist-cs-ma-distance} 
        \includegraphics[clip,trim={0 5.4cm 0 0}, width=0.9\textwidth]{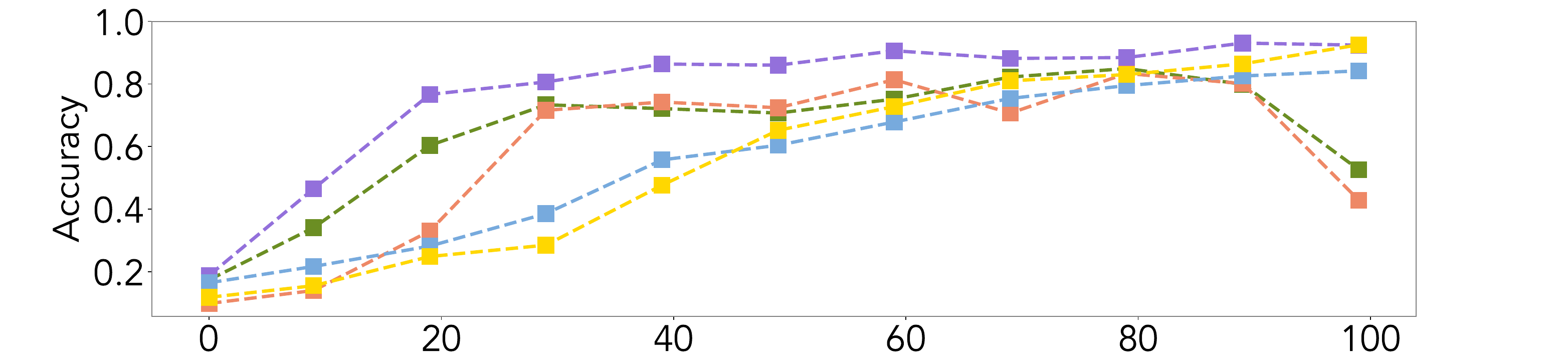}
    \end{subfigure}
    \hfill
    \begin{subfigure}{\textwidth}
        \captionsetup{font={tiny}}
        \centering
        \caption{PGD Attack (constrain-and-scale attack) - Backdoor Accuracy}
        \label{fig:pgd-attacks-mnist-cs-ba-distance} 
        \includegraphics[width=0.9\textwidth]{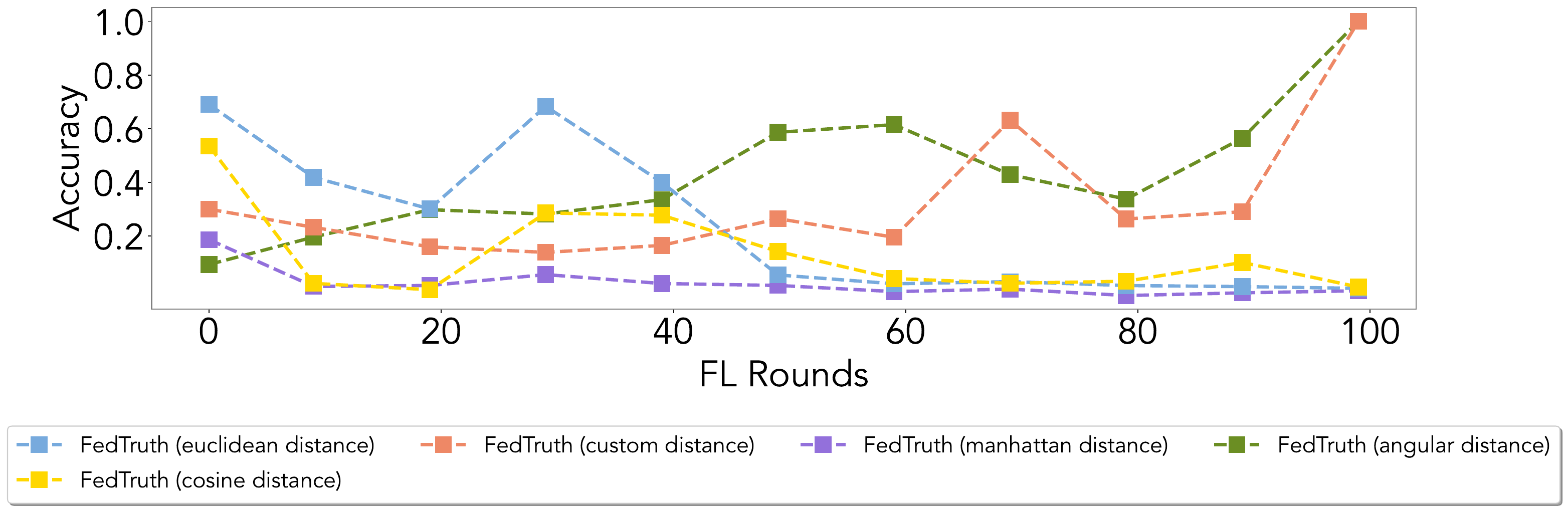}
    \end{subfigure}
  \caption{\normalsize \textbf{PGD Attack - Comparison of Distance Metrics} (MNIST, 3 adversaries)}
  \label{fig:pgd-attacks-mnist-distance} 
\end{figure*}

\textbf{PGD - Comparison of Distance Functions}: 
Figure~\ref{fig:pgd-attacks-mnist-distance} presents our findings for the PGD \textit{base}, \textit{with model-boosting}, and \textit{with constrain-and-scale} attacks. Figures~\ref{fig:pgd-attacks-mnist-base-ma-distance},~\ref{fig:pgd-attacks-mnist-mb-ma-distance}, and~\ref{fig:pgd-attacks-mnist-cs-ma-distance} illustrate how the PGD attacks impacted the \textit{main task} accuracy during these experiments. We observed through these results that all of the algorithms are able to reach a final accuracy above 80\% during all attack combinations, except for FedTruth (custom and angular distance) during the \textit{base} attack.
We observed a slight decrease in the convergence speed of FedTruth for the Euclidean and cosine distance metrics when running this experiment on the \textit{constrain-and-scale} version of the attack, as seen in Figures~\ref{fig:pgd-attacks-mnist-cs-ma-distance}. However, after the 100th iteration, the \textit{main task} accuracy increases back to above 80\% for these metrics. 


Figures~\ref{fig:pgd-attacks-mnist-base-ba-distance},~\ref{fig:pgd-attacks-mnist-mb-ba-distance}, and~\ref{fig:pgd-attacks-mnist-cs-ba-distance} present our findings on the effect of different distance metrics during the PGD attacks. Across all attacks, the cosine distance metric successfully eliminated the backdoor artifacts, reducing the final backdoor accuracy to under $5\%$. 

During the \textit{base} attack (Figure~\ref{fig:pgd-attacks-mnist-base-ba-distance}), Euclidean and custom distances both performed poorly, with backdoor accuracy remaining close to $100\%$ throughout training. Manhattan distance also struggled, reaching a final backdoor accuracy of $80\%$. In contrast, angular distance was more effective, ultimately reducing backdoor accuracy below $20\%$.
For the \textit{model boosting} attack (Figure~\ref{fig:pgd-attacks-mnist-mb-ba-distance}), Euclidean distance again failed to suppress the backdoor, with a final accuracy close to $100\%$. Custom distance showed improvement in this setting, lowering backdoor accuracy to below $20\%$. Manhattan distance achieved moderate results, ending at $40\%$ backdoor accuracy. Angular distance also performed well, though it required more iterations to converge, ultimately reaching a backdoor accuracy below $5\%$. 
In the \textit{constrain-and-scale} attack (Figure~\ref{fig:pgd-attacks-mnist-cs-ba-distance}), both Euclidean and Manhattan distances proved effective, each reducing backdoor accuracy to below $5\%$. The custom and angular distance metrics were ineffective, with a final backdoor accuracy near $100\%$.

Overall, these results demonstrate that cosine distance is consistently robust during all of these attack combinations, reliably reducing backdoor accuracy to below $5\%$. Euclidean and Manhattan distances are only effective when PGD is combined with the constrain-and-scale attacks. This highlights the importance of selecting an appropriate distance metric based on the specific attack scenario. We suspect that the strong performance of Euclidean and Manhattan distances during the PGD with constrain-and-scale attack is due to the model constraining-which slightly reduces the effectiveness of the model-combined with scaling, which makes the magnitude of the model easier to distinguish, thereby allowing these metrics to perform optimally. This effect is somewhat reflected in the model boosting results as well, where Manhattan distance is able to partially remove the backdoor and custom distance achieves an optimal global model.

\begin{figure*}[htp]
    \centering
    \begin{subfigure}{\textwidth}
        \captionsetup{font={tiny}}
        \centering
        \caption{Edge Case Attack (base attack) - Main Task Accuracy}
        \label{fig:egc-attacks-mnist-base-ma-distance} 
        \includegraphics[clip,trim={0 6cm 0 0}, width=0.9\textwidth]{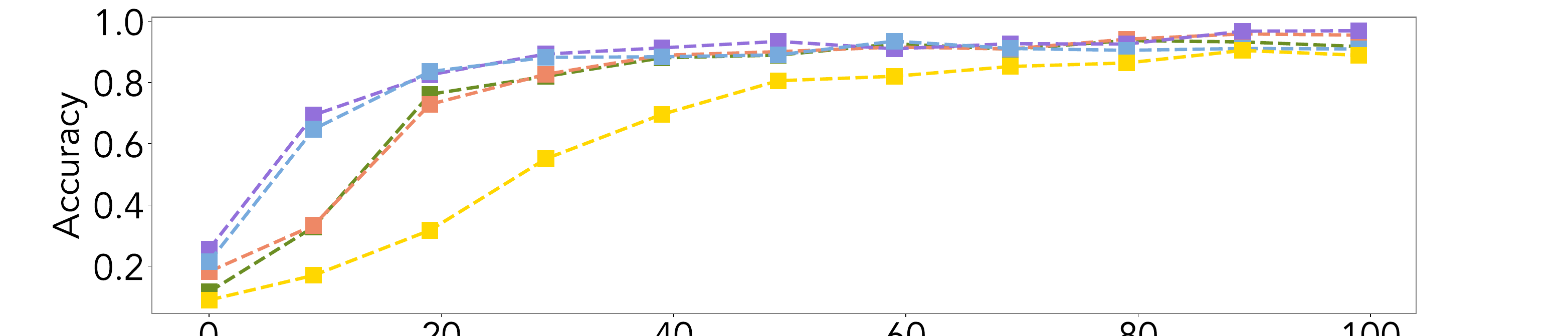}
    \end{subfigure}
    \hfill 
    \begin{subfigure}{\textwidth}
        \captionsetup{font={tiny}} 
        \centering
        \caption{Edge Case Attack (base attack) - Backdoor Accuracy}
        \label{fig:egc-attacks-mnist-base-ba-distance} 
        \includegraphics[clip,trim={0 6cm 0 0}, width=0.9\textwidth]{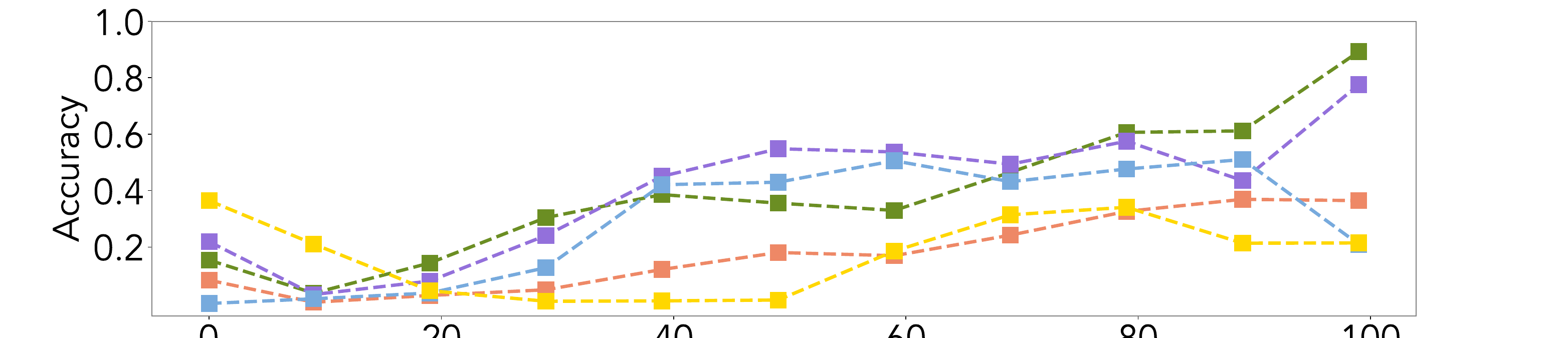}
    \end{subfigure}
    \begin{subfigure}{\textwidth}
        \captionsetup{font={tiny}}
        \centering
        \caption{Edge Case Attack (model-boosting attack) - Main Task Accuracy}
        \label{fig:egc-attacks-mnist-mb-ma-distance} 
        \includegraphics[clip,trim={0 6cm 0 0}, width=0.9\textwidth]{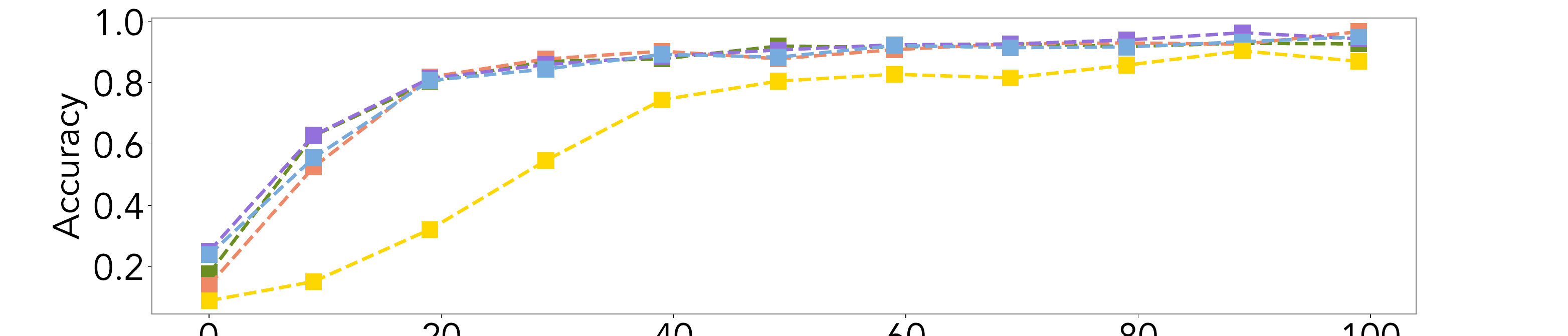}
    \end{subfigure}
    \hfill 
    \begin{subfigure}{\textwidth}
        \captionsetup{font={tiny}}
        \centering
        \caption{Edge Case Attack (model-boosting attack) - Backdoor Accuracy}
        \label{fig:egc-attacks-mnist-mb-ba-distance} 
        \includegraphics[clip,trim={0 6cm 0 0}, width=0.9\textwidth]{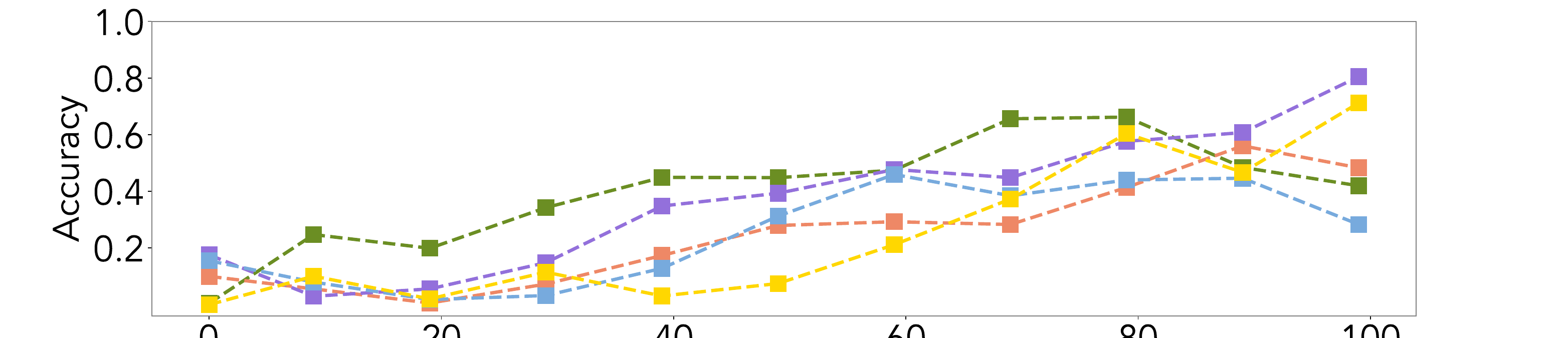}
    \end{subfigure}
    \begin{subfigure}{\textwidth}
        \captionsetup{font={tiny}}
        \centering
        \caption{Edge Case Attack (constrain-and-scale attack) - Main Task Accuracy}
        \label{fig:egc-attacks-mnist-cs-ma-distance} 
        \includegraphics[clip,trim={0 6cm 0 0}, width=0.9\textwidth]{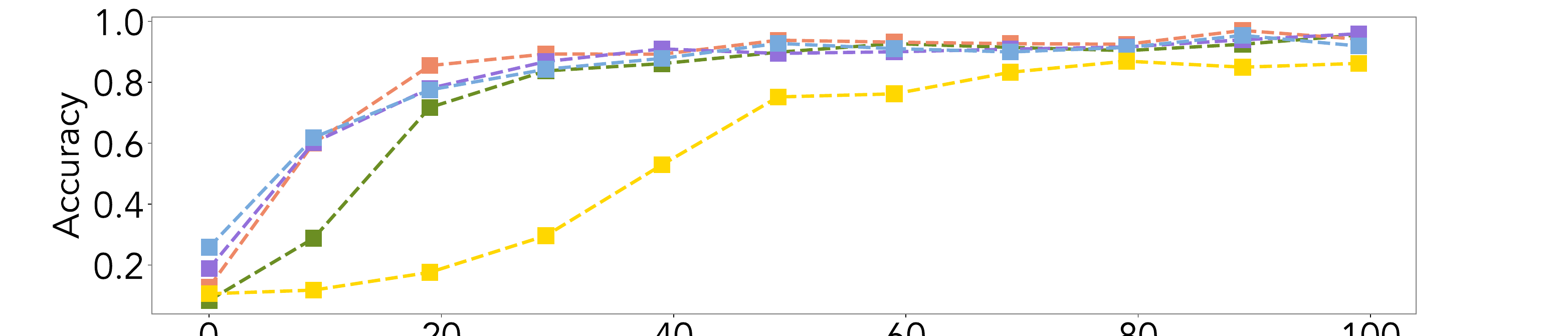}
    \end{subfigure}
    \hfill
    \begin{subfigure}{\textwidth}
        \captionsetup{font={tiny}}
        \centering
        \caption{Edge Case Attack (constrain-and-scale attack) - Backdoor Accuracy}
        \label{fig:egc-attacks-mnist-cs-ba-distance} 
        \includegraphics[width=0.9\textwidth]{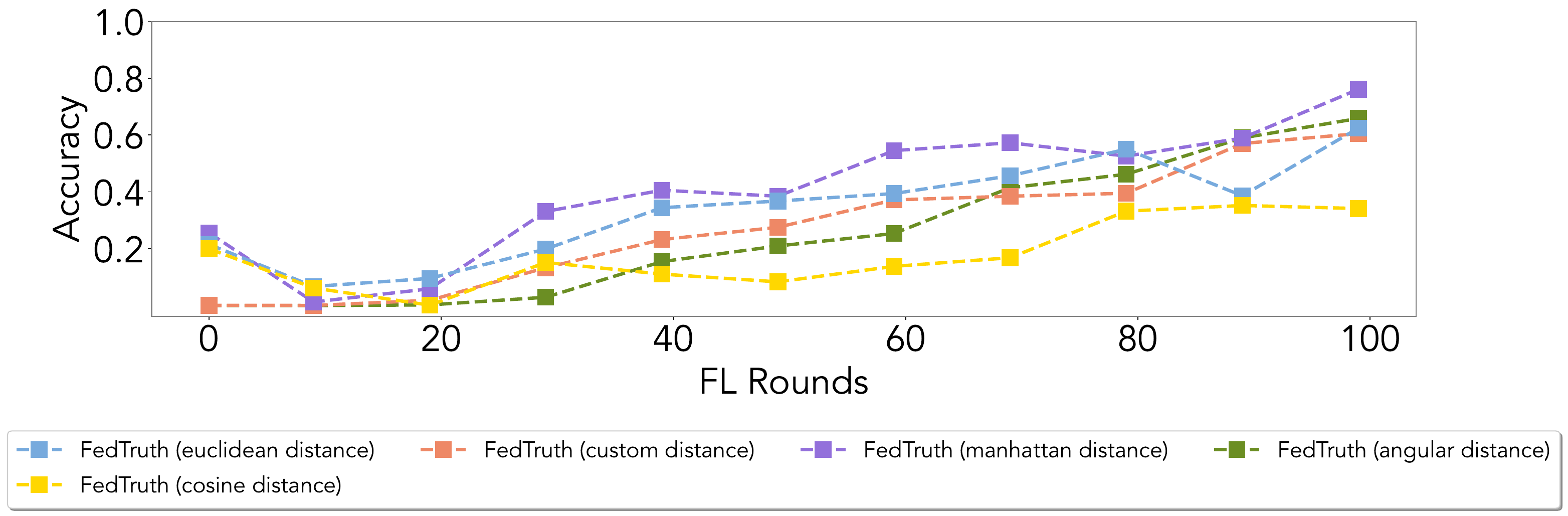}
    \end{subfigure}
  \caption{\normalsize \textbf{Edge Case Attack - Comparison of Distance Metrics} (MNIST, 3 adversaries)}
  \label{fig:egc-attacks-mnist-distance} 
\end{figure*}

\textbf{Edge-case Attack - Comparison of Distance Functions}:
Figure~\ref{fig:egc-attacks-mnist-distance} presents our findings on the impact of different distance metrics for both FedTruth and FedTruth-Layer during the edge-case attack.

Figures~\ref{fig:egc-attacks-mnist-base-ma-distance},~\ref{fig:egc-attacks-mnist-mb-ma-distance}, and~\ref{fig:egc-attacks-mnist-cs-ma-distance} present our findings for \textit{main task} accuracy during the edge-case attacks. We observe that all attacks reach convergence, with a final accuracy above $80\%$ across all attack configurations. However, the convergence rate is significantly slower in all of these attacks when using the cosine distance metric.

Figures~\ref{fig:egc-attacks-mnist-base-ba-distance},~\ref{fig:egc-attacks-mnist-mb-ba-distance}, and~\ref{fig:egc-attacks-mnist-cs-ba-distance} show our results for \textit{targeted task} accuracy during the edge-case attacks.
Figure~\ref{fig:egc-attacks-mnist-base-ba-distance} shows that during the base attack, the cosine distance and Euclidean distance metrics achieve the lowest backdoor accuracy, both falling below $15\%$, while FedTruth-Layer (cosine distance) maintains a final backdoor accuracy below $30\%$. Figure~\ref{fig:egc-attacks-mnist-mb-ba-distance} presents our results for the edge-case attacks with \textit{model-boosting}, where we see the final backdoor accuracy for cosine distance increase to 60\%. This is consistent with our findings for the DBA attack when combined with model-boosting, as seen in Section~{\ref{app:more-distance-functions-dba}}, where we suspect that the minimal change in the adversarial vectors' angles allows boosting to be effective at hijacking the global model. Euclidean distance still offers the best performance during this attack, finishing with a final backdoor accuracy of 20\%. During the constrain-and-scale attack (Figure~\ref{fig:egc-attacks-mnist-cs-ba-distance}), the best performing metric is once again cosine distance, finishing with a backdoor accuracy of 30\%, well below the other approaches. Additionally, Euclidean distance finishes with a backdoor accuracy of approximately 60\%. We suspect this is a result of constraining the amount the model can change during aggregation, causing the model to alter the angular difference enough to weight the adversarial models low enough to remove them during training.

\end{document}